\documentclass[12pt,a4paper]{article}
\usepackage{lmodern}
\usepackage[T1]{fontenc}

\usepackage[lmargin=71pt, tmargin=1.2in]{geometry}
\usepackage{graphicx}
\usepackage{gensymb}
\usepackage{amsmath,amssymb}
\usepackage{float}
\usepackage{algorithm}
\usepackage{algpseudocode}
\usepackage{subcaption}
\usepackage[authoryear]{natbib}
\setcitestyle{round}
\usepackage{matlab-prettifier}
\usepackage{microtype}
\usepackage{diagbox}

\usepackage{xcolor}
\usepackage[colorlinks=true,
            urlcolor={blue!60!black},
            linkcolor=black,
            citecolor=black]{hyperref}

\begin{document}
\thispagestyle{empty}

\begin{center}
    {\LARGE \bfseries
  IMU-Free Body-Frame State Estimation with Sparse Scene Flow for Quadcopters
    \par}
    
    \vspace{0.8em}
    
    {\large Daniel Grønhaug, Sofie Markeset, Mathias Kolberg}\\
    {\normalsize \texttt{dfgronha@math.uio.no, sofielma@ifi.uio.no, mathiko@ifi.uio.no}}
    
    \vspace{1em}
    \rule{0.6\textwidth}{0.4pt}
\end{center}

\vspace{1em}
\begin{abstract}
\noindent We present a vision-only state estimation 
system for X-configuration quadcopters equipped with 
a canonical stereo camera pair and no inertial 
sensors. The system operates entirely in the body 
frame, requiring only synchronised stereo images and 
motor thrust commands as inputs. A continuous-discrete 
extended Kalman filter on a composite manifold state 
$\langle SE(3),\,\mathbb{R}^3,\,\ldots\rangle$ 
maintains estimates of body-frame pose, velocity, 
angular velocity, gravity, and disturbances, 
using stationary scene points as implicit inertial 
references. Feature points are detected (FAST, 
Shi-Tomasi), tracked temporally (SSD, Lucas-Kanade) 
and matched across cameras (NCC), with 
prediction-assisted search regions constructed from 
filter-derived pose and point uncertainty. 
Chi-squared gating on the normalised innovation 
classifies points as stationary or moving; only 
stationary points enter the filter. Beyond the state 
estimate, the system produces a sparse 3D point 
cloud in which each point carries a position, 
velocity, and joint covariance. These are obtained through a 4-view (i.e.\ two stereo pairs for two timestamps) full bundle adjustment that jointly estimates point position and velocity, fusing stereo disparity and temporal parallax across two stereo observations, using the 
filter-derived relative pose as a prior. Feature 
points in the EKF do not enter the solver; their 
measurement information is reflected through the 
pose prior. Point cloud density is spatially 
adaptive: an external focus point directs allocation, 
producing dense coverage in the region of attention 
and sparse peripheral coverage from the feature 
points (akin to foveation). The output --- a body-frame state estimate,
a calibrated pose change, and a sparse
scene flow --- is intended as a measurement source
for a downstream world model (environment state and dynamics) anchored in the current
body frame, without dependence on GPS, IMU, or any
world-frame infrastructure, though the architecture
accommodates their future integration.
\end{abstract}
\vspace{1em}

\newpage

\setcounter{tocdepth}{2}
{\small\tableofcontents}
\newpage
\section{Introduction}

This document describes the design of a vision-only state and scene flow estimation system. The system takes synchronised stereo 
images and motor thrust commands as input and 
produces three outputs: (1)~the quadcopter's 
body-frame state (pose, velocity, angular 
velocity, gravity, disturbance) with covariance, 
(2)~the frame-to-frame pose change with 
covariance, and (3)~a sparse 3D point cloud 
with per-point velocity and uncertainty. \\

\noindent The architecture has two coupled estimators. 
The first is a continuous-discrete extended 
Kalman filter on a composite manifold state 
$\langle SE(3),\,\mathbb{R}^3,\,\ldots\rangle$ 
(Section~\ref{sec:ekf}). The filter maintains 
stationary scene points as inertial references: 
their body-frame positions evolve 
deterministically under the known kinematics 
(Section~\ref{sec:system_model}), and pixel 
measurements correct the state through 
covariance cross-correlations 
(Section~\ref{sec:measurement}). A gravity 
magnitude pseudo-measurement prevents drift 
in the gravity estimate. \\

\noindent The second estimator is a two-frame nonlinear
MAP solver (Section~\ref{sec:joint_solver})
that jointly estimates point positions and 
velocities with a pose change, using 
the filter-derived pose as a Lie algebra prior 
(Section~\ref{sec:relative_pose}) and stereo triangulated point estimates as priors on the points (Section~\ref{sec:depth_estimation}). Feature 
points in the EKF do not enter the solver; 
their information enters through the pose 
prior, avoiding double-counting. The solver 
handles all four stereo/mono visibility 
combinations (SS, SM, MS, MM), with 
mono-mono points constrained to the 
observable perpendicular velocity component 
(Section~\ref{sec:d_perp}), though we have constrained ourselves to the observation combination SS due to unresolved numerical instability. \\

\noindent Points are managed in three sets: feature 
points ($\mathcal{F}$) in the EKF for state 
estimation, pre-admission candidates 
($\mathcal{F}_{\mathrm{pre}}$) awaiting 
their first solve and stationarity test, and 
interest points ($\mathcal{I}$) steered by an 
external focus for task-directed dense 
coverage. All candidate and interest points 
enter the joint solver; the output point cloud 
is uniform across roles 
(Section~\ref{sec:algorithm_overview}). \\

\noindent The system operates without IMU, GPS, or any 
world-frame infrastructure. The body-frame 
formulation means the output is naturally 
suited to a world model anchored in the 
current body frame $B_k$, propagated between 
frames using the solver's pose change 
$\Delta\hat{T}_{\mathrm{solver}}$ rather than 
an accumulating absolute pose.

\subsection{Rotor Configuration}
Viewed from above, the rotors are numbered and 
spin as follows:
\begin{center}
\begin{tabular}{clll}
    \textbf{Rotor} & \textbf{Position} 
    & \textbf{Spin} 
    & \textbf{Body-frame location} \\
    \hline
    1 & Front-right & CW  
    & $+e_1,\;+e_2$ \\
    2 & Rear-right  & CCW 
    & $-e_1,\;+e_2$ \\
    3 & Rear-left   & CW  
    & $-e_1,\;-e_2$ \\
    4 & Front-left  & CCW 
    & $+e_1,\;-e_2$ \\
\end{tabular}
\end{center}
Adjacent rotors spin in opposite directions to 
cancel net reactive torque in steady flight. CW 
rotors (1,\,3) produce negative yaw torque 
(about $+e_3$); CCW rotors (2,\,4) produce 
positive yaw torque. All rotors produce thrust 
along $-e_3$ (upward in FRD, opposing gravity).

\subsection{Camera Configuration}
Two hardware-synchronised cameras with global 
shutter are rigidly mounted facing forward 
(along $+e_1$), arranged in a canonical stereo 
configuration: horizontally separated along 
$e_2$ with parallel optical axes. The left 
camera is at $-e_2$ and the right camera at 
$+e_2$ relative to the body centre, giving a 
baseline $b = \|\mathbf{t}_R - \mathbf{t}_L\|$. \\

\noindent Both cameras share the same orientation: their 
optical axes are aligned with the body $e_1$ 
axis, image $x$-axis with $e_2$, and image 
$y$-axis with $e_3$. The stereo overlap region 
(where both fields of view intersect) is 
centred on the forward direction. The left and 
right fringe zones are visible to only one 
camera. \\

\noindent The cameras are fixed-focus with deep depth of 
field. Images are delivered as synchronised 
$w \times h$ frames at rate $\lambda$\,Hz. 
Fish-eye lenses are preferred for wider 
field of view; HDR or dual-exposure capability 
is beneficial for varying lighting conditions.

\subsection{Priors}
\label{sec:priors}

All quantities that must be known or calibrated 
before the algorithm runs. These are constant 
throughout operation. Runtime values, units, 
sources, and mathematical types (manifolds, 
shapes) are maintained in 
\href{https://github.com/danielftg/blackbird-vio/blob/a8713f352ee5481de798098e39798ac93115d479/src/constants/calibration.yaml}{calibration.yaml}. 
Groups, with cross-references for how each is 
used:

\begin{itemize}
    \item \textbf{Coordinate frames.} Body $B$ 
    (FRD), cameras $C^L, C^R$ (RealSense), 
    Vicon marker $M$. Used throughout.
    
    \item \textbf{Camera parameters.} Intrinsics, 
    distortion, body-to-camera extrinsics in 
    $SE(3)$, frame rate, image resolution. 
    Stereo baseline $b$ derived. See 
    \S\ref{sec:projective}.
    
    \item \textbf{Quadcopter parameters.} Mass, 
    inertia, arm length, per-rotor thrust and 
    torque coefficients, drag, gravity. 
    Determine mixing matrix and dynamics. See 
    \S\ref{sec:dynamics}.
    
    \item \textbf{Noise parameters.} Process 
    noise spectral densities for actuators, 
    torques, gravity, and disturbance; pixel 
    measurement noise; gravity 
    pseudo-measurement noise. See 
    \S\ref{sec:system_model}, 
    \S\ref{sec:measurement}.
    
    \item \textbf{Initial covariance.} Diagonal 
    of $P_0$ for each state component, 
    reflecting the assumption that the quadcopter 
    starts stationary, level, at the body-frame 
    origin. See \S\ref{sec:system_model}.
\end{itemize}

\noindent Several entries are flagged
\texttt{[EST]} in the file. These are educated
guesses rather than measurements: the inertia
tensor $\mathbf{J}$, the drag coefficient $C_d$,
and the camera-to-body rotation
$\mathbf{R}_{BC}$, which is the nominal mount with
any mounting misalignment unmodelled. Identifying
them properly is outstanding work.

\subsection{Inputs}
\label{sec:alg_in}

All objects the algorithm receives at the start 
of time step $t_k$. The control input and focus 
are indexed $k{-}1$ because they reflect 
commands applied during the interval 
$[t_{k-1},\,t_k)$:
\begin{align}
    L_k,\; R_k 
    &\in \mathbb{R}^{1 \times h \times w}
        & &\text{Stereo pair with timestamp 
              $t_k$} \\
    \mathbf{u}_{k-1} 
    &= (T_1,\, T_2,\, T_3,\, T_4)^\top 
    \in \mathbb{R}^4
        & &\text{Rotor thrusts during 
              $[t_{k-1},\,t_k)$ 
              (\S\ref{sec:system_model})} \\
    \mathbf{F}_{k-1} 
    &\in \mathbb{R}^3
        & &\text{Focus point in $B_{k-1}$ 
              (\S\ref{sec:cv})} \\
    \sigma_{F,k-1} 
    &\in \mathbb{R}_{>0}
        & &\text{Focus spread 
              (\S\ref{sec:cv})}
\end{align}

\subsection{Outputs}
\label{sec:alg_out}

All objects the algorithm produces at the end 
of time step $t_k$.

\subsubsection{Core State (body frame)}

The EKF state with feature point rows and 
columns stripped (\S\ref{sec:algorithm_overview}):
\begin{align}
    \hat{T}_{B_k,B_0} &\in SE(3)
        & &\text{Body pose} 
        \\[1ex]
    \hat{\mathbf{v}} 
    &\in \mathbb{R}^3
        & &\text{Linear velocity in $B_k$} \\
    \hat{\boldsymbol{\omega}} 
    &\in \mathbb{R}^3
        & &\text{Angular velocity in $B_k$} \\
    \hat{\mathbf{g}}^B 
    &\in \mathbb{R}^3
        & &\text{Gravity in $B_k$ 
              ($\|\hat{\mathbf{g}}^B\| 
              \approx g$)} \\
    \hat{\mathbf{d}}^B 
    &\in \mathbb{R}^3
        & &\text{Wind/disturbance in $B_k$} \\
    \mathbf{P}^{\mathrm{core}} 
    &\in \mathbb{R}^{18\times18}
        & &\text{Core state covariance}
\end{align}
The pose $\hat{T}_{B_k,B_0}$ accumulates drift and is 
suitable for coarse waypoint navigation; for 
local geometry, use $\Delta\hat{T}$ below.

\subsubsection{Pose Change}

From the joint solver 
(\S\ref{sec:joint_solver}):
\begin{align}
    \Delta\hat{T}_{\mathrm{solver}} 
    &= \hat{T}_{B_k,B_{k-1}} \in SE(3)
        & &\text{Frame-to-frame pose change} \\
    \boldsymbol{\Sigma}_{\Delta\xi}
    ^{\mathrm{solver}} 
    &\in \mathbb{R}^{6\times6}
        & &\text{Pose change covariance 
              (Lie algebra)}
\end{align}
This is the recommended transform for 
propagating a downstream world model anchored 
in the current body frame.

\subsubsection{Point Cloud (body frame $B_k$)}

\begin{equation}
    \mathcal{C}_k = \left\{\left(
    \hat{\mathbf{p}}_i,\;
    \hat{\mathbf{v}}_i^{\mathbf{p}},\;
    \boldsymbol{\Sigma}_i,\;
    \texttt{role}_i,\;
    \texttt{stage}_i
    \right)\right\}_{i=1}^{N}
\end{equation}
where for each point $i$:
\begin{align}
    \hat{\mathbf{p}}_i 
    &\in \mathbb{R}^3
        & &\text{Position in $B_k$} \\
    \hat{\mathbf{v}}_i^{\mathbf{p}} 
    &\in \mathbb{R}^3
        & &\text{Velocity in $B_k$ 
              (stage~2 only)} \\
    \boldsymbol{\Sigma}_i 
    &\in \mathbb{R}^{d_i \times d_i}
        & &\text{Covariance (dimension 
              depends on source)} \\
    \texttt{role}_i 
    &\in \{\mathcal{F},\,
    \mathcal{F}_{\mathrm{pre}},\,
    \mathcal{I}\}
        & &\text{Point role} \\
    \texttt{stage}_i 
    &\in \{1,\, 2\}
        & &\text{Single-frame or two-frame}
\end{align}

The content depends on the point's source:

\paragraph{Stage~2 
($\mathcal{F}_{\mathrm{pre}}$, $\mathcal{I}$).}
Position, velocity, and joint covariance from 
the joint solver, transformed to $B_k$ 
(\S\ref{sec:joint_solver}). For SS/SM/MS: $\boldsymbol{\Sigma}_i 
\in \mathbb{R}^{6\times6}$ (full 
position-velocity). For MM: 
$\boldsymbol{\Sigma}_i 
\in \mathbb{R}^{4\times4}$ 
(position and perpendicular velocity 
$v_\perp$); the velocity field is 
$\hat{\mathbf{v}}_i^{\mathbf{p}} 
= \hat{v}_{\perp,i}\,
\Delta\hat{\mathbf{R}}\,\mathbf{d}_\perp$ 
with the along-epipolar component 
unobservable and set to zero.

\paragraph{Stage~1 stereo 
($\mathcal{F}_{\mathrm{pre}}$, $\mathcal{I}$).}
Position and covariance from stereo disparity 
(\S\ref{sec:depth_estimation}):
$\boldsymbol{\Sigma}_i 
= \boldsymbol{\Sigma}_{p,i} 
\in \mathbb{R}^{3\times3}$. Velocity not yet 
available.

\paragraph{Feature points ($\mathcal{F}$).}
Position and marginal covariance extracted from 
the EKF state and covariance:
$\boldsymbol{\Sigma}_i 
= \boldsymbol{\Sigma}_{p,i}^{\mathrm{EKF}} 
\in \mathbb{R}^{3\times3}$. Velocity not 
returned (the EKF does not maintain per-point 
velocity; feature points are assumed 
stationary, and this assumption is repeatedly tested).

\newpage
\section{Mathematical Preliminaries}
\label{sec:prelim}
\subsection{Lie Algebra on $SE(3)$}
\label{sec:lie_algebra}
We consider the matrix Lie group $T \in SE(3), \ T=\begin{pmatrix}
    \mathbf{R} & \mathbf{t} \\
    \mathbf{0}^\top & 1
\end{pmatrix}$, with 
Lie algebra elements
$\boldsymbol{\xi}^\wedge \in \mathfrak{se}(3)$ and coordinate vectors
$\boldsymbol{\xi} \in \mathbb{R}^6$
\citep{sola2021microlietheorystate}. \\
\noindent A coordinate vector and its Lie algebra element are:
\begin{gather*}
    \boldsymbol{\rho} = \text{Translation}, \quad 
    \boldsymbol{\phi} = \text{Rotation} \\
    \boldsymbol{\xi} = \begin{pmatrix}
    \boldsymbol{\rho} \\ \boldsymbol{\phi}
    \end{pmatrix} = \begin{pmatrix}
    \rho_1 & \rho_2 & \rho_3 & \phi_1 & \phi_2 & \phi_3
    \end{pmatrix}^\top \\
    \boldsymbol{\xi}^{\wedge} 
    = \begin{bmatrix} [\boldsymbol{\phi}]_\times & \boldsymbol{\rho} \\ \mathbf{0}^\top & 0 \end{bmatrix} \in \mathbb{R}^{4\times 4}, \qquad
    (\boldsymbol{\xi}^{\wedge})^{\vee} 
    = \boldsymbol{\xi}
\end{gather*}
The rotation $\boldsymbol{\phi}$ can be represented in angle-axis form $\boldsymbol{\phi} = \theta \mathbf{u}$. A coordinate vector is related to a pose and vice versa by
(where $\exp$ is the matrix exponential and $\log$ its 
inverse):
\begin{gather*}
    T_{\boldsymbol{\xi}} 
    = \operatorname{Exp}(\boldsymbol{\xi}) 
    = \exp(\boldsymbol{\xi}^{\wedge}), \qquad
    \boldsymbol{\xi}_T 
    = \operatorname{Log}(T) 
    = \log(T)^{\vee} \\[3ex]
    \mathbf{R} = \mathrm{Exp}_{so(3)}(\boldsymbol{\phi}) := \mathbf{I} + \sin\theta[\mathbf{u}]_\times + (1 - \cos\theta)[\mathbf{u}]_\times^2 \\
    \boldsymbol{\phi} = \mathrm{Log}_{so(3)}(\mathbf{R}) := \frac{\theta}{2\sin\theta}(\mathbf{R} - \mathbf{R}^\top)^\vee, \qquad \theta = \arccos\!\left(\frac{\mathrm{tr}(\mathbf{R}) - 1}{2}\right) \\
    \mathbf{V}(\boldsymbol{\phi}) = \mathbf{I} + \frac{1 - \cos\theta}{\theta}[\mathbf{u}]_\times + \frac{\theta - \sin\theta}{\theta}[\mathbf{u}]_\times^2 \\
    \mathbf{V}^{-1}(\boldsymbol{\phi}) = \mathbf{I} - \frac{\theta}{2}[\mathbf{u}]_\times + \left(1 - \frac{\theta \sin\theta}{2(1-\cos\theta)}\right)[\mathbf{u}]_\times^2 \\[3ex]
    \mathrm{Exp}(\boldsymbol{\xi}) :=\begin{bmatrix} \mathrm{Exp}_{so(3)}(\boldsymbol{\phi}) & \mathbf{V}(\boldsymbol{\phi})\boldsymbol{\rho} \\ \mathbf{0}^\top & 1 \end{bmatrix} \\
    \mathrm{Log}(T) := \begin{bmatrix} \mathbf{V}^{-1}(\boldsymbol{\phi})\mathbf{t} \\ \boldsymbol{\phi} \end{bmatrix}
\end{gather*}
Addition and subtraction on this Lie group 
(right perturbation convention):
\begin{gather*}
    T \oplus \boldsymbol{\xi} 
    = T \cdot \operatorname{Exp}(\boldsymbol{\xi}) 
    \in SE(3) \\
    T_1 \ominus T_2 
    = \operatorname{Log}(T_2^{-1} \cdot T_1) 
    \in \mathbb{R}^6
\end{gather*}

\paragraph{Adjoint of $SE(3)$.}
The adjoint matrix $\mathrm{Ad}_T$ linearly maps tangent vectors 
between reference frames 
\citep[Example~6, Appendix~D Eq.~175]{sola2021microlietheorystate}:
\begin{equation}\label{eq:adjoint_se3}
    \mathrm{Ad}_T = \begin{pmatrix}
    \mathbf{R} & [\mathbf{t}]_\times\mathbf{R} \\
    \mathbf{0} & \mathbf{R}
    \end{pmatrix} \in \mathbb{R}^{6\times 6}
\end{equation}
satisfying $(\,T\,\boldsymbol{\xi}^\wedge\,T^{-1}\,)^\vee 
= \mathrm{Ad}_T\,\boldsymbol{\xi}$.

\subsection{Composite Manifold State}
\label{sec:composite}

The estimation state contains both Lie group and 
Euclidean components. Following 
\citep[Section~IV, Eqs.~84--90]{sola2021microlietheorystate}, we define a composite manifold 
$\mathcal{M} = \langle SE(3),\,\mathbb{R}^3,\,
\mathbb{R}^3,\,\ldots\rangle$ with state:
\begin{equation}\label{eq:composite_state}
    \mathcal{X} = \langle\, T,\;
    \mathbf{v},\;\boldsymbol{\omega},\;
    \mathbf{g}^B,\;\mathbf{d}^B,\;
    \mathbf{p}_1^B,\;\ldots,\;
    \mathbf{p}_{N_f}^B \,\rangle
\end{equation}
where $T \in SE(3)$ is the body pose and all other 
components are in $\mathbb{R}^3$.

\paragraph{Composite plus and minus.}
The composite operators $\hat{\oplus}$ and $\hat{\ominus}$ 
act on each block according to its own manifold 
\citep[Eqs.~86--87]{sola2021microlietheorystate}:
\begin{equation}\label{eq:composite_plus}
    \mathcal{X} \,\hat{\oplus}\, \delta\mathbf{x} 
    = \langle\, T \oplus \delta\boldsymbol{\xi},\;
    \mathbf{v} + \delta\mathbf{v},\;
    \boldsymbol{\omega} + \delta\boldsymbol{\omega},\;
    \mathbf{g}^B + \delta\mathbf{g}^B,\;
    \mathbf{d}^B + \delta\mathbf{d}^B,\;
    \mathbf{p}_1^B + \delta\mathbf{p}^B_1,\;\ldots\,\rangle
\end{equation}
where $T \oplus \delta\boldsymbol{\xi} 
= T \cdot \mathrm{Exp}(\delta\boldsymbol{\xi})$ is the 
$SE(3)$ right-plus and $+$ is standard vector addition 
\citep[Appendix~E Eq.~186]{sola2021microlietheorystate}. The composite perturbation vector is:
\begin{equation}\label{eq:composite_delta}
    \delta\mathbf{x} = \begin{pmatrix}
    \delta\boldsymbol{\xi} \\ \delta\mathbf{v} \\ 
    \delta\boldsymbol{\omega} \\ \delta\mathbf{g}^B \\ 
    \delta\mathbf{d}^B \\ \delta\mathbf{p}_1^B \\ 
    \vdots \end{pmatrix} 
    \in \mathbb{R}^{18+3N_f}
\end{equation}
with $\delta\boldsymbol{\xi} 
= (\delta\boldsymbol{\rho},\,\delta\boldsymbol{\phi}) 
\in \mathbb{R}^6$ the $SE(3)$ tangent vector. The 
composite minus is the inverse: 
$\delta\mathbf{x} = \mathcal{X} \,\hat{\ominus}\, 
\hat{\mathcal{X}}$ gives 
$\delta\boldsymbol{\xi} = T \ominus \hat{T} 
= \mathrm{Log}(\hat{T}^{-1}T)$ for the $SE(3)$ block 
and $\delta\mathbf{v} = \mathbf{v} - \hat{\mathbf{v}}$ 
for the $\mathbb{R}^3$ blocks.

\paragraph{Composite derivative.}
Jacobians of functions on the composite state are 
computed block-wise 
\citep[Eq.~88--89]{sola2021microlietheorystate}: each 
block $\partial f_i / \partial\mathcal{X}_j$ uses the 
manifold derivative (Eq.~41a) for Lie group components 
and the standard derivative for $\mathbb{R}^3$ 
components. We denote this Jacobian as: $\frac{Df(\mathcal{X})}{D\mathcal{X}}$. This derivative coincides with the perturbation derivative, which in turn reduces to the standard partial derivative for Euclidean components. 

\subsection{Projective Geometry}
\label{sec:projective}

See \citep{szeliski2010computer} and 
\citep{ma2005invitation} for comprehensive treatments.

\subsubsection{Pinhole Camera Model}

A 3D point passes through several transformations to 
become a pixel.

\paragraph{Homogeneous coordinates.}
\begin{align}
    \tilde{\cdot} \,:\, \mathbf{p} &\mapsto 
    \begin{pmatrix} \mathbf{p} \\ 1 \end{pmatrix}, \quad
    (\tilde{\cdot})^{-1} : 
    \begin{pmatrix} \mathbf{p} \\ s \end{pmatrix} 
    \mapsto \frac{\mathbf{p}}{s}
    & &\text{Homogeneous representation}
\end{align}

\paragraph{Perspective divide.}
Projects a 3D camera-frame point to normalised image 
coordinates, discarding depth:
\begin{equation}
    \pi_n(\mathbf{p}) = \frac{1}{p_z}
    \begin{pmatrix} p_x \\ p_y \end{pmatrix} 
    = \begin{pmatrix} x_n \\ y_n \end{pmatrix}
\end{equation}

\paragraph{Lens distortion.}
Maps undistorted normalised coordinates to distorted 
normalised coordinates. Model-dependent 
(Brown-Conrady, equidistant, etc.):
\begin{equation}
    D : \mathbb{R}^2 \to \mathbb{R}^2, \quad
    D(x_n, y_n) = (x_d, y_d)
\end{equation}

\paragraph{Camera intrinsics.}
Maps distorted normalised coordinates to pixel 
coordinates:
\begin{equation}
    \mathbf{K} = \begin{pmatrix} 
    f_u & 0 & c_u \\ 0 & f_v & c_v \\ 0 & 0 & 1 
    \end{pmatrix}, \qquad
    \kappa(x_d, y_d) = \begin{pmatrix} 
    f_u\,x_d + c_u \\ f_v\,y_d + c_v \end{pmatrix}
\end{equation}

\paragraph{Full projection.}
From camera frame to pixel:
\begin{equation}
    \boldsymbol{\pi} = \kappa \circ D \circ \pi_n 
    : \mathbb{R}^3 \to \mathbb{R}^2
\end{equation}

\paragraph{Inverse projection.}
From pixel to normalised coordinates and then to a ray 
(up to unknown depth $Z$):
\begin{equation}
    (u, v) \;\xrightarrow{\kappa^{-1}}\; (x_d, y_d) 
    \;\xrightarrow{D^{-1}}\; (x_n, y_n) 
    \;\xrightarrow{\times\,Z}\; 
    \mathbf{p}^c = Z\begin{pmatrix} x_n \\ y_n \\ 1 
    \end{pmatrix}
\end{equation}
The depth $Z$ cannot be recovered from a single image; 
projection is a many-to-one mapping.

\subsubsection{Camera Extrinsics}

\paragraph{Rotation convention.}
$\mathbf{R}_{AB}$ rotates vectors from frame $B$ into 
frame $A$: $\mathbf{v}^A = \mathbf{R}_{AB}\,\mathbf{v}^B$. 
Frames: $B_0$ (initial body / world), $B$ (body), 
$L$ (left camera), $R$ (right camera).

\paragraph{Body to camera.}
A body-frame point $\mathbf{p}^B$ in camera 
$c \in \{L, R\}$:
\begin{equation}
    \mathbf{p}^c = \mathbf{R}_{cB}\,\mathbf{p}^B 
    + \mathbf{t}_c
\end{equation}
The full projection from body frame to pixel:
\begin{equation}\label{eq:body_to_pixel}
    \mathbf{u} = \boldsymbol{\pi}\!\left(
    \mathbf{R}_{cB}\,\mathbf{p}^B + \mathbf{t}_c\right)
\end{equation}

\paragraph{Camera to camera.}
For canonical stereo with baseline 
$b = \|\mathbf{t}_R - \mathbf{t}_L\|$:
\begin{equation}
    \mathbf{p}^R = \mathbf{R}_{RL}\,\mathbf{p}^L 
    + \mathbf{t}_{RL}
\end{equation}
For parallel cameras: $\mathbf{R}_{RL} = \mathbf{I}$, 
$\mathbf{t}_{RL} = (-b, 0, 0)^\top$.

\subsubsection{Stereo Geometry}

For a rectified stereo pair, corresponding points lie on 
the same horizontal scanline (epipolar line). The 
horizontal disparity encodes depth:
\begin{align}
    d &= u_L - u_R 
        & &\text{Disparity (pixels)} \\
    Z &= \frac{f_u \cdot b}{d} 
        & &\text{Depth from disparity}
\end{align}
Depth uncertainty from first-order propagation of 
disparity noise $\sigma_d$:
\begin{equation}
    \sigma_Z = \left|\frac{\partial Z}{\partial d}
    \right|\sigma_d 
    = \frac{f_u\,b}{d^2}\,\sigma_d 
    = \frac{Z^2}{f_u\,b}\,\sigma_d
\end{equation}
Depth uncertainty grows quadratically with distance.

\subsubsection{Epipolar Constraint}

Given a point in the left image at pixel $(u_L, v_L)$, 
its correspondence in the right image is constrained to 
lie on a line (the epipolar line). For a rectified pair 
this is simply $v_R = v_L$, the same row.

\subsection{Rigid Body Dynamics}
\label{sec:dynamics}

These equations can among others be found in 
\citep{beard2012small}.

\begin{align}
    \sum\mathbf{F} &= m\dot{\mathbf{v}}
        & &\text{Newton's second law} \\
    (\dot{\mathbf{v}})_{\text{inertial}} &=
    (\dot{\mathbf{v}})_{\text{body}}
    + \boldsymbol{\omega} \times \mathbf{v}
        & &\text{Transport theorem}
        \label{eq:transport_general} \\
    \mathbf{J}\dot{\boldsymbol{\omega}} 
    + \boldsymbol{\omega} \times 
    (\mathbf{J}\boldsymbol{\omega}) &= \boldsymbol{\tau}
        & &\text{Euler's rotation equations} \\
    \boldsymbol{\tau} &= \mathbf{r} \times \mathbf{F}
        & &\text{Torque} \\
    \mathbf{F}_{\text{drag}} &= -C_d\,\mathbf{v}
        & &\text{Linear drag} \\
    T &= k_f\,\omega_r^2
        & &\text{Propeller thrust} \\
    Q &= k_m\,\omega_r^2
        & &\text{Reactive propeller torque} \\
    [\mathbf{a}]_\times &= \begin{pmatrix} 
    0 & -a_3 & a_2 \\ a_3 & 0 & -a_1 \\ 
    -a_2 & a_1 & 0 \end{pmatrix}
        & &\text{Skew-symmetric (cross-product) matrix}
\end{align}

\paragraph{Transport theorem for world-fixed vectors.}
A vector $\mathbf{a}^{B_0}$ constant in the $B_0$ frame, 
expressed in the rotating body frame as 
$\mathbf{a}^B = \mathbf{R}_{B,B_0}\,\mathbf{a}^{B_0}$, 
satisfies:
\begin{equation}\label{eq:transport_wfv}
    \dot{\mathbf{a}}^B = -\boldsymbol{\omega}
    \times \mathbf{a}^B
\end{equation}
Derived from 
$\dot{\mathbf{R}}_{B,B_0} = -[\boldsymbol{\omega}]_\times\,
\mathbf{R}_{B,B_0}$  \citep[Example~4, p.~5]{sola2021microlietheorystate}.

\subsection{Covariance and Chi-Squared Tests}
\label{sec:cov_stats}

\subsubsection{Covariance Identities}
\label{sec:cov_identities}

Given a function $\mathbf{y} = g(\mathbf{x})$ with
$\mathbf{x} \sim \mathcal{N}(\bar{\mathbf{x}},
\boldsymbol{\Sigma}_x)$, the output covariance is 
\citep{barshalom2001estimation}:
\begin{equation}\label{eq:cov_prop}
    \boldsymbol{\Sigma}_y 
    = \mathbf{J}\,\boldsymbol{\Sigma}_x\,
    \mathbf{J}^\top, 
    \qquad \mathbf{J} 
    = \frac{\partial g}{\partial \mathbf{x}}
    \bigg|_{\bar{\mathbf{x}}}
\end{equation}

\noindent The following standard identities are used throughout
the document. For zero-mean random vectors
$\mathbf{a}$, $\mathbf{b}$ and deterministic matrices 
$\mathbf{M}$, $\mathbf{N}$ 
\citep{barshalom2001estimation}:

\paragraph{Cross-covariance.}
\begin{equation}\label{eq:cross_cov_def}
    \mathrm{Cov}(\mathbf{a},\,\mathbf{b}) 
    = \mathrm{E}[\mathbf{a}\,\mathbf{b}^\top]
\end{equation}

\paragraph{Linear transformation.}
If $\mathbf{c} = \mathbf{M}\mathbf{a}$, then:
\begin{equation}\label{eq:cov_linear}
    \mathrm{Cov}(\mathbf{c}) 
    = \mathbf{M}\,\mathrm{Cov}(\mathbf{a})\,
    \mathbf{M}^\top
\end{equation}

\paragraph{Cross-covariance under linear 
transformation.}
If $\mathbf{c} = \mathbf{M}\mathbf{a}$, then:
\begin{equation}\label{eq:cross_cov_linear}
    \mathrm{Cov}(\mathbf{c},\,\mathbf{b}) 
    = \mathbf{M}\,\mathrm{Cov}(\mathbf{a},\,
    \mathbf{b})
\end{equation}

\paragraph{Independence.}
If $\mathbf{a}$ and $\mathbf{b}$ are independent:
\begin{equation}\label{eq:cov_indep}
    \mathrm{Cov}(\mathbf{a},\,\mathbf{b}) 
    = \mathbf{0}
\end{equation}

\paragraph{Sum with independent term.}
If $\mathbf{c} = \mathbf{M}\mathbf{a} + \mathbf{n}$ 
with $\mathbf{n}$ independent of both $\mathbf{a}$ 
and $\mathbf{b}$:
\begin{equation}\label{eq:cross_cov_sum}
    \mathrm{Cov}(\mathbf{c},\,\mathbf{b}) 
    = \mathbf{M}\,\mathrm{Cov}(\mathbf{a},\,
    \mathbf{b})
\end{equation}

\subsubsection{Chi-Squared Hypothesis Test}
Following \citep[p.~57]{barshalom2001estimation}, if 
$\mathbf{x} \sim 
\mathcal{N}(\bar{\mathbf{x}}, \mathbf{P})$ is an 
$n$-dimensional Gaussian, then:
\begin{equation}
    q = (\mathbf{x} - \bar{\mathbf{x}})^\top 
    \mathbf{P}^{-1} 
    (\mathbf{x} - \bar{\mathbf{x}}) \sim \chi^2_n,
\end{equation}
with $\mathrm{E}[q] = n$ and 
$\mathrm{var}[q] = 2n$.

\newpage
\section{Extended Kalman Filter}
\label{sec:ekf}
The Extended Kalman Filter on a Composite Manifold State\footnote{Closely related to Multiplicative EKF (MEKF) , Error State KF (ESKF), M-CD-EKF, ...} maintains a Gaussian 
approximation of the state belief: at any time, the state 
estimate $\hat{\mathcal{X}}$ and covariance $\mathbf{P}$ 
represent a multivariate normal distribution 
$\mathcal{N}(\hat{\mathcal{X}}, 
\mathbf{P})$. This approximation is what justifies the 
use of chi-squared tests on innovations 
(Section~\ref{sec:statistics}). \\

\noindent The continuous-discrete formulation 
\citep[Table~3.9]{crassidis2004optimal} 
uses continuous-time dynamics with discrete-time 
measurements. We extend the Euclidean EKF formulation to a Composite Manifold state following \citep[Pages~1-12]{sola2021microlietheorystate}, \citep{huai2023quickguideiteratedextended}, \citep[Section~3]{Bourmaud} and \citep[Section~5]{im2024noteskalmanfilterkf}\footnote{We have not found this composite-manifold continuous-discrete formulation stated as such in the literature. We posit it and verify that it reduces to the formulations we cite; it is an Ansatz, not a proved result.}. We attempt to keep the notation similar to \citep{crassidis2004optimal} as this is what most are familiar with:
\begin{gather}
    \dot{\mathcal{X}}(t) = f( \mathcal{X}(t), \mathbf{u}(t), \mathbf{w}(t)) \\
    \tilde{\mathbf{y}}(t_k) = h(\mathcal{X}(t_k), \mathbf{v}(t_k)) 
\end{gather}
where:
\begin{gather*}
    f,\; h \quad \text{Continuously differentiable} \\
    f \quad \text{System model}, \quad 
    h \quad \text{Measurement model} \\
    \mathcal{X}(t) \quad \text{True State}, \quad 
    \mathbf{u}(t) \quad \text{Control input} \\
    \mathbf{w}(t) \sim \mathcal{N}(\mathbf{0}, 
    \mathcal{Q}(t)), \quad 
    \mathbf{v}(t_k) \sim \mathcal{N}(\mathbf{0}, 
    \mathbf{R}(t_k)) \quad 
    \text{Uncorrelated noise} \\
\delta\mathbf{x} = \mathcal{X} \,\hat{\ominus}\, 
\hat{\mathcal{X}}
    \quad \text{Estimation error}
\end{gather*}

\subsection{Filter Equations}\label{sec:ekf_eqs}

\paragraph{Initialise.}
\begin{equation}
       \hat{\mathcal{X}}(t_0) = \hat{\mathcal{X}}_0, 
    \qquad
    \mathbf{P}_0 = \mathrm{E}\!\left[
    (\mathcal{X}_0 \,\hat{\ominus}\, 
    \hat{\mathcal{X}}_0)\,
    (\mathcal{X}_0 \,\hat{\ominus}\, 
    \hat{\mathcal{X}}_0)^\top\right]
\end{equation}

\paragraph{Gain.}
\begin{gather}
        \mathbf{K}_k = \mathbf{P}_k^-\,\mathbf{H}_k^\top
\left[\mathbf{H}_k\,\mathbf{P}_k^-\,
    \mathbf{H}_k^\top +  \mathbf{H}_\mathbf{v}\mathbf{R}(t_k)\mathbf{H}_\mathbf{v}^\top\right]^{-1} \\
    \mathbf{H}_k 
    := \frac{D h}{D \mathcal{X}}
\bigg|_{\hat{\mathcal{X}}_k^-,\; \mathbf{v}(t_k)=\mathbf{0}}, \quad 
    \mathbf{H}_\mathbf{v} = \frac{\partial h}{\partial \mathbf{v}}\bigg|_{\hat{\mathcal{X}}_k^-,\; \mathbf{v}(t_k)=\mathbf{0}} \
\end{gather}

\paragraph{Update.}
\begin{gather}
    \hat{\mathcal{X}}_k^+ = \hat{\mathcal{X}}_k^- \  
    \hat{\oplus} \ \mathbf{K}_k\left[\tilde{\mathbf{y}}_k 
     \ \hat{\ominus} \  h(\hat{\mathcal{X}}_k^- , \textbf{0})\right] \\
    \mathbf{P}_k^+ = \left[\mathbf{I} 
    -\mathbf{K}_k\,\mathbf{H}_k
    \right]\mathbf{P}_k^-
\end{gather}

\paragraph{Propagation (continuous form).}
\begin{gather}
\dot{\hat{\mathcal{X}}}(t) = f(\hat{\mathcal{X}}(t), 
    \mathbf{u}(t), \textbf{0}) \\
    \dot{\mathbf{P}}(t) = F(t)\,\mathbf{P}(t) 
    + \mathbf{P}(t)\,F^\top(t) 
    + G(t)\,\mathcal{Q}(t)\,G^\top(t) \\
    F(t) := \frac{D f}{D \mathcal{X}}
\bigg|_{\hat{\mathcal{X}}(t),\;\mathbf{u}(t),\; \mathbf{w} = \mathbf{0}}, \quad G(t) = \frac{\partial f}{\partial \mathbf{w}} \bigg|_{\hat{\mathcal{X}}(t),\;\mathbf{u}(t),\; \mathbf{w} = \mathbf{0}} \quad 
\end{gather}

\paragraph{Propagation (Discrete form).}
First-order Forward Euler integration of the estimate,
component-wise according to the system model, with discrete-time process
noise covariance approximated following 
\citep[Equations 3.179, 3.193]{crassidis2004optimal}.

\begin{gather}    \hat{\mathcal{X}}_{k+1}^- = 
\hat{\mathcal{X}}_{k}^+ \,\hat{\oplus}\, \Delta t_k \cdot f(\hat{\mathcal{X}}_k^+,\, 
    \mathbf{u}_k, \mathbf{0})\\
    \Phi_k = \mathbf{I} + \Delta t_k \cdot F_k \\
    \mathbf{P}_{k+1}^- = \Phi_k\,\mathbf{P}_k^+\,
    \Phi_k^\top 
    + \Delta t_k\,G_k\,\mathcal{Q}_k\,G_k^\top
\end{gather}
where $\Delta t_k = t_{k+1} - t_{k}$ is the actual frame 
interval (not assumed constant). \\

 \noindent Note, state propagation being a left or right perturbation depends on the dynamics. Our equation is therefore strictly only right if $f$ is defined as the right acting dynamics, as whether it is a right or left perturbation is dictated by the dynamics and is a per sub-state decision. In our case we'll see the pose dynamics are left acting and use the left perturbation. You can map a left perturbation to a right perturbation using the adjoint. Thereby restating all perturbations as right acting. We don't restate pose propagation as right acting because it costs FLOPS for no benefit. \\

 \noindent The state vector has variable length: feature points 
enter and leave the filter as they are detected, 
lost, or reclassified. Three operations are needed 
\citep{Mourikis2007AMC}.

\paragraph{Adding a state (augmentation).}
\label{sec:augment}

When a new feature with estimated position 
$\hat{\mathbf{p}}_{\text{new}}^B$ is added, the state 
and covariance grow:
\begin{equation}
    \hat{\mathcal{X}} \leftarrow 
    \langle\,\hat{\mathcal{X}},\; 
    \hat{\mathbf{p}}_{\text{new}}^B\,\rangle, \qquad
    \mathbf{P} \leftarrow \begin{pmatrix} 
    \mathbf{P} & 
    \mathbf{P}\,\mathbf{J}_{\text{aug}}^\top \\
    \mathbf{J}_{\text{aug}}\,\mathbf{P} & 
    \boldsymbol{\Sigma}_{\text{new}}
    \end{pmatrix}
\end{equation}
where $\mathbf{J}_{\text{aug}}$ captures how the new 
point's position depends on the current error state, 
and $\boldsymbol{\Sigma}_{\text{new}}$ is the full 
covariance of the point including measurement noise 
(Section~\ref{sec:covariance_init}). The blocks 
$\mathbf{P}\,\mathbf{J}_{\text{aug}}^\top$, 
$\mathbf{J}_{\text{aug}}\,\mathbf{P}$ encode the 
initial cross-correlations between the new point and 
all existing states. The same distinction as for $F$ 
and $H$ applies: the $\mathbb{R}^3$ columns of 
$\mathbf{J}_{\text{aug}}$ are standard partial 
derivatives; the $SE(3)$ column uses the right 
Jacobian \citep[Eq.~41a]{sola2021microlietheorystate}. \\

\noindent The position of augmented feature points depends on the estimated pose change, which itself 
depends on the filter state. Tracing 
$\mathbf{J}_{\text{aug}}$ through the full chain 
(state $\to$ pose $\to$ pose change $\to$ solver $\to$ reconstruction) is complex. We set 
$\mathbf{J}_{\text{aug}} = \delta\,\mathbf{1}$ with a small $\delta > 0$ to seed nonzero 
cross-correlations, avoiding the transient of 
recovering from exactly zero correlation\footnote{$\mathbf{I}$ is the identity. $\mathbf{1}$ is the matrix of only ones.}.

\paragraph{Removing a state (marginalisation).}
When a feature is removed (lost, reclassified, or 
retired), delete the corresponding 3 rows and 3 
columns from both $\hat{\mathcal{X}}$ and 
$\mathbf{P}$. No information about the remaining 
states is lost; their estimates and mutual 
cross-correlations are preserved.

\paragraph{Visibility transition 
(stereo $\leftrightarrow$ mono).}
A stereo feature demoted to mono, or a mono feature 
promoted to stereo, does not change its state 
representation. $\mathbf{p}_i^B \in \mathbb{R}^3$ 
is the same regardless of visibility class. The 
transition affects only:
\begin{itemize}
    \item Which cameras produce measurements for 
          this point (the measurement model row 
          count changes).
    \item The group label ($S$, $L$, $R$) used for 
          bookkeeping.
\end{itemize}
No augmentation or marginalisation is needed for 
visibility transitions.

\subsection{System Model}
\label{sec:system_model}

We derive the system model 
$f(\mathcal{X}, \mathbf{u}, \mathbf{w})$ for the 
continuous-discrete filter 
(Section~\ref{sec:ekf}), followed by the process 
noise covariance $\mathcal{Q}$, the error dynamics 
Jacobian $F$, and the noise Jacobian $G$.

\subsubsection{State Vector}

The composite manifold state 
(Section~\ref{sec:composite}) is:
\begin{equation}\label{eq:state}
    \mathcal{X} = \left\langle\; T,\;
    \mathbf{v},\; \boldsymbol{\omega},\;
    \mathbf{g}^B,\; \mathbf{d}^B,\;
    \mathbf{p}_1^B,\; \ldots,\;
    \mathbf{p}_{N_f}^B \;\right\rangle
\end{equation}
where $T = T_{B_t,B_0} \in SE(3)$ is the body pose 
(Section~\ref{sec:lie_algebra}) and all other 
components are in $\mathbb{R}^3$. The pose acts as 
$\mathbf{p}^{B_t} = \mathbf{R}_{B_t}\,
\mathbf{p}^{B_0} + \mathbf{t}_{B_t}$, where 
$\mathbf{t}_{B_t}$ is the $B_0$ origin expressed 
in the body frame, a world-fixed point, identical 
in nature to the feature positions 
$\mathbf{p}_i^B$. \\

\noindent The composite perturbation vector 
(Section~\ref{sec:composite}, 
Equation~\ref{eq:composite_delta}) is:
\begin{equation}\label{eq:delta_x}
    \delta\mathbf{x} = \begin{pmatrix}
    \delta\boldsymbol{\xi} \\[1ex] 
    \delta\mathbf{v} \\[1ex] 
    \delta\boldsymbol{\omega} \\[1ex] 
    \delta\mathbf{g}^B \\[1ex] 
    \delta\mathbf{d}^B \\[1ex] 
    \delta\mathbf{p}_1^B \\[1ex] 
    \vdots \\[1ex] 
    \delta\mathbf{p}_{N_f}^B
    \end{pmatrix} 
    \in \mathbb{R}^{18+3N_f}
\end{equation}
with $\delta\boldsymbol{\xi} 
= (\delta\boldsymbol{\rho},\,
\delta\boldsymbol{\phi}) \in \mathbb{R}^6$ the 
$SE(3)$ tangent vector and 
$N_f = N_S + N_L + N_R$. The $\mathbb{R}^3$ 
components are:
\begin{align}
    \mathbf{v} &\in \mathbb{R}^3 
        & &\text{Linear velocity in body frame} \\
    \boldsymbol{\omega} &\in \mathbb{R}^3 
        & &\text{Angular velocity in body frame} \\
    \mathbf{g}^B &\in \mathbb{R}^3 
        & &\text{Gravity in body frame 
              ($\|\mathbf{g}^B\| \approx g$)} \\
    \mathbf{d}^B &\in \mathbb{R}^3 
        & &\text{Wind/disturbances in body frame}
\end{align}
Feature positions partitioned by visibility:
\begin{align}
    \mathbf{p}_j^S &\in \mathbb{R}^3,\; 
    j = 1,\ldots,N_S
        & &\text{Stereo features} \\
    \mathbf{p}_j^L &\in \mathbb{R}^3,\; 
    j = 1,\ldots,N_L
        & &\text{Mono-left features} \\
    \mathbf{p}_j^R &\in \mathbb{R}^3,\; 
    j = 1,\ldots,N_R
        & &\text{Mono-right features}
\end{align}
All positions are in body frame. The visibility 
label ($S$, $L$, $R$) determines which 
measurements a point contributes 
(Section~\ref{sec:measurement}). \\

\noindent Gravity is split from disturbances: 
gravity is well approximated as constant in the 
world frame with known magnitude, requiring only 
small process noise. Wind is genuinely stochastic, 
requiring larger process noise. Separate tuning 
avoids compromising one to accommodate the other.

\subsubsection{Input Vector}
\begin{equation}
    \mathbf{u} = \begin{pmatrix} T_1 & T_2 & T_3 
    & T_4 \end{pmatrix}^\top \in \mathbb{R}^4
\end{equation}
where $T_i = k_f\,\omega_{r,i}^2$ is the thrust 
from rotor $i$.

\subsubsection{Dynamics}

\paragraph{Block 0: Pose ($\dot{T}$).}
The body-frame velocity twist 
$\boldsymbol{\xi}_v = ({\mathbf{v}},\,
{\boldsymbol{\omega}})$ drives the pose. The 
continuous dynamics are:
\begin{equation}\label{eq:Tdot}
    \boxed{
    \dot{T} = -\boldsymbol{\xi}_v^\wedge\,T
    }
\end{equation}
\textit{Derivation.} Write 
$T = [\mathbf{R},\,\mathbf{t};\,
\mathbf{0}^\top,\,1]$ and compute the right-hand 
side using the Lie algebra element 
$\boldsymbol{\xi}_v^\wedge 
= [[{\boldsymbol{\omega}}]_\times,\,
{\mathbf{v}};\,\mathbf{0}^\top,\,0]$ 
(Section~\ref{sec:lie_algebra}):
\begin{equation}
    -\boldsymbol{\xi}_v^\wedge\,T 
    = -\begin{pmatrix}
    [{\boldsymbol{\omega}}]_\times & 
    {\mathbf{v}} \\ 
    \mathbf{0}^\top & 0
    \end{pmatrix}
    \begin{pmatrix}
    \mathbf{R} & \mathbf{t} \\ 
    \mathbf{0}^\top & 1
    \end{pmatrix}
    = \begin{pmatrix}
    -[{\boldsymbol{\omega}}]_\times\mathbf{R} 
    & -[{\boldsymbol{\omega}}]_\times\mathbf{t} 
    - {\mathbf{v}} \\ 
    \mathbf{0}^\top & 0
    \end{pmatrix}
\end{equation}
Reading off the blocks: 
$\dot{\mathbf{R}} 
= -[{\boldsymbol{\omega}}]_\times\mathbf{R}$, 
which is the rotation kinematics 
(Section~\ref{sec:dynamics}), and 
$\dot{\mathbf{t}} = -{\mathbf{v}} 
- [{\boldsymbol{\omega}}]_\times\mathbf{t} 
= -{\mathbf{v}} 
- {\boldsymbol{\omega}}\times\mathbf{t}$, 
which is the transport theorem 
(Equation~\ref{eq:transport_general}) applied to 
$\mathbf{t}_{B_t}$, the $B_0$ origin expressed 
in body frame, a world-fixed point. Both known 
component dynamics are recovered. \\

\noindent No process noise enters the pose 
directly: pose uncertainty is inherited entirely 
from velocity and angular velocity through 
integration. \\

\noindent Propagation (left multiply, see 
Section~\ref{sec:ekf}):
\begin{equation}\label{eq:T_euler}
    \hat{T}_{k+1}^- = \mathrm{Exp}(
    -\hat{\boldsymbol{\xi}}_{v,k}\,\Delta t_k)
    \cdot\hat{T}_k^+
\end{equation}
This follows from the ODE $\dot{T} = A\,T$ with 
constant $A = -\boldsymbol{\xi}_v^\wedge$, whose 
solution is $T(t+\Delta t) 
= \exp(A\Delta t)\,T(t) 
= \mathrm{Exp}(-\boldsymbol{\xi}_v\Delta t)
\cdot T(t)$.

\paragraph{Block 1: Translational 
($\dot{\mathbf{v}}$).}
Newton's second law in the rotating body frame:
\begin{equation}\label{eq:vdot}
    \boxed{
    \dot{\mathbf{v}} = \frac{1}{m}
    \mathbf{F}_{\text{thrust}}(\mathbf{u} 
    + \mathbf{w}_u) 
    - \boldsymbol{\omega} \times \mathbf{v} 
    - \frac{C_d}{m}\mathbf{v} 
    + \mathbf{g}^B + \mathbf{d}^B
    }
\end{equation}
with $\mathbf{F}_{\text{thrust}} 
= (0, 0, -\sum (T_i + w_{u,i}))^\top$ in FRD. 
Actuator noise $\mathbf{w}_u$ enters through the 
thrust mapping $B_\mathbf{v}$ 
(Equation~\ref{eq:B_v}).

\paragraph{Block 2: Rotational 
($\dot{\boldsymbol{\omega}}$).}
Euler's rotation equations, assuming the inertia 
tensor is diagonal $\mathbf{J} 
= \operatorname{diag}(J_{xx}, J_{yy}, J_{zz})$ 
(valid for a symmetric X-configuration where the 
body axes align with the principal axes of 
inertia):
\begin{equation}\label{eq:omegadot}
    \boxed{
    \dot{\boldsymbol{\omega}} = \mathbf{J}^{-1}
    \left(\boldsymbol{\tau}(\mathbf{u} 
    + \mathbf{w}_u) 
    - \boldsymbol{\omega} \times 
    (\mathbf{J}\boldsymbol{\omega})\right) 
    + \mathbf{w}_\omega
    }
\end{equation}
Actuator noise $\mathbf{w}_u$ enters through the 
torque mapping $B_\omega$ 
(Equation~\ref{eq:B_omega}). The additional 
process noise $\mathbf{w}_\omega$ accounts for 
unmodelled aerodynamic torques (e.g.\ asymmetric 
wind loading) that are not captured by the 
actuator channel. \\

\noindent Torques from the 
X-configuration mixing matrix (Noise adds to each $T_i$, omitted here for clarity. $l$ is the arm length, and we have assumed the body frame aligns with the centre of mass.):
\begin{align}
    \tau_1 &= \tfrac{l}{\sqrt{2}}
    (T_4 + T_3 - T_1 - T_2) 
        & &\text{Roll} \\
    \tau_2 &= \tfrac{l}{\sqrt{2}}
    (T_1 + T_4 - T_2 - T_3) 
        & &\text{Pitch} \\
    \tau_3 &= \tfrac{k_m}{k_f}
    (T_2 + T_4 - T_1 - T_3) 
        & &\text{Yaw}
\end{align}

\paragraph{Block 3: Gravity ($\dot{\mathbf{g}}^B$).}
From the transport theorem for world-fixed vectors
(Equation~\ref{eq:transport_wfv}):
\begin{equation}\label{eq:gdot}
    \boxed{
    \dot{\mathbf{g}}^B = -\boldsymbol{\omega} 
    \times \mathbf{g}^B + \mathbf{w}_g
    }
\end{equation}
Process noise $\mathbf{w}_g$ accounts for the error in the
approximation that gravity is constant in the 
world frame (small spectral density 
$\mathcal{Q}_g$).

\paragraph{Block 4: Disturbances 
($\dot{\mathbf{d}}^B$).}
Same transport equation:
\begin{equation}\label{eq:ddot}
    \boxed{
    \dot{\mathbf{d}}^B = -\boldsymbol{\omega} 
    \times \mathbf{d}^B + \mathbf{w}_d
    }
\end{equation}
Process noise $\mathbf{w}_d$ drives the 
disturbance as a random walk in the world frame 
(larger spectral density $\mathcal{Q}_d$).

\paragraph{Block 5: Feature positions 
($\dot{\mathbf{p}}_i^B$).}
A world-stationary point in the body frame 
satisfies (derived from 
$\mathbf{p}_i^B 
= \mathbf{R}_{B,B_0}(\mathbf{p}_i^{B_0} 
- \mathbf{t}_{B_0})$ using 
$\dot{\mathbf{p}}_i^{B_0} = \mathbf{0}$ and the transport theorem, Equation~\ref{eq:transport_general}):
\begin{equation}\label{eq:pdot}
    \boxed{
    \dot{\mathbf{p}}_i^B = -\mathbf{v} 
    - \boldsymbol{\omega} \times \mathbf{p}_i^B
    }
\end{equation}
Identical for all visibility groups. Points are 
assumed stationary in the world frame; violations 
of this assumption are detected by the 
chi-squared test 
(Section~\ref{sec:statistics}) and the point is 
removed. Note that the translation component 
$\mathbf{t}_{B_t}$ of the pose satisfies the 
same equation (Block~0 derivation above).

\subsubsection{Process Noise}

The process noise vector is:
\begin{equation}\label{eq:w_vec}
    \mathbf{w} = \begin{pmatrix}
    \mathbf{w}_u \\[1ex]
    \mathbf{w}_\omega \\[1ex] 
    \mathbf{w}_g \\[1ex] 
    \mathbf{w}_d
    \end{pmatrix} 
    \in \mathbb{R}^{13}
\end{equation}
where $\mathbf{w}_u \in \mathbb{R}^4$ is actuator 
noise on the rotor thrusts, 
$\mathbf{w}_\omega \in \mathbb{R}^3$ is 
unmodelled aerodynamic torque, and the remaining 
terms are as before. The continuous-time spectral 
density (Section~\ref{sec:ekf}):
\begin{equation}\label{eq:Q}
    \mathcal{Q} = \operatorname{diag}\!\left(
    \mathcal{Q}_u,\;
    \mathcal{Q}_\omega,\;
    \mathcal{Q}_g,\;
    \mathcal{Q}_d
    \right) 
    \in \mathbb{R}^{13\times13}
\end{equation}
where $\mathcal{Q}_u \in \mathbb{R}^{4\times4}$ 
is the actuator spectral density and 
$\mathcal{Q}_\omega$, $\mathcal{Q}_g$, 
$\mathcal{Q}_d$ are each 
$3\times 3$ (units: state$^2$/time). Actuator 
noise enters both the velocity and angular 
velocity through the thrust and torque mappings. 
No noise enters the pose ($T$). Nor does any enter the linear 
velocity ($\mathbf{v}$) directly beyond the 
actuator channel: pose uncertainty is inherited 
from $\mathbf{v}$ and $\boldsymbol{\omega}$ 
through integration, and velocity uncertainty 
is inherited from the forcing terms 
$\mathbf{g}^B$, $\mathbf{d}^B$, 
$\boldsymbol{\omega}$, and the actuators through 
the dynamics (Equation~\ref{eq:vdot}).
\subsubsection{Complete System Model}

The $SE(3)$ dynamics are 
$\dot{T} = -\boldsymbol{\xi}_v^\wedge T$ 
(Equation~\ref{eq:Tdot}). The $\mathbb{R}^3$ 
dynamics are assembled as:
\begin{equation}\label{eq:f}
    \dot{\mathbf{x}}_{\mathbb{R}} 
    = f_{\mathbb{R}}(\mathcal{X}, \mathbf{u}, 
    \mathbf{w}) 
    = \begin{pmatrix}
        \frac{1}{m}\mathbf{F}_{\text{thrust}}
        (\mathbf{u} + \mathbf{w}_u) 
        - \boldsymbol{\omega} \times \mathbf{v} 
        - \frac{C_d}{m}\mathbf{v} 
        + \mathbf{g}^B + \mathbf{d}^B \\[6pt]
        \mathbf{J}^{-1}(\boldsymbol{\tau}
        (\mathbf{u} + \mathbf{w}_u) 
        - \boldsymbol{\omega} \times 
        (\mathbf{J}\boldsymbol{\omega})) 
        + \mathbf{w}_\omega \\[6pt]
        -\boldsymbol{\omega} \times \mathbf{g}^B 
        + \mathbf{w}_g \\[4pt]
        -\boldsymbol{\omega} \times \mathbf{d}^B 
        + \mathbf{w}_d \\[6pt]
        -\mathbf{v} - \boldsymbol{\omega} 
        \times \mathbf{p}_1^B \\[1ex]
        \vdots \\[1ex]
        -\mathbf{v} - \boldsymbol{\omega} 
        \times \mathbf{p}_{N_f}^B
    \end{pmatrix}
\end{equation}
Together with $\dot{T}$ these form the complete 
system model $f(\mathcal{X}, \mathbf{u}, 
\mathbf{w})$. Setting $\mathbf{w} = \mathbf{0}$ 
recovers the nominal dynamics used for state 
propagation (Section~\ref{sec:ekf}).

\subsubsection{Noise Jacobian $G$}
\label{sec:G_matrix}

The noise Jacobian $G = \partial f / 
\partial\mathbf{w}$ is the standard partial 
derivative (Section~\ref{sec:composite}) since 
$\mathbf{w} \in \mathbb{R}^{13}$ is 
Euclidean. Define the actuator-to-force and 
actuator-to-torque mappings:
\begin{equation}\label{eq:B_v}
    B_\mathbf{v} = \frac{1}{m}\,
    \frac{\partial\mathbf{F}_{\mathrm{thrust}}}
    {\partial\mathbf{u}} 
    = \frac{1}{m}\begin{pmatrix}
    0 & 0 & 0 & 0 \\[1ex]
    0 & 0 & 0 & 0 \\[1ex]
    -1 & -1 & -1 & -1
    \end{pmatrix}
    \in \mathbb{R}^{3\times4}
\end{equation}
\begin{equation}\label{eq:B_omega}
    B_\omega = \mathbf{J}^{-1}\,
    \frac{\partial\boldsymbol{\tau}}
    {\partial\mathbf{u}}
    = \mathbf{J}^{-1}
    \begin{pmatrix}
    -\frac{l}{\sqrt{2}} & -\frac{l}{\sqrt{2}} 
    & \frac{l}{\sqrt{2}} & \frac{l}{\sqrt{2}} 
    \\[1ex]
    \frac{l}{\sqrt{2}} & -\frac{l}{\sqrt{2}} 
    & -\frac{l}{\sqrt{2}} & \frac{l}{\sqrt{2}} 
    \\[1ex]
    -\frac{k_m}{k_f} & \frac{k_m}{k_f} 
    & -\frac{k_m}{k_f} & \frac{k_m}{k_f}
    \end{pmatrix}
    \in \mathbb{R}^{3\times4}
\end{equation}
The full $G$ matrix maps the 
$(13)$-dimensional noise into the 
$(18+3N_f)$-dimensional perturbation space. 
Column groups correspond to 
$\mathbf{w}_u$, $\mathbf{w}_\omega$, 
$\mathbf{w}_g$, $\mathbf{w}_d$:
\begin{equation}\label{eq:G}
    G = \begin{pmatrix}
    \mathbf{0}_{6\times4} & \mathbf{0}_{6\times3} 
    & \mathbf{0}_{6\times3} & \mathbf{0}_{6\times3} 
    \\[1ex]
    B_\mathbf{v} & \mathbf{0}_{3\times3} 
    & \mathbf{0} & \mathbf{0} \\[1ex]
    B_\omega & \mathbf{I}_3 
    & \mathbf{0} & \mathbf{0} \\[1ex]
    \mathbf{0}_{3\times4} & \mathbf{0} 
    & \mathbf{I}_3 & \mathbf{0} \\[1ex]
    \mathbf{0}_{3\times4} & \mathbf{0} 
    & \mathbf{0} & \mathbf{I}_3 \\[1ex]
    \mathbf{0}_{3\times4} & \mathbf{0} 
    & \mathbf{0} & \mathbf{0} \\[1ex]
    \vdots & & & \vdots \\[1ex]
    \mathbf{0}_{3\times4} & \mathbf{0} 
    & \mathbf{0} & \mathbf{0}
    \end{pmatrix}
    \in \mathbb{R}^{(18+3N_f)\times 13}
\end{equation}
Row groups correspond to 
$\delta\boldsymbol{\xi}$, $\delta\mathbf{v}$, 
$\delta\boldsymbol{\omega}$, 
$\delta\mathbf{g}^B$, $\delta\mathbf{d}^B$ , $\delta\mathbf{p}_i^B$. The pose row is zero 
(no direct noise). The velocity row receives 
actuator noise through $B_\mathbf{v}$. The 
angular velocity row receives both actuator 
noise through $B_\omega$ (structured, via the 
mixing matrix) and unmodelled torque 
$\mathbf{w}_\omega$ (unstructured).

\subsubsection{Error Dynamics Jacobian $F$}
\label{sec:F_matrix}

$F$ is the composite Jacobian 
(Section~\ref{sec:composite}) of the error 
dynamics 
$\dot{\delta\mathbf{x}} \approx 
F\,\delta\mathbf{x}$, evaluated at 
$\mathbf{w} = \mathbf{0}$. The noise terms do 
not affect $F$: they enter linearly through $G$ 
and do not couple to the state.

\paragraph{$\mathbb{R}^3$ blocks.}
For the $\mathbb{R}^3$ states, $F$ is obtained 
by standard differentiation of 
$f_{\mathbb{R}}$ with $\mathbf{w} = \mathbf{0}$ 
\citep[Eqs.~3.239--3.244]{crassidis2004optimal}:
\begin{align}
    F_{vv} &= -[\hat{\boldsymbol{\omega}}]_\times 
    - \tfrac{C_d}{m}I_3 
        & F_{v\omega} &= 
        [\hat{\mathbf{v}}]_\times \\[1ex]
    F_{\omega\omega} &= \mathbf{J}^{-1}(
    [\mathbf{J}\hat{\boldsymbol{\omega}}]_\times 
    - [\hat{\boldsymbol{\omega}}]_\times
    \mathbf{J})
        & & \\[1ex]
    F_{g\omega} &= [\hat{\mathbf{g}}^B]_\times 
        & F_{gg} &= 
        -[\hat{\boldsymbol{\omega}}]_\times \\[1ex]
    F_{d\omega} &= [\hat{\mathbf{d}}^B]_\times 
        & F_{dd} &= 
        -[\hat{\boldsymbol{\omega}}]_\times \\[1ex]
    F_{pv} &= -I_3 
        & F_{p_i\omega} &= 
        [\hat{\mathbf{p}}_i^B]_\times \\[1ex]
    F_{p_i p_i} &= 
    -[\hat{\boldsymbol{\omega}}]_\times 
        & &
\end{align}
Cross-point terms vanish: 
$\partial\dot{\mathbf{p}}_i^B 
/ \partial\mathbf{p}_j^B 
= \mathbf{0}$ for $j \neq i$.

\paragraph{$SE(3)$ blocks.}
For the pose, the error dynamics are 
$\dot{\delta\boldsymbol{\xi}} 
= -\mathrm{Ad}_{\hat{T}^{-1}}
(\delta\mathbf{v},\,
\delta\boldsymbol{\omega})^\top$. \\

\noindent \textit{Derivation.} The true pose 
satisfies 
$\dot{T}_{\mathrm{true}} 
= -(\boldsymbol{\xi}_v 
+ \delta\boldsymbol{\xi}_v)^\wedge\,
T_{\mathrm{true}}$ where 
$\delta\boldsymbol{\xi}_v 
= (\delta\mathbf{v},\,
\delta\boldsymbol{\omega})$. With 
$T_{\mathrm{true}} 
= \hat{T}\cdot\mathrm{Exp}(
\delta\boldsymbol{\xi})$ 
\citep[Eq.~25]{sola2021microlietheorystate}, 
expand the left side via the product rule:
\begin{equation}
    \dot{\hat{T}}\,\mathrm{Exp}(
    \delta\boldsymbol{\xi}) 
    + \hat{T}\,\tfrac{d}{dt}\mathrm{Exp}(
    \delta\boldsymbol{\xi}) 
    = -(\boldsymbol{\xi}_v 
    + \delta\boldsymbol{\xi}_v)^\wedge\,
    \hat{T}\,\mathrm{Exp}(
    \delta\boldsymbol{\xi})
\end{equation}
The estimate dynamics 
$\dot{\hat{T}} 
= -\boldsymbol{\xi}_v^\wedge\hat{T}$ 
(Equation~\ref{eq:Tdot}) cancel from both sides:
\begin{equation}
    \hat{T}\,\tfrac{d}{dt}\mathrm{Exp}(
    \delta\boldsymbol{\xi}) 
    = -(\delta\boldsymbol{\xi}_v)^\wedge\,
    \hat{T}\,\mathrm{Exp}(
    \delta\boldsymbol{\xi})
\end{equation}
Left-multiply by $\hat{T}^{-1}$ and apply the 
adjoint conjugation property 
$\mathcal{X}^{-1}\boldsymbol{\tau}^\wedge
\mathcal{X} 
= (\mathrm{Ad}_{\mathcal{X}^{-1}}
\boldsymbol{\tau})^\wedge$ 
\citep[Eq.~29]{sola2021microlietheorystate}:
\begin{equation}
    \tfrac{d}{dt}\mathrm{Exp}(
    \delta\boldsymbol{\xi}) 
    = -(\mathrm{Ad}_{\hat{T}^{-1}}\,
    \delta\boldsymbol{\xi}_v)^\wedge\,
    \mathrm{Exp}(\delta\boldsymbol{\xi})
\end{equation}
First order: 
$\mathrm{Exp}(\delta\boldsymbol{\xi}) 
\approx \mathcal{E} 
+ \delta\boldsymbol{\xi}^\wedge$ 
\citep[Eq.~16, truncated after the first term]{sola2021microlietheorystate}, so 
$\frac{d}{dt}\mathrm{Exp}(
\delta\boldsymbol{\xi}) 
\approx (\dot{\delta\boldsymbol{\xi}})^\wedge$ 
and 
$\mathrm{Exp}(\delta\boldsymbol{\xi}) 
\approx \mathcal{E}$ on the right. Extracting 
the vector:
\begin{equation}\label{eq:se3_error_dyn}
    \dot{\delta\boldsymbol{\xi}} 
    = -\mathrm{Ad}_{\hat{T}^{-1}}\,
    \delta\boldsymbol{\xi}_v 
    = -\mathrm{Ad}_{\hat{T}^{-1}}
    \begin{pmatrix}\delta\mathbf{v} \\[1ex] 
    \delta\boldsymbol{\omega}\end{pmatrix}
\end{equation}
The resulting $F$ blocks are:
\begin{equation}\label{eq:F_se3}
    F_{\delta\boldsymbol{\xi},\,
    \delta\boldsymbol{\xi}} 
    = \mathbf{0}_{6\times 6}, \qquad
    F_{\delta\boldsymbol{\xi},\,
    (\delta\mathbf{v},\,
    \delta\boldsymbol{\omega})} 
    = -\mathrm{Ad}_{\hat{T}^{-1}}
\end{equation}
with $\mathrm{Ad}_{\hat{T}^{-1}}$ from 
Equation~\ref{eq:adjoint_se3} evaluated at 
$\hat{T}^{-1}$. All other entries in the 
$\delta\boldsymbol{\xi}$ row are zero 
($\delta\mathbf{g}$, $\delta\mathbf{d}$, 
$\delta\mathbf{p}_i$ do not enter the pose 
error dynamics). All entries in the 
$\delta\boldsymbol{\xi}$ column from existing 
rows are zero (no $\mathbb{R}^3$ dynamics 
depend on pose).

\paragraph{Full $F$ matrix.}
Column order: $\delta\boldsymbol{\xi},\;
\delta\mathbf{v},\;
\delta\boldsymbol{\omega},\;
\delta\mathbf{g}^B,\;\delta\mathbf{d}^B,\;
\delta\mathbf{p}_1^B,\;\ldots,\;
\delta\mathbf{p}_{N_f}^B$:
\begin{equation}\label{eq:F}
    F = \left(\begin{array}{c|cc|cc|ccc}
        \mathbf{0}_{6} & 
        \multicolumn{2}{c|}{
        -\mathrm{Ad}_{\hat{T}^{-1}}}
        & \mathbf{0} & \mathbf{0} 
        & \mathbf{0} & \cdots & \mathbf{0} 
        \\[1ex] \hline
        \mathbf{0} & 
        F_{vv} & F_{v\omega} 
        & I_3 & I_3 
        & \mathbf{0} & \cdots & \mathbf{0} 
        \\[1ex]
        \mathbf{0} & 
        \mathbf{0} & F_{\omega\omega} 
        & \mathbf{0} & \mathbf{0} 
        & \mathbf{0} & \cdots & \mathbf{0} 
        \\[1ex] \hline
        \mathbf{0} & 
        \mathbf{0} & F_{g\omega} 
        & F_{gg} & \mathbf{0} 
        & \mathbf{0} & \cdots & \mathbf{0} 
        \\[1ex]
        \mathbf{0} & 
        \mathbf{0} & F_{d\omega} 
        & \mathbf{0} & F_{dd} 
        & \mathbf{0} & \cdots & \mathbf{0} 
        \\[1ex] \hline
        \mathbf{0} & 
        F_{pv} & F_{p_1\omega} 
        & \mathbf{0} & \mathbf{0} 
        & F_{p_1p_1} & \cdots & \mathbf{0} 
        \\[1ex]
        \vdots & 
        \vdots & \vdots 
        & & 
        & & \ddots & \\[1ex]
        \mathbf{0} & 
        F_{pv} & F_{p_{N_f}\omega} 
        & \mathbf{0} & \mathbf{0} 
        & \mathbf{0} & \cdots & F_{p_{N_f}p_{N_f}}
    \end{array}\right)
\end{equation}
The lower-right block is block-diagonal in the 
per-point states: each point $i$ has its own 
diagonal block $F_{p_ip_i} 
= -[\hat{\boldsymbol{\omega}}]_\times$ and 
off-diagonal coupling $F_{p_i\omega} 
= [\hat{\mathbf{p}}_i^B]_\times$ to angular 
velocity. All points share the same 
$F_{pv} = -I_3$. Cross-point terms vanish: 
$\partial\dot{\mathbf{p}}_i^B 
/ \partial\mathbf{p}_j^B 
= \mathbf{0}$ for $j \neq i$. The first row 
(pose error dynamics) couples to velocity and 
angular velocity through 
$-\mathrm{Ad}_{\hat{T}^{-1}}$. The first 
column is zero throughout (no $\mathbb{R}^3$ 
dynamics depend on pose).

\subsubsection{Initialisation}
\begin{align}
    \hat{T}_0 &= \begin{pmatrix} \mathbf{I} & 
    \mathbf{0} \\ \mathbf{0}^\top & 1 
    \end{pmatrix}
        & &\text{Body at $B_0$ origin, aligned} 
        \\[1ex]
    \hat{\mathbf{v}}_0 &= \mathbf{0}
        & &\text{Stationary start} \\[1ex]
    \hat{\boldsymbol{\omega}}_0 &= \mathbf{0}
        & &\text{No initial rotation} \\[1ex]
    \hat{\mathbf{g}}_0^B &= (0, 0, g)^\top
        & &\text{FRD: down $= +e_3$} \\[1ex]
    \hat{\mathbf{d}}_0^B &= -\hat{\mathbf{g}}_0^B
        & &\text{Normal force}
\end{align}
No feature points are present at initialisation. 
All points must pass the velocity entry test 
(Section~\ref{sec:statistics}) before being 
augmented into the filter. The first frame 
provides stereo and mono detections; these are 
tracked to the second frame where the joint 
solver (Section~\ref{sec:joint_solver}) produces 
position and velocity estimates with covariance, 
at which point they are augmented 
(Section~\ref{sec:augment}). Initial covariance:
\begin{equation}
    \mathbf{P}_0 = \operatorname{diag}\!\left(
    \sigma_\xi^2 I_6,\;
    \sigma_v^2 I_3,\;
    \sigma_\omega^2 I_3,\;
    \sigma_g^2 I_3,\;
    \sigma_d^2 I_3
    \right)
\end{equation}
with no feature blocks until augmentation.

\subsection{Measurement Model}
\label{sec:measurement}

We derive the measurement model 
$\tilde{\mathbf{y}}(t_k) = h(\mathcal{X}(t_k), 
\mathbf{v}(t_k))$ for the continuous-discrete 
filter (Section~\ref{sec:ekf}), followed by the 
measurement noise $\mathbf{R}$, and the Jacobians 
$\mathbf{H}_k$ and $\mathbf{H}_\mathbf{v}$.

\subsubsection{Observations}

At each frame $t_k$, the observations are 
LK-tracked pixel coordinates of feature points in 
one or both cameras:
\begin{equation}\label{eq:observations}
    \tilde{\mathbf{y}}(t_k) = \begin{pmatrix}
    \tilde{\mathbf{y}}_1 \\[1ex] 
    \vdots \\[1ex] 
    \tilde{\mathbf{y}}_{N_f}
    \end{pmatrix}
    \in \mathbb{R}^{\sum_i \nu_i}
\end{equation}
where $\tilde{\mathbf{y}}_i 
\in \mathbb{R}^{\nu_i}$ are the pixel 
coordinates of point $i$, with $\nu_i = 4$ 
(stereo: left and right cameras) or $\nu_i = 2$ 
(mono: single camera).

\subsubsection{Measurement Function}

The measurement prediction maps the state to 
pixel coordinates via direct geometric projection 
(Equation~\ref{eq:body_to_pixel}). For a single 
point $i$ in camera $c \in \{L, R\}$:
\begin{equation}\label{eq:h_single}
    h_{i,c}(\mathcal{X}, \mathbf{v}_{i,c}) 
    = \boldsymbol{\pi}(\mathbf{R}_{cB}\,
    \mathbf{p}_i^B + \mathbf{t}_c) 
    + \mathbf{v}_{i,c}
\end{equation}
where $\mathbf{v}_{i,c} \in \mathbb{R}^2$ is 
additive pixel noise in camera $c$ for point 
$i$. The noise is additive because the dominant 
error source (LK sub-pixel localisation) acts 
directly on the pixel coordinates.

\paragraph{Stereo feature ($\mathcal{F}_S$).}
A stereo feature $\mathbf{p}_i^B$ is observed 
in both cameras:
\begin{equation}\label{eq:h_stereo}
    h_{S,i}(\mathcal{X}, \mathbf{v}_i) 
    = \begin{pmatrix}
    \boldsymbol{\pi}(\mathbf{R}_{LB}\,
    \mathbf{p}_i^B + \mathbf{t}_L) \\[1ex]
    \boldsymbol{\pi}(\mathbf{R}_{RB}\,
    \mathbf{p}_i^B + \mathbf{t}_R)
    \end{pmatrix} + \mathbf{v}_i
    = \begin{pmatrix}
    u_L \\[1ex] v_L \\[1ex] u_R \\[1ex] v_R
    \end{pmatrix} 
    \in \mathbb{R}^4
\end{equation}
with $\mathbf{v}_i \in \mathbb{R}^4$.

\paragraph{Why not a separate disparity 
measurement?}
Disparity $d = u_L - u_R$ is a linear function 
of the four pixel coordinates already observed. 
Adding it as a 5th row adds no new information. A 
depth error produces different innovations in 
$u_L$ and $u_R$ because the Jacobians 
$\partial h / \partial \mathbf{p}_i^B$ differ 
between the two cameras (different extrinsics). 
The filter exploits this geometric discrepancy 
through the existing four rows.

\paragraph{Mono feature ($\mathcal{F}_M$).}
A mono feature is observed in one camera 
$c \in \{L, R\}$:
\begin{equation}\label{eq:h_mono}
    h_{M,i}(\mathcal{X}, \mathbf{v}_i) 
    = \boldsymbol{\pi}(\mathbf{R}_{cB}\,
    \mathbf{p}_i^B + \mathbf{t}_c) 
    + \mathbf{v}_i
    = \begin{pmatrix} u_c \\[1ex] v_c 
    \end{pmatrix} 
    \in \mathbb{R}^2
\end{equation}
with $\mathbf{v}_i \in \mathbb{R}^2$.

\paragraph{Assembled measurement function.}
Stacking all points:
\begin{equation}\label{eq:h_full}
    h(\mathcal{X}, \mathbf{v}) = \begin{pmatrix}
    h_{\cdot,1}(\mathcal{X}, \mathbf{v}_1) \\[1ex]
    \vdots \\[1ex]
    h_{\cdot,N_f}(\mathcal{X}, \mathbf{v}_{N_f})
    \end{pmatrix}
    \in \mathbb{R}^{\sum_i \nu_i}
\end{equation}
where each $h_{\cdot,i}$ is $h_{S,i}$ or 
$h_{M,i}$ according to the visibility of point 
$i$.

\subsubsection{Measurement Noise}

The noise vector 
$\mathbf{v} = (\mathbf{v}_1, \ldots, 
\mathbf{v}_{N_f})^\top$ has covariance:
\begin{equation}\label{eq:R}
    \mathbf{R} = \operatorname{diag}\!\left(
    \sigma_{\mathrm{px}}^2\,
    \mathbf{I}_{\nu_1},\;\ldots,\;
    \sigma_{\mathrm{px}}^2\,
    \mathbf{I}_{\nu_{N_f}}
    \right)
    = \sigma_{\mathrm{px}}^2\,
    \mathbf{I}_{\sum_i \nu_i}
\end{equation}
where $\sigma_{\mathrm{px}} \approx 0.3$--$1.0$
px is a tuning parameter representing LK
sub-pixel accuracy; the runtime value is
maintained in
\href{https://github.com/danielftg/blackbird-vio/blob/a8713f352ee5481de798098e39798ac93115d479/src/constants/algorithm.yaml}{algorithm.yaml}.
All pixel measurements share
the same noise level; the block-diagonal 
structure reflects independence between points 
and between cameras.

\subsubsection{Measurement Jacobian $\mathbf{H}_k$}

Following Section~\ref{sec:ekf}, 
$\mathbf{H}_k = \frac{Dh}{D\mathcal{X}}
\big|_{\hat{\mathcal{X}}_k^-,\,\mathbf{v}=
\mathbf{0}}$ is the composite Jacobian 
(Section~\ref{sec:composite}). Since the 
measurement noise is additive, setting 
$\mathbf{v} = \mathbf{0}$ simply removes it.

\paragraph{Projection Jacobian.}
For camera $c$, define 
$\mathbf{p}_i^c = \mathbf{R}_{cB}\,
\mathbf{p}_i^B + \mathbf{t}_c$. The chain rule 
through $\boldsymbol{\pi} = \kappa \circ D 
\circ \pi_n$ 
(Section~\ref{sec:projective}) gives:
\begin{equation}\label{eq:dpi_dpc}
    \frac{\partial\boldsymbol{\pi}}
    {\partial\mathbf{p}^c} 
    = \mathbf{K}_{2\times 2}\,\mathbf{J}_D\,
    \frac{1}{p_z^c}\begin{pmatrix} 
    1 & 0 & -x_n \\[1ex] 
    0 & 1 & -y_n \end{pmatrix}
\end{equation}
\begin{equation}
    \mathbf{K}_{2\times 2} = \begin{pmatrix}
        f_u & 0 \\[1ex]
        0   & f_v
    \end{pmatrix}, \quad 
    \mathbf{J}_D = \frac{\partial D}
    {\partial (x_n, y_n)}
\end{equation}
The Jacobian with respect to the body-frame 
position:
\begin{equation}\label{eq:H_pi}
    \frac{\partial\boldsymbol{\pi}}
    {\partial\mathbf{p}_i^B}\bigg|_c
    = \frac{\partial\boldsymbol{\pi}}
    {\partial\mathbf{p}^c}\,\mathbf{R}_{cB}
\end{equation}

\paragraph{Per-point row.}
The measurement depends only on 
$\mathbf{p}_i^B$ — not on $T$, $\mathbf{v}$, 
$\boldsymbol{\omega}$, $\mathbf{g}^B$, 
$\mathbf{d}^B$, or any other point 
$\mathbf{p}_j^B$ ($j \neq i$). The full 
Jacobian row for point $i$ in camera $c$:
\begin{equation}\label{eq:H_row}
    \mathbf{H}_i^c = \begin{pmatrix}
    \mathbf{0}_{2\times 6} & 
    \mathbf{0}_{2\times 3} & 
    \mathbf{0}_{2\times 3} & 
    \mathbf{0}_{2\times 3} & 
    \mathbf{0}_{2\times 3} & 
    \cdots & 
    \frac{\partial\boldsymbol{\pi}}
    {\partial\mathbf{p}_i^B}\big|_c & 
    \cdots & \mathbf{0}_{2\times 3}
    \end{pmatrix}
\end{equation}
Nonzero only in the columns corresponding to 
$\mathbf{p}_i^B$. For a stereo feature, stack 
$\mathbf{H}_i^L$ and $\mathbf{H}_i^R$ 
(4 rows). For a mono feature, a single 
$\mathbf{H}_i^c$ (2 rows).

\paragraph{Assembled $\mathbf{H}_k$.}
Stacking all points:
\begin{equation}\label{eq:H_full}
    \mathbf{H}_k = \begin{pmatrix}
    \mathbf{H}_1 \\[1ex] 
    \vdots \\[1ex] 
    \mathbf{H}_{N_f}
    \end{pmatrix}
    \in \mathbb{R}^{(\sum_i \nu_i) 
    \times (18+3N_f)}
\end{equation}
The matrix is block-sparse: each point 
contributes a nonzero block only in its own 
columns.

\subsubsection{Noise Jacobian $\mathbf{H}_\mathbf{v}$}

Since the noise is additive 
(Equation~\ref{eq:h_single}):
\begin{equation}\label{eq:Hv}
    \mathbf{H}_\mathbf{v} 
    = \frac{\partial h}{\partial \mathbf{v}}
    \bigg|_{\hat{\mathcal{X}}_k^-,\,
    \mathbf{v}=\mathbf{0}} 
    = \mathbf{I}_{\sum_i \nu_i}
\end{equation}
so the gain equation 
(Section~\ref{sec:ekf_eqs}) simplifies to:
\begin{equation}
    \mathbf{K}_k = \mathbf{P}_k^-\,
    \mathbf{H}_k^\top
    \left[\mathbf{H}_k\,\mathbf{P}_k^-\,
    \mathbf{H}_k^\top + \mathbf{R}
    \right]^{-1}
\end{equation}

\subsubsection{Indirect Correction of 
Non-Observed States}

The measurement Jacobian has zeros in the 
$\delta\boldsymbol{\xi}$, $\delta\mathbf{v}$, 
$\delta\boldsymbol{\omega}$, 
$\delta\mathbf{g}^B$, and $\delta\mathbf{d}^B$ 
columns: the measurement model projects 
body-frame feature positions through fixed 
camera extrinsics, so pose, velocity, rotation 
rate, gravity, and disturbance do not appear. 
These states are corrected indirectly through 
covariance cross-correlations built during 
propagation: 
$\dot{\mathbf{p}}_i^B$ depends on $\mathbf{v}$ 
and $\boldsymbol{\omega}$ 
(Equation~\ref{eq:pdot}), and 
$\dot{T}$ depends on $\mathbf{v}$ and 
$\boldsymbol{\omega}$ 
(Equation~\ref{eq:Tdot}), so $\mathbf{P}$ 
develops off-diagonal blocks that couple 
feature-point innovations to velocity, rotation, 
pose, gravity, and disturbance corrections.

\subsubsection{Gravity Pseudo-Measurement}

The gravity magnitude $\|\mathbf{g}^B\| \approx g$ is 
known a priori. We encode this as a scalar 
measurement to prevent magnitude drift:
\begin{equation}\label{eq:h_gravity}
    \tilde{\mathbf{y}}_g = h_g(\mathcal{X}, v_g) 
    = \|\mathbf{g}^B\|^2 + v_g 
    = g^2 + v_g
\end{equation}
where $v_g \sim \mathcal{N}(0, R_g)$ is scalar 
measurement noise. The "observation" is the 
known constant $\tilde{\mathbf{y}}_g = g^2$. $R_g$ should reflect the minor variations in gravity ($g=9.81$ is a common approximation whose true value varies along the earth's surface and decays with height). $R_g$ should be small; unmodelled forces ought to be reflected in $\mathbf{d}^B$.

\paragraph{Jacobians.}
The state Jacobian 
$H_g = \frac{Dh_g}{D\mathcal{X}}
\big|_{\hat{\mathcal{X}}_k,\,v_g=0}$. Since 
$h_g$ depends only on $\mathbf{g}^B 
\in \mathbb{R}^3$ (Euclidean), the composite 
derivative reduces to the standard partial 
derivative:
\begin{equation}
    \frac{\partial\,\|\mathbf{g}^B\|^2}
    {\partial\,\mathbf{g}^B} 
    = 2(\mathbf{g}^B)^\top
\end{equation}
All other state components do not appear in 
$h_g$, giving:
\begin{equation}
    H_g = \begin{pmatrix}
    \mathbf{0}_{1\times 6} & 
    \mathbf{0}_{1\times 3} & 
    \mathbf{0}_{1\times 3} & 
    2(\hat{\mathbf{g}}^B)^\top & 
    \mathbf{0}_{1\times 3} & 
    \mathbf{0}_{1\times 3} & 
    \cdots & \mathbf{0}_{1\times 3}
    \end{pmatrix}
    \in \mathbb{R}^{1\times(18+3N_f)}
\end{equation}
The noise Jacobian is:
\begin{equation}
    H_{v_g} = \frac{\partial h_g}{\partial v_g}
    \bigg|_{v_g=0} = 1
\end{equation}
so the gain equation gives:
\begin{equation}
    K_g = \mathbf{P}\,H_g^\top
    \left[H_g\,\mathbf{P}\,H_g^\top 
    + R_g\right]^{-1} 
    \in \mathbb{R}^{(18+3N_f)\times 1}
\end{equation}
which is a scalar division (the bracketed term 
is $1\times1$).

\paragraph{Sequential update.}
The gravity pseudo-measurement can be processed as a 
separate update step after the feature-point 
update. This is valid because the Kalman update 
equations can be applied any number of times 
within a single time step by splitting the 
measurement vector into independent groups 
\citep{Wescott2022SequentialKalmanUpdate}. 
Each group updates 
$\hat{\mathcal{X}}$ and $\mathbf{P}$ using the 
posterior from the previous group as the prior 
for the next. The result is algebraically 
equivalent to processing all measurements 
simultaneously in a single large update, 
provided the measurement noises are independent. 
This is not to be confused with the iterated EKF 
(IEKF), which re-linearises the measurement 
model at the updated state and solves an 
optimisation problem iteratively.

\noindent We stack all measurements into one update however since deriving the pose change and its covariance becomes easier. 
\subsection{Relative Pose from Filter State}
\label{sec:relative_pose}

The relative pose 
$\Delta T = T_{B_k,B_{k-1}} \in SE(3)$ and its 
covariance are extracted directly from the filter 
state.

\paragraph{Relative Pose}

After propagation at step $k$, the discrete estimated 
poses are $\hat{T}_{k-1}^+$ and $\hat{T}_k^-$ (or 
$\hat{T}_k^+$ after the update). From 
$\Delta T = T_{B_k,B_0}\cdot T_{B_{k-1},B_0}^{-1}$:
\begin{equation}\label{eq:DeltaT}
    \Delta\hat{T} = \hat{T}_k\cdot\hat{T}_{k-1}^{-1}
\end{equation}
with rotation and translation blocks:
\begin{equation}
    \Delta\hat{\mathbf{R}} 
    = \hat{\mathbf{R}}_{B_k}\,
    \hat{\mathbf{R}}_{B_{k-1}}^\top, \qquad
    \Delta\hat{\mathbf{t}} 
    = \hat{\mathbf{t}}_{B_k} 
    - \Delta\hat{\mathbf{R}}\,
    \hat{\mathbf{t}}_{B_{k-1}}
\end{equation}
\textit{Derivation.} From 
$\mathbf{p}^{B_k} 
= \mathbf{R}_{B_k}\mathbf{p}^{B_0} 
+ \mathbf{t}_{B_k}$ and 
$\mathbf{p}^{B_0} 
= \mathbf{R}_{B_{k-1}}^\top
(\mathbf{p}^{B_{k-1}} - \mathbf{t}_{B_{k-1}})$, 
substitute:
\begin{equation}
    \mathbf{p}^{B_k} 
    = \underbrace{\mathbf{R}_{B_k}
    \mathbf{R}_{B_{k-1}}^\top}_{\Delta\mathbf{R}}
    \,\mathbf{p}^{B_{k-1}} 
    + \underbrace{\mathbf{t}_{B_k} 
    - \mathbf{R}_{B_k}\mathbf{R}_{B_{k-1}}^\top
    \mathbf{t}_{B_{k-1}}}_{\Delta\mathbf{t}}
\end{equation}

\paragraph{Perturbation of the Relative Pose}

The filter uses right perturbation 
(Section~\ref{sec:composite}): 
$T_{B_j,\mathrm{true}} 
= \hat{T}_{B_j}\cdot\mathrm{Exp}(
\delta\boldsymbol{\xi}_j)$ where 
$\delta\boldsymbol{\xi}_j \in \mathbb{R}^6$ lives 
in $B_0$ (the domain frame of $T_{B_j,B_0}$) for 
both $j = k{-}1$ and $j = k$. The relative pose 
perturbation 
$\delta\Delta\boldsymbol{\xi} \in \mathbb{R}^6$, 
defined by 
$\Delta T_{\mathrm{true}} 
= \Delta\hat{T}\cdot\mathrm{Exp}(
\delta\Delta\boldsymbol{\xi})$, lives in $B_{k-1}$ 
(the domain frame of $\Delta T_{B_k,B_{k-1}}$). \\

\noindent\textit{Derivation.} Perturb both poses:
\begin{equation}
    \Delta T_{\mathrm{true}} 
    = \hat{T}_{B_k}\,\mathrm{Exp}(
    \delta\boldsymbol{\xi}_k)
    \cdot\left(\hat{T}_{B_{k-1}}\,\mathrm{Exp}(
    \delta\boldsymbol{\xi}_{k-1})\right)^{-1}
\end{equation}
Invert the second factor:
\begin{equation}
    = \hat{T}_{B_k}\,\mathrm{Exp}(
    \delta\boldsymbol{\xi}_k)\,
    \mathrm{Exp}(-\delta\boldsymbol{\xi}_{k-1})\,
    \hat{T}_{B_{k-1}}^{-1}
\end{equation}
Both $\delta\boldsymbol{\xi}_k$ and 
$\delta\boldsymbol{\xi}_{k-1}$ are small filter 
perturbations. Apply 
\citep[Eq.~69]{sola2021microlietheorystate}:
\begin{equation}
    \mathrm{Exp}(\boldsymbol{\tau})\,
    \mathrm{Exp}(\delta\boldsymbol{\tau}) 
    \approx \mathrm{Exp}(\boldsymbol{\tau} 
    + \mathbf{J}_r^{-1}(\boldsymbol{\tau})\,
    \delta\boldsymbol{\tau})
\end{equation}
with $\boldsymbol{\tau} 
= \delta\boldsymbol{\xi}_k$ and 
$\delta\boldsymbol{\tau} 
= -\delta\boldsymbol{\xi}_{k-1}$. Since 
$\delta\boldsymbol{\xi}_k$ is small, 
$\mathbf{J}_r^{-1}(\delta\boldsymbol{\xi}_k) 
\approx \mathbf{I}_6$: the $SE(3)$ right Jacobian 
evaluated at the origin gives $\mathbf{I}_6$, as 
$\mathbf{J}_l(\boldsymbol{\theta})$ 
\citep[Eq.~145]{sola2021microlietheorystate} 
has $\sin\theta/\theta \to 1$ and 
$[\mathbf{u}]_\times$ terms vanishing, 
$Q(\boldsymbol{\rho},\boldsymbol{\theta})$ 
\citep[Eq.~180]{sola2021microlietheorystate} 
has every term proportional to 
$\boldsymbol{\rho}_\times$ or 
$\boldsymbol{\theta}_\times$ so 
$Q(\mathbf{0},\mathbf{0}) = \mathbf{0}$, giving 
$\mathbf{J}_l(\mathbf{0}) = \mathbf{I}_6$ 
\citep[Eq.~179a]{sola2021microlietheorystate}, 
and $\mathbf{J}_r(\boldsymbol{\tau}) 
= \mathbf{J}_l(-\boldsymbol{\tau})$ 
\citep[Eq.~76]{sola2021microlietheorystate}. 
Therefore:
\begin{equation}
    \Delta T_{\mathrm{true}} 
    = \hat{T}_{B_k}\,\mathrm{Exp}(
    \delta\boldsymbol{\xi}_k 
    - \delta\boldsymbol{\xi}_{k-1})\,
    \hat{T}_{B_{k-1}}^{-1}
\end{equation}
Insert $\mathcal{E} 
= \hat{T}_{B_{k-1}}^{-1}\hat{T}_{B_{k-1}}$ 
between $\hat{T}_{B_k}$ and $\mathrm{Exp}(\cdot)$:
\begin{equation}
    = \underbrace{\hat{T}_{B_k}\,
    \hat{T}_{B_{k-1}}^{-1}}_{\Delta\hat{T}}
    \cdot\hat{T}_{B_{k-1}}\,\mathrm{Exp}(
    \delta\boldsymbol{\xi}_k 
    - \delta\boldsymbol{\xi}_{k-1})\,
    \hat{T}_{B_{k-1}}^{-1}
\end{equation}
Apply the adjoint property 
$\mathcal{X}\,\mathrm{Exp}(\boldsymbol{\tau})\,
\mathcal{X}^{-1} 
= \mathrm{Exp}(\mathrm{Ad}_\mathcal{X}\,
\boldsymbol{\tau})$, obtained from 
\citep[Eq.~32]{sola2021microlietheorystate}: 
$\mathcal{X} \oplus \boldsymbol{\tau} 
= (\mathrm{Ad}_\mathcal{X}\,\boldsymbol{\tau}) 
\oplus \mathcal{X}$, expanded as 
$\mathcal{X}\cdot\mathrm{Exp}(\boldsymbol{\tau}) 
= \mathrm{Exp}(\mathrm{Ad}_\mathcal{X}\,
\boldsymbol{\tau})\cdot\mathcal{X}$ and 
right-multiplied by $\mathcal{X}^{-1}$. With 
$\mathcal{X} = \hat{T}_{B_{k-1}}$:
\begin{equation}
    \Delta T_{\mathrm{true}} 
    = \Delta\hat{T}\cdot\mathrm{Exp}\!\left(
    \mathrm{Ad}_{\hat{T}_{B_{k-1}}}(
    \delta\boldsymbol{\xi}_k 
    - \delta\boldsymbol{\xi}_{k-1})\right)
\end{equation}
Reading off the right perturbation of $\Delta T$:
\begin{equation}\label{eq:delta_DeltaXi}
    \boxed{
    \delta\Delta\boldsymbol{\xi} 
    = \mathrm{Ad}_{\hat{T}_{B_{k-1}}}(
    \delta\boldsymbol{\xi}_k 
    - \delta\boldsymbol{\xi}_{k-1})
    }
\end{equation}
where $\mathrm{Ad}_{\hat{T}_{B_{k-1}}}$ 
(Equation~\ref{eq:adjoint_se3}) maps from $B_0$ 
(where $\delta\boldsymbol{\xi}_j$ live) to $B_{k-1}$ 
(where $\delta\Delta\boldsymbol{\xi}$ lives). \\

\noindent\textit{Verification.} At 
$\hat{T}_{B_{k-1}} = \mathcal{E}$: 
$\mathrm{Ad}_\mathcal{E} = \mathbf{I}_6$, so 
$\delta\Delta\boldsymbol{\xi} 
= \delta\boldsymbol{\xi}_k 
- \delta\boldsymbol{\xi}_{k-1}$. $\checkmark$

\paragraph{Jacobian and Covariance}

From Equation~\ref{eq:delta_DeltaXi}, the 
perturbation is linear in 
$(\delta\boldsymbol{\xi}_{k-1},\,
\delta\boldsymbol{\xi}_k)$:
\begin{equation}\label{eq:J_Delta}
    \delta\Delta\boldsymbol{\xi} 
    = \underbrace{\begin{pmatrix}
    -\mathrm{Ad}_{\hat{T}_{B_{k-1}}} & 
    \mathrm{Ad}_{\hat{T}_{B_{k-1}}}
    \end{pmatrix}}_{\mathbf{J}_\Delta 
    \,\in\, \mathbb{R}^{6\times12}}
    \begin{pmatrix}
    \delta\boldsymbol{\xi}_{k-1} \\
    \delta\boldsymbol{\xi}_k
    \end{pmatrix}
\end{equation}
By Equation~\ref{eq:cov_linear}:
\begin{equation}\label{eq:Sigma_DeltaT}
    \boldsymbol{\Sigma}_{\Delta\xi} 
    = \mathbf{J}_\Delta\,
    \mathbf{P}_{\mathrm{pose,joint}}\,
    \mathbf{J}_\Delta^\top 
    \in \mathbb{R}^{6\times6}
\end{equation}
where $\mathbf{P}_{\mathrm{pose,joint}}$ is the 
$12\times12$ joint covariance of 
$(\delta\boldsymbol{\xi}_{k-1},\,
\delta\boldsymbol{\xi}_k)$, derived below.

\paragraph{Cross-covariance between time steps.}
The filter covariance $\mathbf{P}$ at any single 
time step contains only the perturbation at that 
time. To assemble 
$\mathbf{P}_{\mathrm{pose,joint}}$, we need the 
cross-covariance between the perturbations at 
$k{-}1$ and $k$. \\

\noindent The discrete covariance propagation 
$\mathbf{P}_{k}^- = \Phi_{k-1}\,
\mathbf{P}_{k-1}^+\,\Phi_{k-1}^\top 
+ \Delta t_{k-1}\,G_{k-1}\,\mathcal{Q}_{k-1}\,
G_{k-1}^\top$ 
(Section~\ref{sec:ekf_eqs}) corresponds to the 
linearised error relation:
\begin{equation}\label{eq:error_prop}
    \delta\mathbf{x}_k^- 
    = \Phi_{k-1}\,\delta\mathbf{x}_{k-1}^+ 
    + \mathbf{n}_{k-1}
\end{equation}
where $\Phi_{k-1} = \mathbf{I} 
+ \Delta t_{k-1}\,F_{k-1}$ is the first-order 
state transition matrix (the discretisation of 
the continuous error dynamics 
$\dot{\delta\mathbf{x}} = F\,\delta\mathbf{x}$, 
Section~\ref{sec:F_matrix}), and 
$\mathbf{n}_{k-1}$ is the discretised process 
noise with covariance 
$\Delta t_{k-1}\,G_{k-1}\,\mathcal{Q}_{k-1}\,
G_{k-1}^\top$, independent of 
$\delta\mathbf{x}_{k-1}^+$.

\noindent\textit{Verification}: applying 
Equation~\ref{eq:cov_linear} and 
Equation~\ref{eq:cov_indep}:
\begin{align}
    \mathrm{Cov}(\delta\mathbf{x}_k^-) 
    &= \Phi_{k-1}\,\mathrm{Cov}(
    \delta\mathbf{x}_{k-1}^+)\,\Phi_{k-1}^\top 
    + \mathrm{Cov}(\mathbf{n}_{k-1}) \nonumber \\
    &= \Phi_{k-1}\,\mathbf{P}_{k-1}^+\,
    \Phi_{k-1}^\top 
    + \Delta t_{k-1}\,G_{k-1}\,\mathcal{Q}_{k-1}\,
    G_{k-1}^\top 
    = \mathbf{P}_k^-
\end{align}
which recovers the propagation equation. 
$\checkmark$ \\

\noindent The cross-covariance follows from 
Equation~\ref{eq:cross_cov_sum} with 
$\mathbf{c} = \delta\mathbf{x}_k^-$, 
$\mathbf{M} = \Phi_{k-1}$, 
$\mathbf{a} = \delta\mathbf{x}_{k-1}^+$, 
$\mathbf{b} = \delta\mathbf{x}_{k-1}^+$, and 
$\mathbf{n} = \mathbf{n}_{k-1}$ independent of 
$\delta\mathbf{x}_{k-1}^+$:
\begin{equation}\label{eq:cross_cov_prop}
    \mathrm{Cov}(\delta\mathbf{x}_k^-,\;
    \delta\mathbf{x}_{k-1}^+) 
    = \Phi_{k-1}\,\underbrace{\mathrm{Cov}(
    \delta\mathbf{x}_{k-1}^+,\;
    \delta\mathbf{x}_{k-1}^+)}_{
    \mathbf{P}_{k-1}^+} 
    = \Phi_{k-1}\,\mathbf{P}_{k-1}^+
\end{equation}

\paragraph{Pre-update joint covariance.}
Using $\hat{T}_k^-$ (after propagation, before 
update at $k$). Let subscript $\xi$ denote the 
$6\times6$ pose block:
\begin{equation}\label{eq:P_joint_pre}
    \mathbf{P}_{\mathrm{pose,joint}}^{-} 
    = \begin{pmatrix}
    \mathbf{P}_{\xi\xi}^{k-1,+} 
    & \left(\Phi_{k-1}\,\mathbf{P}^{k-1,+}
    \right)_{\xi\xi}^\top \\[4pt]
    \left(\Phi_{k-1}\,\mathbf{P}^{k-1,+}
    \right)_{\xi\xi} 
    & \mathbf{P}_{\xi\xi}^{k,-}
    \end{pmatrix}
\end{equation}
where:
\begin{itemize}
    \item $\mathbf{P}_{\xi\xi}^{k-1,+}$: the 
    $6\times6$ pose block after the update at 
    $k{-}1$. 
    \item $\mathbf{P}_{\xi\xi}^{k,-}$: the 
    $6\times6$ pose block after propagation to 
    $k$. Larger than 
    $\mathbf{P}_{\xi\xi}^{k-1,+}$ because 
    propagation adds process noise.
    \item $(\Phi_{k-1}\,\mathbf{P}^{k-1,+}
    )_{\xi\xi}$: pose rows of 
    $\Phi_{k-1}\,\mathbf{P}^{k-1,+}$ restricted 
    to pose columns. This is \emph{not} simply 
    $(\Phi_{k-1})_{\xi\xi}\,
    \mathbf{P}_{\xi\xi}^{k-1,+}$: the full 
    product is needed because $\Phi_{k-1}$ 
    couples pose to velocity and rotation through 
    $F_{\delta\boldsymbol{\xi},
    (\delta\mathbf{v},\delta\boldsymbol{\omega})} 
    = -\mathrm{Ad}_{\hat{T}^{-1}}$ 
    (Equation~\ref{eq:F_se3}), so the 
    cross-covariance picks up contributions from 
    the velocity and rotation uncertainty at 
    $k{-}1$.
\end{itemize}

\paragraph{Post-update joint covariance.}
Using $\hat{T}_k^+$ (after the update at $k$). The 
update gives 
$\mathbf{P}_k^+ = (\mathbf{I} 
- \mathbf{K}_k\mathbf{H}_k)\mathbf{P}_k^-$, 
corresponding to the error transformation:
\begin{equation}\label{eq:error_update}
    \delta\mathbf{x}_k^+ 
    = (\mathbf{I} - \mathbf{K}_k\mathbf{H}_k)\,
    \delta\mathbf{x}_k^- 
    + \mathbf{K}_k\mathbf{v}_k
\end{equation}
where $\mathbf{v}_k$ is measurement noise, 
independent of $\delta\mathbf{x}_{k-1}^+$. 
\textit{Verification}: applying 
Equation~\ref{eq:cov_linear} and 
Equation~\ref{eq:cov_indep} (noting 
$\mathrm{Cov}(\mathbf{K}\mathbf{v}) 
= \mathbf{K}\mathbf{R}_k\mathbf{K}^\top$ and 
$(\mathbf{I} - \mathbf{KH})\mathbf{P}^-
(\mathbf{I} - \mathbf{KH})^\top 
+ \mathbf{K}\mathbf{R}_k\mathbf{K}^\top 
= (\mathbf{I} - \mathbf{KH})\mathbf{P}^-$ 
by the Joseph form identity when $\mathbf{K}$ is 
the optimal gain):
\begin{equation}
    \mathrm{Cov}(\delta\mathbf{x}_k^+) 
    = (\mathbf{I} - \mathbf{K}_k\mathbf{H}_k)\,
    \mathbf{P}_k^- = \mathbf{P}_k^+ 
    \quad\checkmark
\end{equation}
The cross-covariance with 
$\delta\mathbf{x}_{k-1}^+$ follows from 
Equation~\ref{eq:cross_cov_sum} with 
$\mathbf{c} = \delta\mathbf{x}_k^+$, 
$\mathbf{M} = (\mathbf{I} 
- \mathbf{K}_k\mathbf{H}_k)$, 
$\mathbf{a} = \delta\mathbf{x}_k^-$, 
$\mathbf{b} = \delta\mathbf{x}_{k-1}^+$, and 
$\mathbf{n} = \mathbf{K}_k\mathbf{v}_k$ 
independent of $\delta\mathbf{x}_{k-1}^+$:
\begin{align}\label{eq:cross_cov_post}
    \mathrm{Cov}(\delta\mathbf{x}_k^+,\;
    \delta\mathbf{x}_{k-1}^+) 
    &= (\mathbf{I} 
    - \mathbf{K}_k\mathbf{H}_k)\,
    \mathrm{Cov}(\delta\mathbf{x}_k^-,\;
    \delta\mathbf{x}_{k-1}^+) \nonumber \\
    &= (\mathbf{I} 
    - \mathbf{K}_k\mathbf{H}_k)\,
    \Phi_{k-1}\,\mathbf{P}^{k-1,+}
\end{align}
substituting Equation~\ref{eq:cross_cov_prop} in 
the second step. The $12\times12$ joint covariance 
becomes:
\begin{equation}\label{eq:P_joint_post}
    \mathbf{P}_{\mathrm{pose,joint}}^{+} 
    = \begin{pmatrix}
    \mathbf{P}_{\xi\xi}^{k-1,+} 
    & \left((\mathbf{I} 
    - \mathbf{K}_k\mathbf{H}_k)\,\Phi_{k-1}\,
    \mathbf{P}^{k-1,+}\right)_{\xi\xi}^\top 
    \\[4pt]
    \left((\mathbf{I} 
    - \mathbf{K}_k\mathbf{H}_k)\,\Phi_{k-1}\,
    \mathbf{P}^{k-1,+}\right)_{\xi\xi} 
    & \mathbf{P}_{\xi\xi}^{k,+}
    \end{pmatrix}
\end{equation}

\paragraph{Which version to use.}
The pre-update version 
(Equation~\ref{eq:P_joint_pre}) is simpler: it is 
available immediately after propagation and requires 
only $\Phi_{k-1}\,\mathbf{P}^{k-1,+}$. The 
post-update version 
(Equation~\ref{eq:P_joint_post}) is tighter: the 
update at $k$ reduces 
$\mathbf{P}_{\xi\xi}^{k,+}$ relative to 
$\mathbf{P}_{\xi\xi}^{k,-}$ and tightens the 
cross-covariance, yielding a more informative prior. 
Since the pose is corrected indirectly through 
feature cross-correlations 
(Section~\ref{sec:measurement}), the improvement 
depends on the quality of feature matches at $k$. 
In practice, the post-update version should be used
when the joint solver runs after the filter update. \\

\noindent The post-update covariance occasionally
comes out with eigenvalues near $-10^{-7}$, well
above the floating-point noise floor. We clip
eigenvalues at $10^{-12}$, which masks the symptom
rather than fixing it. The likely sources are the
selector construction across a changing state
dimension and the $\mathbf{J}_r^{-1} \approx
\mathbf{I}$ approximation above.

\paragraph{Implementation.}
Before propagation at step $k$, store the pose rows 
of $\mathbf{P}^{k-1,+}$ (a $6\times n$ slice where 
$n = 18 + 3N_f$). After propagation, compute 
$(\Phi_{k-1}\,\mathbf{P}^{k-1,+})_{\xi\xi}$ by 
extracting the pose rows of the product and 
restricting to pose columns. For the post-update 
version, apply $(\mathbf{I} 
- \mathbf{K}_k\mathbf{H}_k)$ to this slice after 
the update. This is $O(n)$ additional storage per 
frame.

\paragraph{Conversion to $SE(3)$ Lie Algebra}

For the joint solver 
(Section~\ref{sec:joint_solver}), the prior is 
expressed in the Lie algebra:
\begin{equation}\label{eq:DeltaXi_prior}
    \Delta\hat{\boldsymbol{\xi}} 
    = \mathrm{Log}(\Delta\hat{T}) 
    \in \mathbb{R}^6, \qquad
    \boldsymbol{\Sigma}_{\mathrm{prior}} 
    = \mathbf{J}_r^{-1}\,
    \boldsymbol{\Sigma}_{\Delta\xi}\,
    \mathbf{J}_r^{-\top}
\end{equation}
where $\mathbf{J}_r^{-1}$ is evaluated at 
$\Delta\hat{\boldsymbol{\xi}}$ 
\citep[Eq.~79]{sola2021microlietheorystate}, which gives
$\mathbf{J}^{\mathrm{Log}(\mathcal{X})}_\mathcal{X}
= \mathbf{J}_r^{-1}(\boldsymbol{\tau})$ with
$\boldsymbol{\tau}
= \mathrm{Log}(\mathcal{X})$. For one frame
interval at typical frame rates, the relative pose 
is near the identity, 
$\mathbf{J}_r^{-1} \approx \mathbf{I}_6$
\citep[Eqs.~179a,~145,~76]{sola2021microlietheorystate}, and
$\boldsymbol{\Sigma}_{\mathrm{prior}} 
\approx \boldsymbol{\Sigma}_{\Delta\xi}$.

\newpage
\section{Joint MAP Solver}
\label{sec:map_solver}

\subsection{Nonlinear MAP with Gauss-Newton}
\label{sec:map}

Following the formulation in 
\citep{haavardsholm2026tek5030wk9}.

\subsubsection{Residual and Noise Model}

For measurements $\mathbf{z}_i$ and latent state 
$\mathbf{x}$, assume:
\begin{equation}
    \mathbf{z}_i = h_i(\mathbf{x}) 
    + \boldsymbol{\eta}_i, 
    \qquad 
    \boldsymbol{\eta}_i \sim \mathcal{N}(\mathbf{0}, 
    \boldsymbol{\Sigma}_i)
\end{equation}
The predicted measurement is 
$\hat{\mathbf{z}}_i = h_i(\mathbf{x})$, and the 
measurement residual is:
\begin{equation}
    e_i(\mathbf{x}) = h_i(\mathbf{x}) - \mathbf{z}_i
\end{equation}

\subsubsection{Nonlinear MAP as Weighted Least Squares}

Under Gaussian noise, the MAP estimate is obtained by 
solving the nonlinear least squares problem:
\begin{equation}
    \mathbf{x}^* = \arg\min_{\mathbf{x}} 
    \sum_{i=1}^n 
    \|h_i(\mathbf{x}) 
    - \mathbf{z}_i\|^2_{\boldsymbol{\Sigma}_i}
\end{equation}
where the squared Mahalanobis norm is:
\begin{equation}
    \|\mathbf{e}\|^2_{\boldsymbol{\Sigma}} 
    = \mathbf{e}^\top\,\boldsymbol{\Sigma}^{-1}\,
    \mathbf{e} 
    = \|\boldsymbol{\Sigma}^{-1/2}\,
    \mathbf{e}\|^2
\end{equation}
This yields the whitened Jacobian and residual:
\begin{align}
    A_i &= \boldsymbol{\Sigma}_i^{-1/2}\,
    J_{\mathbf{x}}^{h_i}\big|_{\hat{\mathbf{x}}^t} 
    \\
    b_i &= \boldsymbol{\Sigma}_i^{-1/2}\,
    (\mathbf{z}_i - h_i(\hat{\mathbf{x}}^t))
\end{align}
Stack all measurements:
\begin{equation}
    A = \begin{pmatrix} A_1 \\ \vdots \\ A_n 
    \end{pmatrix}, \qquad
    b = \begin{pmatrix} b_1 \\ \vdots \\ b_n 
    \end{pmatrix}
\end{equation}

\subsubsection{Gauss-Newton Iteration}

At iteration $t$, linearise the residual around the 
current estimate $\hat{\mathbf{x}}^t$:
\begin{equation}
    h_i(\mathbf{x}) \approx h_i(\hat{\mathbf{x}}^t) 
    + J_{\mathbf{x}}^{h_i}\big|_{\hat{\mathbf{x}}^t}
    \,\boldsymbol{\tau}
\end{equation}
where $\boldsymbol{\tau}$ is the increment. The 
Gauss-Newton step solves:
\begin{equation}
    \boldsymbol{\tau}^* = \arg\min_{\boldsymbol{\tau}} 
    \|A\boldsymbol{\tau} - b\|^2
\end{equation}
via the normal equations 
$A^\top A\,\boldsymbol{\tau}^* = A^\top b$. Update:
\begin{equation}
    \hat{\mathbf{x}}^{t+1} = \hat{\mathbf{x}}^t 
    \oplus \boldsymbol{\tau}^*
\end{equation}
where $\oplus$ is addition in Euclidean space or the 
right plus operator on the appropriate manifold. 
Note the Jacobian must also be defined on the 
manifold.

\subsubsection{Estimated Covariance}

The covariance of the MAP estimate is approximated by 
the inverse of the approximate Hessian at the solution:
\begin{equation}
    \boldsymbol{\Sigma}_{\hat{\mathbf{x}}} 
    \approx (A^\top A)^{-1}
\end{equation}
so that $\mathbf{x} \sim 
\mathcal{N}(\hat{\mathbf{x}}, 
\boldsymbol{\Sigma}_{\hat{\mathbf{x}}})$.

\subsection{Including a Prior}
\label{sec:prior}

If a prior on part or all of the state is available, 
it enters as an additional term in the objective. 
This is seen by noticing the prior adds a quadratic 
penalty to the negative log-posterior 
\citep[Equation~34]{wilkinson2022bayesnewtonmethodsapproximatebayesian}. The prior is then treated as a virtual 
measurement and stacked into $A$ and $b$ alongside 
the measurement terms.

\paragraph{Full prior on Euclidean state.}
If $\mathbf{x} \in \mathbb{R}^n$ with prior 
$\mathbf{x} \sim \mathcal{N}(\boldsymbol{\mu}_0,\,
\boldsymbol{\Sigma}_0)$:
\begin{equation}
    A_0 = \boldsymbol{\Sigma}_0^{-1/2}\,\mathbf{I}_n, 
    \qquad
    b_0 = \boldsymbol{\Sigma}_0^{-1/2}\,
    (\boldsymbol{\mu}_0 - \hat{\mathbf{x}}^t)
\end{equation}
The residual $b_0$ is \emph{prior minus estimate}, 
consistent with the general formula 
$b_i = \boldsymbol{\Sigma}_i^{-1/2}
(\mathbf{z}_i - h_i(\hat{\mathbf{x}}^t))$.

\paragraph{Partial prior.}
If the prior applies only to a subset of the state 
(e.g.\ $\mathbf{x} = (\mathbf{x}_a,\,\mathbf{x}_b)$ 
with prior on $\mathbf{x}_a$ only), the selection 
matrix replaces the identity:
\begin{equation}
    A_0 = \boldsymbol{\Sigma}_0^{-1/2}
    \begin{pmatrix}\mathbf{I}_{n_a} & 
    \mathbf{0}_{n_a \times n_b}\end{pmatrix}, 
    \qquad
    b_0 = \boldsymbol{\Sigma}_0^{-1/2}\,
    (\boldsymbol{\mu}_0 - \hat{\mathbf{x}}_a^t)
\end{equation}
where $n_a$ is the dimension of $\mathbf{x}_a$. 
This penalises only the $\boldsymbol{\tau}_a$ 
increment, leaving $\boldsymbol{\tau}_b$ 
unconstrained by the prior.

\paragraph{Prior on a manifold state.}
If the state (or the prior subset) includes a 
component $\mathcal{X} \in \mathcal{M}$ on a Lie 
group with prior mean $\bar{\mathcal{X}}$ and 
covariance $\boldsymbol{\Sigma}_0$ defined on the 
tangent space, the Euclidean difference 
$\boldsymbol{\mu}_0 - \hat{\mathbf{x}}^t$ is 
replaced by the manifold $\ominus$ operator 
(Section~\ref{sec:lie_algebra}):
\begin{equation}
    b_0 = \boldsymbol{\Sigma}_0^{-1/2}\,
    (\bar{\mathcal{X}} \ominus \hat{\mathcal{X}}^t)
    = \boldsymbol{\Sigma}_0^{-1/2}\,
    \mathrm{Log}\!\left(
    (\hat{\mathcal{X}}^t)^{-1}\cdot
    \bar{\mathcal{X}}\right)
\end{equation}
This is \emph{prior $\ominus$ estimate}, the 
manifold analogue of prior minus estimate. At 
$\hat{\mathcal{X}}^t = \bar{\mathcal{X}}$ the 
residual is $\mathrm{Log}(\mathcal{E}) 
= \mathbf{0}$, as required. The sign follows from 
the virtual measurement formulation: the residual 
$e_0 = \mathcal{X} \ominus \bar{\mathcal{X}} 
= \mathrm{Log}(\bar{\mathcal{X}}^{-1}\cdot
\mathcal{X})$ and $b_0 = -\boldsymbol{\Sigma}_0^{-1/2}
\,e_0|_{\hat{\mathcal{X}}^t}$, using 
$-\mathrm{Log}(X) = \mathrm{Log}(X^{-1})$ 
\citep[Eqs.~19,~23,~24]{sola2021microlietheorystate}. 
The $A_0$ matrix is unchanged:
\begin{equation}
    A_0 = \boldsymbol{\Sigma}_0^{-1/2}\,\mathbf{I}_m
\end{equation}
where $m = \dim(\mathcal{M})$ is the tangent space
dimension (e.g.\ $m = 6$ for $SE(3)$), padded with
zeros if the prior is partial. \\

\noindent $A_0$ is the whitened Jacobian of the
prior's measurement function. In the Euclidean case
that function is the identity on the state, so the
Jacobian is $\mathbf{I}$ exactly. On a manifold the
derivative is the composite one
(Section~\ref{sec:composite}): it is taken with
respect to a perturbation of $\hat{\mathcal{X}}^t$,
whereas the residual and $\boldsymbol{\Sigma}_0$
live in the tangent space at $\bar{\mathcal{X}}$.
Writing $\mathbf{I}$ asserts that no conversion
between those two tangent spaces is needed. \\

\noindent We believe the exact form to be
$A_0 = \boldsymbol{\Sigma}_0^{-1/2}\,
\mathbf{J}_l^{-1}(\boldsymbol{\delta})$, with
$\boldsymbol{\delta}
= \bar{\mathcal{X}} \ominus \hat{\mathcal{X}}^t$,
following the expansion
$\mathrm{Log}(\mathrm{Exp}(\delta\boldsymbol{\tau})\,
\mathrm{Exp}(\boldsymbol{\tau}))
\approx \boldsymbol{\tau}
+ \mathbf{J}_l^{-1}(\boldsymbol{\tau})\,
\delta\boldsymbol{\tau}$
\citep[Eq.~74, p.~10]{sola2021microlietheorystate}.
We use $\mathbf{J}_l^{-1} \approx \mathbf{I}$, which
is valid near the prior mean; over a single frame
interval the prior residual is small.

\paragraph{Combined manifold and Euclidean prior.}
If the prior covers both a Lie group component 
$\mathcal{X} \in \mathcal{M}$ and a Euclidean 
component $\mathbf{x}_e \in \mathbb{R}^n$ 
(e.g.\ pose and a point position), with joint prior 
mean $(\bar{\mathcal{X}},\,\boldsymbol{\mu}_e)$ 
and joint covariance $\boldsymbol{\Sigma}_0 
\in \mathbb{R}^{(m+n)\times(m+n)}$, the residual 
is assembled block-wise using the appropriate 
operator for each component:
\begin{equation}
    b_0 = \boldsymbol{\Sigma}_0^{-1/2}
    \begin{pmatrix}
    \bar{\mathcal{X}} \ominus \hat{\mathcal{X}}^t \\
    \boldsymbol{\mu}_e - \hat{\mathbf{x}}_e^t
    \end{pmatrix}
\end{equation}
with $A_0 = \boldsymbol{\Sigma}_0^{-1/2}
(\mathbf{I}_{m+n},\;\mathbf{0})$ if the prior is 
partial, or $A_0 = \boldsymbol{\Sigma}_0^{-1/2}
\mathbf{I}_{m+n}$ if it covers the full state. The 
joint covariance $\boldsymbol{\Sigma}_0$ may have 
nonzero off-diagonal blocks encoding correlations 
between the manifold and Euclidean components; the 
whitening $\boldsymbol{\Sigma}_0^{-1/2}$ handles 
this automatically.
\subsection{Joint Point-Velocity Solver
with Pose Prior}
\label{sec:joint_solver}

The two-frame point-velocity solver estimates 
position and velocity for all tracked points jointly 
with the pose change, using the 
filter-derived pose 
(Section~\ref{sec:relative_pose}) as a prior. The 
structure follows the MAP framework of 
Section~\ref{sec:map}, applied to a joint state 
containing one $SE(3)$ pose and $N$ Euclidean 
per-point states. This is a four-view bundle
adjustment over two stereo pairs
\citep{Haavardsholm2026FullBundleAdjustment}. An interesting observation is that for every $n$'th frame correspondence you can estimate the $n-1$'th order derivative of position. 

\paragraph{Joint State}

The state to be estimated is:
\begin{equation}\label{eq:joint_state}
    \mathbf{x}_{\mathrm{joint}} = \left\langle\,
    \Delta T,\;
    \mathbf{s}_1,\;\ldots,\;\mathbf{s}_N
    \,\right\rangle
\end{equation}
where $\Delta T = T_{B_k,B_{k-1}} \in SE(3)$ is 
the pose change, maintained as a group 
element. Per point:
\begin{itemize}
    \item SS, SM, MS: $\mathbf{s}_i 
    = (\mathbf{p}_i^{B_{k-1}},\,
    \mathbf{v}_i^{\mathbf{p}}) \in \mathbb{R}^6$
    \item MM: $\mathbf{s}_i 
    = (\mathbf{p}_i^{B_{k-1}},\,v_{\perp,i}) 
    \in \mathbb{R}^4$
\end{itemize}
The Gauss-Newton increment 
(Section~\ref{sec:map}) is:
\begin{equation}\label{eq:increment}
    \boldsymbol{\tau} = \begin{pmatrix}
    \boldsymbol{\tau}_\xi \\[1ex] 
    \boldsymbol{\tau}_{s_1} \\[1ex] 
    \vdots \\[1ex] 
    \boldsymbol{\tau}_{s_N}
    \end{pmatrix}
    \in \mathbb{R}^{6+\sum_i\dim(\mathbf{s}_i)}
\end{equation}
where $\boldsymbol{\tau}_\xi \in \mathbb{R}^6$ is 
a fresh tangent vector at each iteration. The update uses $\oplus$ 
appropriate to each component:
\begin{equation}
    \Delta T^{t+1} = \Delta T^t \cdot 
    \mathrm{Exp}(\boldsymbol{\tau}_\xi), \qquad
    \mathbf{s}_i^{t+1} = \mathbf{s}_i^t 
    + \boldsymbol{\tau}_{s_i}
\end{equation}
The $SE(3)$ update composes on the right 
(Section~\ref{sec:lie_algebra}), so 
$\boldsymbol{\tau}_\xi$ is always a small 
perturbation near the identity regardless of 
how far $\Delta T$ is from $\mathcal{E}$. \\

\noindent MM is observable only when the camera centre
moves between frames: 
$\|\mathbf{t}_{\mathrm{base}}^c\| > 0$ is 
required for the cross product in 
Equation~\ref{eq:n_epi} to define a non-degenerate 
epipolar plane. Pure rotation gives 
$\mathbf{t}_{\mathrm{base}}^c = \mathbf{0}$ 
(Section~\ref{sec:d_perp}); a numerical threshold 
on $\|\mathbf{t}_{\mathrm{base}}^c\|$ excludes MM 
points from the joint solver when this fails.

\paragraph{Measurement Model}

Given the joint state, the measurement prediction 
proceeds in two steps: (1) transport the point 
from $B_{k-1}$ to $B_k$ using $\Delta T$; 
(2) project through the camera.

\paragraph{Step 1: Transport.}
For SS/SM/MS (full velocity), the transported 
point is the $SE(3)$ action 
$\Delta T \cdot \mathbf{p}$ with 
$\mathbf{p} = \mathbf{p}_i^{B_{k-1}} 
+ \mathbf{v}_i^{\mathbf{p}}\,\Delta t_{k-1}$:
\begin{equation}\label{eq:transport_joint}
    \mathbf{p}_i^{B_k} 
    = \Delta T \cdot \mathbf{p}
    = \Delta\mathbf{R}\,
    (\mathbf{p}_i^{B_{k-1}} 
    + \mathbf{v}_i^{\mathbf{p}}\,\Delta t_{k-1}) 
    + \Delta\mathbf{t}
\end{equation}
where $\Delta\mathbf{R}$ and $\Delta\mathbf{t}$ 
are the rotation and translation blocks of 
$\Delta T$. For MM (perpendicular velocity only):
\begin{equation}\label{eq:transport_mm}
    \mathbf{p}_i^{B_k} 
    = \Delta\mathbf{R}\,
    (\mathbf{p}_i^{B_{k-1}} 
    + v_{\perp,i}\,\mathbf{d}_\perp\,
    \Delta t_{k-1}) 
    + \Delta\mathbf{t}
\end{equation}
where $\mathbf{d}_\perp$ is the perpendicular 
direction (Section~\ref{sec:d_perp}).

\paragraph{Step 2: Projection.}
Project into camera $c \in \{L,R\}$ 
(Equation~\ref{eq:body_to_pixel}):
\begin{equation}
    h_{i,c}^k = \boldsymbol{\pi}(
    \mathbf{R}_{cB}\,\mathbf{p}_i^{B_k} 
    + \mathbf{t}_c)
\end{equation}
At $k{-}1$, the projection uses 
$\mathbf{p}_i^{B_{k-1}}$ directly (no transport):
\begin{equation}
    h_{i,c}^{k-1} = \boldsymbol{\pi}(
    \mathbf{R}_{cB}\,\mathbf{p}_i^{B_{k-1}} 
    + \mathbf{t}_c)
\end{equation}

\paragraph{Assembled measurement models.}
Stacking the projections for each visibility type:
\begin{align*}
    \textbf{SS:}\; \mathbf{h}_i &= \begin{pmatrix}
    h_{i,L}^{k-1} \\[1ex] h_{i,R}^{k-1} \\[1ex]
    h_{i,L}^{k} \\[1ex] h_{i,R}^{k}
    \end{pmatrix} \in \mathbb{R}^8, \quad
    \textbf{SM:}\; \mathbf{h}_i = \begin{pmatrix}
    h_{i,L}^{k-1} \\[1ex] h_{i,R}^{k-1} \\[1ex] 
    h_{i,c}^{k}
    \end{pmatrix} \in \mathbb{R}^6, \\[1ex]
    \textbf{MS:}\; \mathbf{h}_i &= \begin{pmatrix}
    h_{i,c}^{k-1} \\[1ex] 
    h_{i,L}^{k} \\[1ex] h_{i,R}^{k}
    \end{pmatrix} \in \mathbb{R}^6, \quad
    \textbf{MM:}\; \mathbf{h}_i = \begin{pmatrix}
    h_{i,c}^{k-1} \\[1ex] h_{i,c}^{k}
    \end{pmatrix} \in \mathbb{R}^4
\end{align*}
The corresponding measurements $\mathbf{z}_i$ are 
the tracked pixel coordinates at both frames. 
Measurement noise: 
$\boldsymbol{\Sigma}_{z,i} 
= \sigma_{\mathrm{px}}^2\,\mathbf{I}_{\nu_i}$ 
with $\nu_i$ matching the row count.

\paragraph{Measurement Jacobian}

Following Section~\ref{sec:map}, the Jacobian 
$J_{\mathbf{x}}^{\mathbf{h}_i}$ is the derivative 
of $\mathbf{h}_i$ with respect to the increment 
$\boldsymbol{\tau}$ 
(Equation~\ref{eq:increment}), evaluated at 
$\boldsymbol{\tau} = \mathbf{0}$. For the 
$\mathbb{R}^3$ per-point states this is the 
standard partial derivative. For the $SE(3)$ pose 
it is the manifold Jacobian: the derivative with 
respect to the right perturbation 
$\boldsymbol{\tau}_\xi$ 
\citep[Eq.~41a]{sola2021microlietheorystate}. \\

\noindent For the SS case, the full Jacobian with 
respect to the joint state is:
\begin{gather*}
    \textbf{SS:}\;
    J_{\mathbf{x}}^{\mathbf{h}_i} = 
    \begin{pmatrix}
    \frac{D\,h_{i,L}^{k-1}}
    {D\,\Delta T} & 
    \frac{\partial\,h_{i,L}^{k-1}}
    {\partial\,\mathbf{s}_1} & \cdots & 
    \frac{\partial\,h_{i,L}^{k-1}}
    {\partial\,\mathbf{s}_N} \\[1ex]
    \frac{D\,h_{i,R}^{k-1}}
    {D\,\Delta T} & 
    \frac{\partial\,h_{i,R}^{k-1}}
    {\partial\,\mathbf{s}_1} & \cdots & 
    \frac{\partial\,h_{i,R}^{k-1}}
    {\partial\,\mathbf{s}_N} \\[1ex]
    \frac{D\,h_{i,L}^{k}}
    {D\,\Delta T} & 
    \frac{\partial\,h_{i,L}^{k}}
    {\partial\,\mathbf{s}_1} & \cdots & 
    \frac{\partial\,h_{i,L}^{k}}
    {\partial\,\mathbf{s}_N} \\[1ex]
    \frac{D\,h_{i,R}^{k}}
    {D\,\Delta T} & 
    \frac{\partial\,h_{i,R}^{k}}
    {\partial\,\mathbf{s}_1} & \cdots & 
    \frac{\partial\,h_{i,R}^{k}}
    {\partial\,\mathbf{s}_N}
    \end{pmatrix}
\end{gather*}
where $D / D\,\Delta T$ denotes the 
manifold derivative with respect to the right 
perturbation $\boldsymbol{\tau}_\xi$ at 
$\boldsymbol{\tau}_\xi = \mathbf{0}$, 
and $\partial / \partial\,\mathbf{s}_j$ is the 
standard partial derivative. For other 
correspondence types, remove the corresponding 
row; for MM recall that $\mathbf{h}_i$ is a 
different function. \\

\noindent Only the transported point 
$\mathbf{p}_i^{B_k}$ depends on the pose change, 
and there is no dependence between points, so 
only the column corresponding to point $i$ is 
nonzero:
\begin{gather*}
    \textbf{SS:}\;
    J_{\mathbf{x}}^{\mathbf{h}_i} = 
    \begin{pmatrix}
    \mathbf{0} & \mathbf{0} & \cdots & 
    \mathbf{0} & 
    \frac{\partial\,h_{i,L}^{k-1}}
    {\partial\,\mathbf{s}_i} & 
    \mathbf{0} & \cdots & \mathbf{0} \\[1ex]
    \mathbf{0} & \mathbf{0} & \cdots & 
    \mathbf{0} & 
    \frac{\partial\,h_{i,R}^{k-1}}
    {\partial\,\mathbf{s}_i} & 
    \mathbf{0} & \cdots & \mathbf{0} \\[1ex]
    \frac{D\,h_{i,L}^{k}}
    {D\,\Delta T} & 
    \mathbf{0} & \cdots & \mathbf{0} & 
    \frac{\partial\,h_{i,L}^{k}}
    {\partial\,\mathbf{s}_i} & 
    \mathbf{0} & \cdots & \mathbf{0} \\[1ex]
    \frac{D\,h_{i,R}^{k}}
    {D\,\Delta T} & 
    \mathbf{0} & \cdots & \mathbf{0} & 
    \frac{\partial\,h_{i,R}^{k}}
    {\partial\,\mathbf{s}_i} & 
    \mathbf{0} & \cdots & \mathbf{0}
    \end{pmatrix}
\end{gather*}
We can simplify further since only the transported 
point depends on velocity:
\begin{gather*}
    \textbf{SS:}\;
    J_{\mathbf{x}}^{\mathbf{h}_i} = 
    \begin{pmatrix}
    \mathbf{0} & \mathbf{0} & \cdots & 
    \mathbf{0} & 
    \frac{\partial\,h_{i,L}^{k-1}}
    {\partial\,\mathbf{p}_i^{B_{k-1}}} & 
    \mathbf{0} & \mathbf{0} & \cdots & 
    \mathbf{0} \\[1ex]
    \mathbf{0} & \mathbf{0} & \cdots & 
    \mathbf{0} & 
    \frac{\partial\,h_{i,R}^{k-1}}
    {\partial\,\mathbf{p}_i^{B_{k-1}}} & 
    \mathbf{0} & \mathbf{0} & \cdots & 
    \mathbf{0} \\[1ex]
    \frac{D\,h_{i,L}^{k}}
    {D\,\Delta T} & 
    \mathbf{0} & \cdots & \mathbf{0} & 
    \frac{\partial\,h_{i,L}^{k}}
    {\partial\,\mathbf{p}_i^{B_{k-1}}} & 
    \frac{\partial\,h_{i,L}^{k}}
    {\partial\,\mathbf{v}_i^{\mathbf{p}}} & 
    \mathbf{0} & \cdots & \mathbf{0} \\[1ex]
    \frac{D\,h_{i,R}^{k}}
    {D\,\Delta T} & 
    \mathbf{0} & \cdots & \mathbf{0} & 
    \frac{\partial\,h_{i,R}^{k}}
    {\partial\,\mathbf{p}_i^{B_{k-1}}} & 
    \frac{\partial\,h_{i,R}^{k}}
    {\partial\,\mathbf{v}_i^{\mathbf{p}}} & 
    \mathbf{0} & \cdots & \mathbf{0}
    \end{pmatrix}
\end{gather*}
It remains to find each nonzero block. Let 
$\mathbf{M}_c^{k-1} 
= \frac{\partial\boldsymbol{\pi}}
{\partial\mathbf{q}}\big|_{k-1}\,\mathbf{R}_{cB}$ 
and $\mathbf{M}_c^{k} 
= \frac{\partial\boldsymbol{\pi}}
{\partial\mathbf{q}}\big|_{k}\,\mathbf{R}_{cB}$ 
denote the projection-times-extrinsic Jacobians 
(Section~\ref{sec:measurement}).

\paragraph{Per-point Jacobians w.r.t.\ 
$\mathbf{s}_i$.} At $k{-}1$ (no dependence on 
velocity or pose):
\begin{equation}
    \frac{\partial h_{i,c}^{k-1}}
    {\partial\mathbf{p}_i^{B_{k-1}}} 
    = \mathbf{M}_c^{k-1}, \qquad
    \frac{\partial h_{i,c}^{k-1}}
    {\partial\mathbf{v}_i^{\mathbf{p}}} 
    = \mathbf{0}
\end{equation}
At $k$ (SS/SM/MS), by the chain rule through 
Equation~\ref{eq:transport_joint}:
\begin{equation}\label{eq:Jh_v_stereo}
    \frac{\partial h_{i,c}^k}
    {\partial\mathbf{p}_i^{B_{k-1}}} 
    = \mathbf{M}_c^{k}\,\Delta\hat{\mathbf{R}}, 
    \qquad
    \frac{\partial h_{i,c}^k}
    {\partial\mathbf{v}_i^{\mathbf{p}}} 
    = \mathbf{M}_c^{k}\,\Delta\hat{\mathbf{R}}\,
    \Delta t_{k-1}
\end{equation}
At $k$ (MM), by the chain rule through 
Equation~\ref{eq:transport_mm}:
\begin{equation}
    \frac{\partial h_{i,c}^k}
    {\partial\mathbf{p}_i^{B_{k-1}}} 
    = \mathbf{M}_c^k\,\Delta\hat{\mathbf{R}}, 
    \qquad
    \frac{\partial h_{i,c}^k}
    {\partial v_{\perp,i}} 
    = \mathbf{M}_c^k\,\Delta\hat{\mathbf{R}}\,
    \mathbf{d}_\perp\,\Delta t_{k-1}
\end{equation}

\paragraph{Jacobian w.r.t.\ $\Delta T$.}
The measurement at $k$ is the composition 
$h_{i,c}^k = \boldsymbol{\pi} \circ 
(\mathbf{R}_{cB}\,(\cdot) + \mathbf{t}_c) \circ 
(\Delta T \cdot \mathbf{p})$. By the chain rule, 
the manifold Jacobian with respect to $\Delta T$ 
is the product of the Jacobians of each stage 
\citep[Eq.~41]{sola2021microlietheorystate}:
\begin{equation}\label{eq:Jh_chain}
    J_{\Delta T}^{h_{i,c}^k} 
    = \underbrace{
    \frac{\partial\boldsymbol{\pi}}
    {\partial\mathbf{q}}\bigg|_k
    }_{\text{projection}}
    \cdot\;
    \underbrace{\mathbf{R}_{cB}
    }_{\text{extrinsics}}
    \cdot\;
    \underbrace{J_{\Delta T}^{\Delta T \cdot 
    \mathbf{p}}
    }_{\text{action}}
\end{equation}
The action Jacobian 
$J_{\Delta T}^{\Delta T \cdot \mathbf{p}}$ is the 
derivative of the $SE(3)$ action on 
$\mathbf{p}$ with respect to the right 
perturbation of $\Delta T$ 
\citep[Appendix~D Eq.~182]{sola2021microlietheorystate}:
\begin{equation}\label{eq:J_action}
    J_{\Delta T}^{\Delta T \cdot \mathbf{p}} 
    = \begin{pmatrix}\Delta\hat{\mathbf{R}} 
    & -\Delta\hat{\mathbf{R}}
    [\mathbf{p}]_\times\end{pmatrix}
    \in \mathbb{R}^{3\times6}
\end{equation}
This is defined as 
$\frac{\partial}
{\partial\boldsymbol{\tau}_\xi}
[(\Delta T \cdot 
\mathrm{Exp}(\boldsymbol{\tau}_\xi)) \cdot 
\mathbf{p}]\big|_{\boldsymbol{\tau}_\xi 
= \mathbf{0}}$ --- the differentiation through 
$\mathrm{Exp}$ is already performed in deriving 
Eq.~182; no additional chain rule is needed. 
Since $\Delta T$ acts directly on $\mathbf{p}$ 
(no inversion), there is no inversion Jacobian 
\citep[Eq.~178]{sola2021microlietheorystate} 
unlike the motion-only BA formulation of 
\citep{Haavardsholm2026FullBundleAdjustment}. 
Substituting Equation~\ref{eq:J_action} into 
Equation~\ref{eq:Jh_chain}:
\begin{equation}\label{eq:Jh_xi}
    J_{\Delta T}^{h_{i,c}^k}
    = \mathbf{M}_c^k
    \begin{pmatrix}\Delta\hat{\mathbf{R}} 
    & -\Delta\hat{\mathbf{R}}
    [\mathbf{p}]_\times\end{pmatrix}
\end{equation}
At $k{-}1$: 
$J_{\Delta T}^{h_{i,c}^{k-1}} = \mathbf{0}$ 
(the $k{-}1$ projection does not depend on the 
pose change).

\paragraph{Compact nonzero block for one SS point.}
Extracting the nonzero columns 
($\Delta T$ and $\mathbf{s}_i$ only, 
8 rows $\times$ 12 cols):
\begin{equation}\label{eq:J_SS}
    J_i^{\mathrm{nz}} = \begin{pmatrix}
    \mathbf{0}_{2\times6} 
    & \mathbf{M}_L^{k-1} & \mathbf{0} \\[1ex]
    \mathbf{0}_{2\times6} 
    & \mathbf{M}_R^{k-1} & \mathbf{0} \\[1ex]
    J_{\Delta T}^{h_{i,L}^k}
    & \mathbf{M}_L^k\Delta\hat{\mathbf{R}} 
    & \mathbf{M}_L^k\Delta\hat{\mathbf{R}}
    \Delta t \\[1ex]
    J_{\Delta T}^{h_{i,R}^k}
    & \mathbf{M}_R^k\Delta\hat{\mathbf{R}} 
    & \mathbf{M}_R^k\Delta\hat{\mathbf{R}}
    \Delta t
    \end{pmatrix}
\end{equation}
SM: remove the second-camera row at $k$ 
(6 measurements). MS: remove the second-camera 
row at $k{-}1$ (6 measurements). MM: single 
camera at both frames, velocity column is 
$\mathbf{M}_c^k\Delta\hat{\mathbf{R}}
\mathbf{d}_\perp\Delta t$ (scalar, 
4 measurements).

\paragraph{Perpendicular Direction (MM)}
\label{sec:d_perp}

A two-frame mono correspondence is geometrically 
restricted: velocity in the epipolar plane is 
indistinguishable from a stationary point at a 
different depth. Only the perpendicular component 
is observable. \\

\noindent Compute the baseline vector between 
camera $c$ at $k{-}1$ and camera $c$ at $k$ via 
$T_{c_{k-1},c_k} = T_{cB}\cdot\Delta T^{-1}
\cdot T_{Bc}$, using 
$T_{cB} = [\mathbf{R}_{cB},\,\mathbf{t}_c;\,
\mathbf{0}^\top,\,1]$, 
$T_{Bc} = [\mathbf{R}_{Bc},\,
-\mathbf{R}_{Bc}\mathbf{t}_c;\,
\mathbf{0}^\top,\,1]$, and 
$\Delta T^{-1} = [\Delta\mathbf{R}^\top,\,
-\Delta\mathbf{R}^\top\Delta\mathbf{t};\,
\mathbf{0}^\top,\,1]$ 
\citep[Appendix~D Eq.~170]{sola2021microlietheorystate}. The transform $T_{c_{k-1},c_k}$ maps 
from $c_k$ to $c_{k-1}$; its translation column 
is the $c_k$ origin in $c_{k-1}$ --- the baseline 
from $c_{k-1}$ toward $c_k$, expressed in 
$c_{k-1}$. Computing the $4\times4$ product:
\begin{equation}\label{eq:t_base}
    \mathbf{t}_{\mathrm{base}}^c 
    = \mathbf{t}_c 
    - \mathbf{R}_{cB}\Delta\mathbf{R}^\top
    (\mathbf{R}_{Bc}\mathbf{t}_c 
    + \Delta\mathbf{t})
\end{equation}
Verification: $\Delta\mathbf{R} = \mathbf{I}$, 
$\Delta\mathbf{t} = \mathbf{0}$ gives 
$\mathbf{t}_{\mathrm{base}}^c 
= \mathbf{0}$. $\checkmark$ \\

\noindent For a point with undistorted normalised 
coordinates $(x_n, y_n)$ in camera $c$ at $k{-}1$, 
the ray is $\tilde{\mathbf{x}} 
= (x_n,\,y_n,\,1)^\top$. The ray and baseline 
both live in $c_{k-1}$. Their cross product gives 
the normal to the epipolar plane 
\citep[slide~15]{opsahl2023epipolar}:
\begin{equation}\label{eq:n_epi}
    \mathbf{n}^c = \mathbf{t}_{\mathrm{base}}^c 
    \times \tilde{\mathbf{x}}
\end{equation}
The perpendicular direction in body frame:
\begin{equation}\label{eq:d_perp}
    \mathbf{d}_\perp 
    = \frac{\mathbf{R}_{Bc}\,\mathbf{n}^c}
    {\|\mathbf{n}^c\|}
\end{equation}
Since $\mathbf{d}_\perp$ depends on 
$\Delta T$ (through 
$\Delta\mathbf{R}$ and $\Delta\mathbf{t}$ in 
$\mathbf{t}_{\mathrm{base}}^c$), it is recomputed
at each Gauss-Newton iteration. It is then held
constant within the iteration: the Jacobian above
omits the $\mathbf{d}_\perp$ derivative through
$\Delta T$. The full product rule remains to be
derived and included.

\paragraph{Pose Prior}

The filter-derived relative pose 
$\Delta\hat{T}_{\mathrm{EKF}}$ with covariance 
$\boldsymbol{\Sigma}_{\mathrm{prior}}$ 
(Section~\ref{sec:relative_pose}, 
Equation~\ref{eq:DeltaXi_prior}) enters as a 
partial manifold prior on $\Delta T$ 
(Section~\ref{sec:prior}):
\begin{equation}\label{eq:pose_prior}
    A_0 = \boldsymbol{\Sigma}_{\mathrm{prior}}
    ^{-1/2}
    \begin{pmatrix}\mathbf{I}_6 & \mathbf{0} 
    & \cdots\end{pmatrix}, \quad
    b_0 = \boldsymbol{\Sigma}_{\mathrm{prior}}
    ^{-1/2}\,(\Delta\hat{T}_{\mathrm{EKF}} 
    \ominus \Delta T^t)
\end{equation}
where $\Delta\hat{T}_{\mathrm{EKF}} \ominus 
\Delta T^t = \mathrm{Log}((\Delta T^t)^{-1}
\cdot\Delta\hat{T}_{\mathrm{EKF}})$ is prior 
$\ominus$ estimate 
(Section~\ref{sec:prior}). At 
$\Delta T^t = \Delta\hat{T}_{\mathrm{EKF}}$, 
$b_0 = \mathbf{0}$. $\checkmark$ \\

\noindent The $(\mathbf{I}_6,\,\mathbf{0},\,
\ldots)$ selects only the 
$\boldsymbol{\tau}_\xi$ columns, leaving the 
per-point states unconstrained by the prior. The 
pose prior anchors the 6-DOF gauge freedom of the 
two-view problem, ensuring the Hessian is 
non-singular 
\citep{Haavardsholm2026FullBundleAdjustment}.

\paragraph{Point Priors}
\label{sec:point_priors}

Per-point priors on the stereo-observed position 
regularise the joint solve. SS is observable 
without a prior, but the projection Jacobians 
for $\mathbf{p}$ and $\mathbf{v}^{\mathbf{p}}$ 
at frame $k$ differ only by a factor of 
$\Delta t_{k-1}$ 
(Equation~\ref{eq:Jh_v_stereo}), giving condition 
number $\propto 1/\Delta t_{k-1}$ on the per-point 
block. For SM and MS only one camera observes the
point at one of the two frames. That view
constrains the point in two directions and leaves
the third unconstrained. With
$\mathbf{v}^{\mathbf{p}}$ included the per-point
block is then rank-deficient without further
information. \\

\noindent The prior comes from stereo triangulation (and can come from the previous joint solve) at the 
frame where both cameras observe the point 
(Section~\ref{sec:depth_estimation}). Stereo at 
frame $k{-}1$ (SS, SM) gives 
$(\hat{\mathbf{p}}_i^{B_{k-1}},\,
\boldsymbol{\Sigma}_{p,i}^{B_{k-1}})$ as a direct 
prior on $\mathbf{p}_i^{B_{k-1}}$. Stereo at 
frame $k$ (MS) gives 
$(\hat{\mathbf{p}}_i^{B_k},\,
\boldsymbol{\Sigma}_{p,i}^{B_k})$ as a prior on 
the transported point 
$\mathbf{q}_i = \Delta T\cdot(\mathbf{p}_i^{B_{k-1}} 
+ \mathbf{v}_i^{\mathbf{p}}\,\Delta t_{k-1})$ 
(Equation~\ref{eq:transport_joint}). In principle MM could use a previous-solve prior; currently MM is excluded entirely due to numerical instability. 

\paragraph{Prior at $k{-}1$.}
The state component $\mathbf{p}_i^{B_{k-1}}$ is 
directly part of $\mathbf{s}_i$; the prior reduces 
to a standard partial Gaussian prior 
(Section~\ref{sec:prior}):
\begin{equation}\label{eq:prior_kml}
    A_{p,i} = 
    \boldsymbol{\Sigma}_{p,i}^{-1/2}
    \begin{pmatrix}\mathbf{0}_{3\times 6} & 
    \cdots & \mathbf{S}_{p,i} & 
    \cdots\end{pmatrix}, \quad
    b_{p,i} = \boldsymbol{\Sigma}_{p,i}^{-1/2}\,
    (\hat{\mathbf{p}}_i^{B_{k-1}} 
    - \mathbf{p}_i^{B_{k-1},t})
\end{equation}
where $\mathbf{S}_{p,i}$ selects rows 1{-}3 of 
the $\mathbf{s}_i$ columns.

\paragraph{Prior at $k$.}
The prior is on the transported point 
$\mathbf{q}_i(\mathbf{s}_i,\,\Delta T)$. 
Linearising at the current iterate gives the 
Jacobian blocks (using 
Equation~\ref{eq:Jh_v_stereo} and 
Equation~\ref{eq:J_action}):
\begin{align}
    \frac{\partial\mathbf{q}_i}
    {\partial\mathbf{p}_i^{B_{k-1}}} 
    &= \Delta\hat{\mathbf{R}}, \quad
    \frac{\partial\mathbf{q}_i}
    {\partial\mathbf{v}_i^{\mathbf{p}}} 
    = \Delta\hat{\mathbf{R}}\,\Delta t_{k-1} \\[1ex]
    \frac{D\mathbf{q}_i}{D\,\Delta T} 
    &= \begin{pmatrix}\Delta\hat{\mathbf{R}} & 
    -\Delta\hat{\mathbf{R}}\,
    [\mathbf{p}_i + \mathbf{v}_i^{\mathbf{p}}\,
    \Delta t_{k-1}]_\times\end{pmatrix}
\end{align}
The whitened row stacks these into the pose and 
per-point columns:
\begin{equation}\label{eq:prior_k}
    A_{q,i} = 
    \boldsymbol{\Sigma}_{p,i}^{-1/2}
    \begin{pmatrix}\frac{D\mathbf{q}_i}
    {D\,\Delta T} & \mathbf{0} & \cdots & 
    \Delta\hat{\mathbf{R}} & 
    \Delta\hat{\mathbf{R}}\,\Delta t_{k-1} & 
    \cdots\end{pmatrix}, \quad
    b_{q,i} = \boldsymbol{\Sigma}_{p,i}^{-1/2}\,
    (\hat{\mathbf{p}}_i^{B_k} 
    - \mathbf{q}_i(\mathbf{s}_i^t,\,\Delta T^t))
\end{equation}
This is a linearised manifold prior 
(Section~\ref{sec:prior}, manifold case): $A$ and 
$b$ are rebuilt each Gauss-Newton iteration 
because $\mathbf{q}_i$ depends nonlinearly on 
both $\Delta T$ and $\mathbf{s}_i$. Previous-solve priors on points are deferred to future work; stereo triangulation is currently the only source. MM points have been excluded entirely for now due to numerical instability for which we've found no solution for yet. 

\paragraph{Whitened System and Solution}

Following Section~\ref{sec:map}, whiten each 
per-point measurement block:
\begin{equation}
    A_i = \boldsymbol{\Sigma}_{z,i}^{-1/2}\,
    J_{\mathbf{x}}^{\mathbf{h}_i}, 
    \qquad
    b_i = \boldsymbol{\Sigma}_{z,i}^{-1/2}\,
    (\mathbf{z}_i - \mathbf{h}_i(
    \hat{\mathbf{x}}_{\mathrm{joint}}^t))
\end{equation}
Stack the prior and all point measurements:
\begin{equation}\label{eq:stacked}
    A = \begin{pmatrix}
    A_0 \\[1ex] A_1 \\[1ex] \vdots \\[1ex] A_N
    \end{pmatrix}, \quad
    b = \begin{pmatrix}
    b_0 \\[1ex] b_1 \\[1ex] \vdots \\[1ex] b_N
    \end{pmatrix}
\end{equation}
Solve the normal equations 
(Section~\ref{sec:map}):
\begin{equation}
    (A^\top A)\,\boldsymbol{\tau}^* = A^\top b
\end{equation}
Update:
\begin{equation}
    \Delta T^{t+1} = \Delta T^t\cdot
    \mathrm{Exp}(\boldsymbol{\tau}_\xi^*), \qquad
    \mathbf{s}_i^{t+1} = \mathbf{s}_i^t 
    + \boldsymbol{\tau}_{s_i}^*
\end{equation}
Iterate until 
$\|\boldsymbol{\tau}^*\| < \epsilon_1$ or $\|b\| < \epsilon_2$.

\paragraph{Covariance at convergence.}
\begin{equation}
    \boldsymbol{\Sigma}_{\hat{\mathbf{x}}}
    \approx (A^\top A)^{-1}
\end{equation}
The diagonal blocks give the marginal covariances:
$\boldsymbol{\Sigma}_{\Delta T}^{\mathrm{post}}$ 
($6\times6$) for the pose, and 
$\boldsymbol{\Sigma}_{s_i}$ for each point 
($6\times6$ or $4\times4$).

\paragraph{Initialisation}

$\Delta T^0 = \Delta\hat{T}_{\mathrm{EKF}}$ from 
Section~\ref{sec:relative_pose}. Per-point 
initialisation comes from the point prior or from the search step 
(Section~\ref{sec:search_region}): the input 
sample (depth, velocity) that produced the 
winning LK candidate provides 
$(\hat{\mathbf{p}}_i^0,\,
\hat{\mathbf{v}}_i^{\mathbf{p},0})$.

\paragraph{Transform to $B_k$.}
After convergence, transform position and 
velocity to the current body frame. For 
SS/SM/MS points 
($\mathbf{s}_i = (\mathbf{p}_i,\,
\mathbf{v}_i^{\mathbf{p}}) 
\in \mathbb{R}^6$):
\begin{align}
    \mathbf{p}_i^{B_k} &= 
    \Delta\hat{\mathbf{R}}\,
    (\hat{\mathbf{p}}_i^{B_{k-1}} 
    + \hat{\mathbf{v}}_i^{\mathbf{p}}\,
    \Delta t_{k-1}) 
    + \Delta\hat{\mathbf{t}} \\[1ex]
    \mathbf{v}_i^{\mathbf{p},B_k} 
    &= \Delta\hat{\mathbf{R}}\,
    \hat{\mathbf{v}}_i^{\mathbf{p}}
\end{align}
The Jacobian of this transformation with respect 
to $\mathbf{s}_i$ is:
\begin{equation}\label{eq:T_Delta_SS}
    \mathbf{T}_{\Delta,i} = \begin{pmatrix}
    \Delta\hat{\mathbf{R}} & 
    \Delta\hat{\mathbf{R}}\,\Delta t_{k-1} 
    \\[1ex]
    \mathbf{0} & \Delta\hat{\mathbf{R}}
    \end{pmatrix}
    \in \mathbb{R}^{6\times6}
\end{equation}
The off-diagonal block 
$\Delta\hat{\mathbf{R}}\,\Delta t_{k-1}$ 
captures the velocity-to-position coupling from 
the transport. For MM points 
($\mathbf{s}_i = (\mathbf{p}_i,\,v_{\perp,i}) 
\in \mathbb{R}^4$):
\begin{equation}\label{eq:T_Delta_MM}
    \mathbf{T}_{\Delta,i}^{\mathrm{MM}} 
    = \begin{pmatrix}
    \Delta\hat{\mathbf{R}} & 
    \Delta\hat{\mathbf{R}}\,
    \mathbf{d}_\perp\,\Delta t_{k-1} \\[1ex]
    \mathbf{0}_{1\times3} & 1
    \end{pmatrix}
    \in \mathbb{R}^{4\times4}
\end{equation}
The scalar $v_\perp$ is a magnitude along 
$\mathbf{d}_\perp$ and does not rotate, but 
contributes to position uncertainty through 
$\mathbf{d}_\perp\,\Delta t_{k-1}$.

\paragraph{Covariance transformation.}
The full joint solver covariance 
$\boldsymbol{\Sigma}_{\hat{\mathbf{x}}} 
= (A^\top A)^{-1}$ 
(Section~\ref{sec:joint_solver}) has block 
structure over 
$(\Delta T,\,\mathbf{s}_1,\,\ldots,\,
\mathbf{s}_N)$. The pose perturbation 
$\delta\boldsymbol{\xi}$ lives in the tangent 
space at $\Delta T$ (anchored in $B_{k-1}$) 
and is not transformed. The per-point states 
are transformed by $\mathbf{T}_{\Delta,i}$. 
The full block-diagonal transformation:
\begin{equation}\label{eq:T_block}
    \mathbf{T} = \operatorname{diag}\!\left(
    \mathbf{I}_6,\;
    \mathbf{T}_{\Delta,1},\;\ldots,\;
    \mathbf{T}_{\Delta,N}
    \right)
\end{equation}
Applied via 
(Equation~\ref{eq:cov_linear}):
\begin{equation}\label{eq:Sigma_Bk}
    \boldsymbol{\Sigma}^{B_k} 
    = \mathbf{T}\,
    \boldsymbol{\Sigma}_{\hat{\mathbf{x}}}\,
    \mathbf{T}^\top
\end{equation}
This leaves
$\boldsymbol{\Sigma}_{\xi\xi}$ unchanged,
transforms per-point blocks
$\boldsymbol{\Sigma}_{s_is_j}^{B_k}
= \mathbf{T}_{\Delta,i}\,
\boldsymbol{\Sigma}_{s_is_j}\,
\mathbf{T}_{\Delta,j}^\top$, and transforms
the pose--point cross-covariances on the point
side only:
$\boldsymbol{\Sigma}_{\xi s_i}^{B_k}
= \boldsymbol{\Sigma}_{\xi s_i}\,
\mathbf{T}_{\Delta,i}^\top$. \\

\noindent This is an approximation. It understates
the point covariance by the pose-uncertainty
contribution.
\newpage
\section{Measurement Pipeline}
\label{sec:pipeline}
\subsection{Computer Vision Primitives}
\label{sec:cv}

\subsubsection{Feature Detection: FAST}

FAST (Features from Accelerated Segment Test) 
\citep{rosten2006machine} examines a circle of 16 pixels 
around each candidate. If a contiguous arc of $N$ pixels 
(typically 9 or 12) are all brighter or darker than the 
centre by a threshold, it is a corner.

\subsubsection{Feature Quality: Shi-Tomasi}

The Shi-Tomasi cornerness score \citep{shi1994good} is 
$\min(\lambda_1, \lambda_2)$ where $\lambda_1, \lambda_2$ 
are eigenvalues of the structure tensor 
$\mathbf{M} = \sum_{\text{patch}} 
\nabla I\,\nabla I^\top$. A high score means strong 
gradients in two independent directions --- the patch is 
uniquely localised and suitable for tracking. \\

\noindent A minimum score $\tau_{\min}$ rejects unreliable 
detections. In featureless or overexposed scenes, the 
pool may be empty and this is the correct behaviour.

\subsubsection{Temporal Tracking: Lucas-Kanade}

LK \citep{lucas1981iterative} (pyramidal implementation 
\citep{bouguet2000pyramidal}) finds the displacement 
$\delta\mathbf{u}$ that minimises 
$\sum_{\text{patch}} [I_k(\mathbf{u} + \delta\mathbf{u}) 
- I_{k-1}(\mathbf{u})]^2$. Linearising gives
\citep[\S9.1.3, eqs.~9.26--9.35, pp.~566--567]{szeliski2010computer}:
\begin{equation}
    \mathbf{M}\,\delta\mathbf{u} 
    = -\sum_{\text{patch}} (I_k - I_{k-1})\,\nabla I_k
\end{equation}
where $\mathbf{M}$ is the same structure tensor. The 
system is well-conditioned when both eigenvalues are 
large (Shi-Tomasi score high). Pyramidal LK handles
large displacements by solving coarse-to-fine. We use
OpenCV's \texttt{calcOpticalFlowPyrLK} (4.13.0), which
iterates the update at each pyramid level until the
displacement change falls below a tolerance or an
iteration cap is reached; see
\href{https://github.com/danielftg/blackbird-vio/blob/a8713f352ee5481de798098e39798ac93115d479/src/modules/vision.py}{vision.py}.

\paragraph{Forward-backward check.}
Track forward $k{-}1 \to k$ to get $\mathbf{u}'$, then 
backward $k \to k{-}1$ from $\mathbf{u}'$ to get 
$\mathbf{u}''$. If 
$\|\mathbf{u}'' - \mathbf{u}\| > \epsilon_{\text{fb}}$, 
the match is inconsistent and the point is dropped. This 
is our primary failure detector.

\subsubsection{Stereo Matching: NCC}
Normalised Cross-Correlation
\citep[\S9.1, eq.~9.11, p.~561]{szeliski2010computer}
matches a patch from one image along the epipolar line in 
the other:
\begin{equation}
    \text{NCC}(\mathbf{u}_L, \mathbf{u}_R) 
    = \frac{\sum(I_L - \bar{I}_L)(I_R - \bar{I}_R)}
    {\sqrt{\sum(I_L - \bar{I}_L)^2\,
    \sum(I_R - \bar{I}_R)^2}}
\end{equation}
Invariant to brightness and contrast differences between 
cameras. For rectified stereo the epipolar line is a 
horizontal row, so the search reduces to 1D over the 
disparity range $[d_{\min}, d_{\max}]$. The location of 
the highest NCC value, provided it exceeds threshold 
$\text{NCC}_{\min}$, is the match. Sub-pixel refinement by 
parabolic fitting on the three response values around the 
peak: 
\begin{equation}
    \Delta u = \frac{1}{2}
    \frac{y_{-1} - y_{+1}}{y_{-1} - 2y_0 + y_{+1}}, \qquad
    u^* = u_{\text{peak}} + \Delta u
\end{equation}
clamped to $[-1, 1]$. Boundary peaks skip refinement.

\subsubsection{Grid Selection and Focus Weighting}
The image is divided into a $\text{cols}\times\text{rows}$ 
grid. Both selection routines first filter the candidate 
pool against the pixels of currently-tracked points, 
discarding any candidate closer than $d_{\min}^{\text{px}}$ 
to an existing one.

\paragraph{Feature point allocation.} The top 
$N_{\text{feat},gl}$ candidates by Shi-Tomasi cornerness 
are picked globally. The remaining candidates are binned 
into the grid, and the top $N_{\text{feat},lo}$ per cell 
are added. The grid is purely a spatial spread mechanism 
and need not match the interest-point grid.\\

\noindent Feature replenishment ought to be weighted by 
expected remaining cell lifetime (from estimated velocity 
and angular velocity): cells the quadcopter is moving toward 
get more features. This also avoids spending compute on 
points that fall out of the image at the next time step 
anyway. Feature points should remain in frame for at least 
3 frames, so 50ms at 60Hz frame rate. Ideally more.

\paragraph{Interest point allocation.}
The focus point $\mathbf{F}_{k-1} \in \mathbb{R}^3$ 
projects into each camera as 
$\mathbf{f}_c = \boldsymbol{\pi}(\mathbf{R}_{cB}\,
\mathbf{F}_{k-1} + \mathbf{t}_c)$, $c \in \{L, R\}$. 
A 2D isotropic Gaussian 
$\mathcal{N}(\mathbf{f}_c,\,\sigma_F^2\mathbf{I})$ 
defined over the image plane assigns a probability mass to 
each grid cell. Isotropy makes the two axes independent
\citep[\S5.5, p.~331]{devore2021modern}, so the
joint mass factors:
\begin{equation}
    P(\text{cell}_{r,c}) 
    = \frac{P(x \in [x_c, x_{c+1}])\,
            P(y \in [y_r, y_{r+1}])}
           {Z_x \cdot Z_y}
\end{equation}
where each one-dimensional probability is a difference of
normal CDFs \citep[\S4.3, p.~217--219]{devore2021modern}, and
$Z_x = P(0 \le x \le W),\ Z_y = P(0 \le y \le H)$ 
re-normalise the distribution over the image rectangle.

\noindent Each camera receives $N_{\mathcal{I}}^{\max}/2$ 
points, allocated 
$\lfloor (N_{\mathcal{I}}^{\max}/2)\cdot 
P(\text{cell}_{r,c}) \rfloor$ per cell, with the top 
candidates per cell by Shi-Tomasi cornerness chosen. Any 
residual from the floor operation is distributed in 
weight-descending order, peak cell first, then outward. 
Cells that cannot fulfil their allocation (too few 
candidates) overflow their residual into the same 
descending order. Smaller $\sigma_F$ concentrates points 
near the focus; larger $\sigma_F$ spreads them. NCC 
stereo matching then attempts to pair each interest point 
across cameras; successful matches contribute stereo 
interest points, failures contribute mono.
\newpage
\subsection{Geometric Estimation}
\label{sec:depth_estimation}

\paragraph{Stereo Depth}

From a stereo match with pixel coordinates 
$(u_L, v_L)$ in the left camera and $(u_R, v_R)$ in 
the right camera, first undistort to normalised 
coordinates:
\begin{equation}
    (x_n, y_n) = D^{-1}\!\left(
    \kappa^{-1}(u_L, v_L)\right)
\end{equation}
Compute depth from disparity and reconstruct the 3D 
point in the left camera frame:
\begin{equation}
    Z = \frac{f_u\,b}{u_L - u_R}, \qquad
    \mathbf{p}^c = Z
    \begin{pmatrix} x_n \\ y_n \\ 1 \end{pmatrix}
\end{equation}
Taking the disparity in raw pixel coordinates
presumes a rectified pair. Our input is rectified.
$D^{-1}$ is then the identity and
$(x_n, y_n) = \kappa^{-1}(u_L, v_L)$. We keep
$D^{-1}$ above and its Jacobian below for the
unrectified case. There the disparity must be taken
on undistorted coordinates as well. \\

\noindent Transform to body frame:
\begin{equation}
    \mathbf{p}^B = \mathbf{R}_{BL}(
    \mathbf{p}^c - \mathbf{t}_L)
\end{equation}

\paragraph{Covariance Initialisation}
\label{sec:covariance_init}

The covariance of the reconstructed 3D position is
obtained by first-order uncertainty propagation
(Equation~\ref{eq:cov_prop}).

\noindent The raw measurements are 
$(u_L, v_L, u_R)$ --- three independent pixel 
coordinates, each with noise 
$\sigma_{\text{px}}$. Note that $u_L$ affects 
both the normalised coordinates $(x_n, y_n)$ and 
the depth $Z = f_u b / (u_L - u_R)$, so these are 
correlated and cannot be treated independently. \\

\noindent We derive 
$\partial\mathbf{p}^c / \partial(u_L, v_L, u_R)$ 
using the product rule on 
$\mathbf{p}^c = Z \cdot (x_n, y_n, 1)^\top$.

\paragraph{Intermediate quantities.}
Depth derivatives:
\begin{equation}
    \frac{\partial Z}{\partial u_L} 
    = -\frac{f_u b}{(u_L - u_R)^2} 
    = -\frac{Z}{d}, 
    \quad
    \frac{\partial Z}{\partial v_L} = 0, \quad
    \frac{\partial Z}{\partial u_R} = \frac{Z}{d}
\end{equation}
where $d = u_L - u_R$.

Normalised coordinate derivatives (chain rule through 
undistortion):
\begin{equation}
    \frac{\partial(x_n, y_n)}{\partial(u_L, v_L)} 
    = \underbrace{\frac{\partial D^{-1}}
    {\partial(x_d, y_d)}}_{\mathbf{J}_{D^{-1}}} 
    \cdot 
    \underbrace{\frac{\partial \kappa^{-1}}
    {\partial(u, v)}}_{\mathbf{K}_{2\times 2}^{-1}}
    \triangleq \mathbf{C} 
    = \begin{pmatrix} C_{11} & C_{12} \\ 
    C_{21} & C_{22} \end{pmatrix}
\end{equation}
where $\kappa^{-1}$: $x_d = (u - c_u)/f_u$, 
$y_d = (v - c_v)/f_v$ gives 
$\mathbf{K}_{2\times 2}^{-1} 
= \operatorname{diag}(1/f_u, 1/f_v)$, and 
$\mathbf{J}_{D^{-1}} 
= \partial D^{-1}/\partial(x_d, y_d) 
\in \mathbb{R}^{2\times 2}$ is the Jacobian of the 
undistortion model (model-dependent; identity if no 
distortion). $(x_n, y_n)$ do not depend on $u_R$.

\paragraph{Product rule.}
Since $p_x^c = Z \cdot x_n$, 
$p_y^c = Z \cdot y_n$, $p_z^c = Z$:
\begin{equation}
    \frac{\partial(Z \cdot x_n)}{\partial(\cdot)} 
    = \frac{\partial Z}{\partial(\cdot)}\,x_n 
    + Z\,\frac{\partial x_n}{\partial(\cdot)}
\end{equation}

\paragraph{Row 1: $p_x^c = Z\,x_n$.}
\begin{align}
    \frac{\partial p_x^c}{\partial u_L} 
    &= Z\!\left(C_{11} - \frac{x_n}{d}\right) \\
    \frac{\partial p_x^c}{\partial v_L} 
    &= Z\,C_{12} \\
    \frac{\partial p_x^c}{\partial u_R} 
    &= \frac{Z\,x_n}{d}
\end{align}

\paragraph{Row 2: $p_y^c = Z\,y_n$.}
\begin{align}
    \frac{\partial p_y^c}{\partial u_L} 
    &= Z\!\left(C_{21} - \frac{y_n}{d}\right) \\
    \frac{\partial p_y^c}{\partial v_L} 
    &= Z\,C_{22} \\
    \frac{\partial p_y^c}{\partial u_R} 
    &= \frac{Z\,y_n}{d}
\end{align}

\paragraph{Row 3: $p_z^c = Z$.}
\begin{align}
    \frac{\partial p_z^c}{\partial u_L} 
    &= -\frac{Z}{d}, \quad
    \frac{\partial p_z^c}{\partial v_L} = 0, \quad
    \frac{\partial p_z^c}{\partial u_R} 
    = \frac{Z}{d}
\end{align}

\paragraph{Assembled Jacobian.}
\begin{equation}\label{eq:Jpc_stereo}
    \frac{\partial\mathbf{p}^c}
    {\partial(u_L, v_L, u_R)} 
    = Z \begin{pmatrix}
    C_{11} - x_n/d & C_{12} & x_n/d \\
    C_{21} - y_n/d & C_{22} & y_n/d \\
    -1/d & 0 & 1/d
    \end{pmatrix}
\end{equation}

\paragraph{Camera-frame covariance.}
\begin{equation}\label{eq:Sigma_pc_stereo}
    \boldsymbol{\Sigma}_{p^c} 
    = \frac{\partial\mathbf{p}^c}
    {\partial(u_L, v_L, u_R)}\,
    \sigma_{\text{px}}^2\,\mathbf{I}_3\,
    \frac{\partial\mathbf{p}^c}
    {\partial(u_L, v_L, u_R)}^\top
\end{equation}

\paragraph{Body-frame covariance.}
\begin{equation}\label{eq:Sigma_pB}
    \boldsymbol{\Sigma}_{p^B} 
    = \mathbf{R}_{Bc}\,
    \boldsymbol{\Sigma}_{p^c}\,
    \mathbf{R}_{Bc}^\top
\end{equation}

\paragraph{Mono features.}
Initial covariance and point come from the
point-velocity solver
(Section~\ref{sec:joint_solver}).

\subsection{Search Region Construction}
\label{sec:search_region}

For filter feature points, the search region is the
projection of the $\chi^2_{\alpha,3}$ ellipsoid of
the filter's body-frame position estimate
$\hat{\mathbf{p}}_i^B$ and its $3\times3$ block of
$\mathbf{P}_k^-$. For all
other points (interest and pre-admission features),
the search region is constructed from the prediction 
function and the uncertainty in its inputs.

\paragraph{Prediction function.}
The predicted pixel position at frame $k$ is a function of pose change, position, and velocity:
\begin{equation}\label{eq:u_pred}
    \hat{\mathbf{u}}_k(\Delta T,\,
    \mathbf{p}_i^{B_{k-1}},\,
    \mathbf{v}_i^{\mathbf{p}}) 
    = \boldsymbol{\pi}\!\left(
    \mathbf{R}_{cB}\left(\Delta\mathbf{R}\,
    (\mathbf{p}_i^{B_{k-1}} 
    + \mathbf{v}_i^{\mathbf{p}}\,\Delta t_{k-1}) 
    + \Delta\mathbf{t}\right) 
    + \mathbf{t}_c\right)
\end{equation}

\paragraph{Uncertainty sets.}
The EKF-derived relative pose 
$\Delta\hat{T},\,
\boldsymbol{\Sigma}_{\Delta\xi}$ 
(Section~\ref{sec:relative_pose}) is always 
available. Position and velocity may or may not be 
available from previous iterations. When an estimate 
with covariance exists, the uncertainty set is the 
$n\sigma$ ellipsoid. When no estimate exists, the 
uncertainty set is a physically bounded domain. \\

\noindent \textit{Statistical sets} ($n\sigma$ 
ellipsoids at confidence level $\alpha$):
\begin{align}
    \mathcal{E}_{\boldsymbol{\xi}} &= \left\{
    \Delta\hat{T} \circ 
    \operatorname{Exp}(\boldsymbol{\xi})
    \;\middle|\;
    \boldsymbol{\xi}^\top\,
    \boldsymbol{\Sigma}_{\Delta\xi}^{-1}\,
    \boldsymbol{\xi} 
    \leq \chi^2_{\alpha,6}
    \right\} \\[1ex]
    \mathcal{E}_{\mathbf{p}} &= \left\{
    \mathbf{p}
    \;\middle|\;
    (\mathbf{p} - \hat{\mathbf{p}}_i)^\top\,
    \boldsymbol{\Sigma}_{pp}^{-1}\,
    (\mathbf{p} - \hat{\mathbf{p}}_i) 
    \leq \chi^2_{\alpha,3}
    \right\} \\[1ex]
    \mathcal{E}_{\mathbf{p},\mathbf{v}} &= \left\{
    \begin{pmatrix} \mathbf{p} \\ \mathbf{v} 
    \end{pmatrix}
    \;\middle|\;
    \left(
    \begin{pmatrix} \mathbf{p} \\ \mathbf{v} 
    \end{pmatrix}
    - \begin{pmatrix} \hat{\mathbf{p}}_i \\ 
    \hat{\mathbf{v}}_i^{\mathbf{p}} \end{pmatrix}
    \right)^\top
    \boldsymbol{\Sigma}_i^{-1}
    \left(
    \begin{pmatrix} \mathbf{p} \\ \mathbf{v} 
    \end{pmatrix}
    - \begin{pmatrix} \hat{\mathbf{p}}_i \\ 
    \hat{\mathbf{v}}_i^{\mathbf{p}} \end{pmatrix}
    \right)
    \leq \chi^2_{\alpha,6}
    \right\}
\end{align}

\noindent \textit{Bounded sets} (physical limits, 
used when no estimate exists):
\begin{align}
    \mathcal{S}_Z &= \left\{
    Z\,(x_n,\, y_n,\, 1)^\top
    \;\middle|\;
    Z \in [Z_{\min},\, Z_{\max}]\right\}
    & &\text{Back-projected depth range
    (\S\ref{sec:depth_estimation})} \\
    \mathcal{S}_{\mathbf{v}} &= \{\mathbf{v} 
    \in \mathbb{R}^3 \mid 
    \|\mathbf{v}\| \leq v_{\max}\} 
    & &\text{Velocity bound} \\
    \mathcal{S}_{\mathbf{a}} &= \{\mathbf{a} 
    \in \mathbb{R}^3 \mid 
    \|\mathbf{a}\| \leq a_{\max}\} 
    & &\text{Acceleration bound}
\end{align}

\noindent \textit{Acceleration expansion.} For 
points with a velocity estimate, the velocity may 
have changed since the last frame:
\begin{equation}
    \mathcal{E}_{\mathbf{p},\mathbf{v}} 
    \uplus \mathcal{S}_{\mathbf{a}} 
    = \left\{
    (\mathbf{p},\; \mathbf{v} 
    + \mathbf{a}\,\Delta t_{k-1})
    \;\middle|\;
    (\mathbf{p}, \mathbf{v}) 
    \in \mathcal{E}_{\mathbf{p},\mathbf{v}},\;
    \mathbf{a} \in \mathcal{S}_{\mathbf{a}}
    \right\}
\end{equation}

\paragraph{Search regions by available information.}
In each case, the search region is 
$\hat{\mathbf{u}}_k[\cdot] \cap [0,w]\times[0,h]$:
\begin{center}
\renewcommand{\arraystretch}{1.4}
\begin{tabular}{ll}
\textbf{Available} & \textbf{Input domain} \\
\hline
$\hat{\mathbf{p}}_i$, 
$\hat{\mathbf{v}}_i^{\mathbf{p}}$ 
    & $\hat{\mathbf{u}}_k\!\left[
    \mathcal{E}_{\boldsymbol{\xi}} \times
    (\mathcal{E}_{\mathbf{p},\mathbf{v}} 
    \uplus \mathcal{S}_{\mathbf{a}})
    \right]$ \\[1ex]
$\hat{\mathbf{p}}_i$ only 
    & $\hat{\mathbf{u}}_k\!\left[
    \mathcal{E}_{\boldsymbol{\xi}} \times
    \mathcal{E}_{\mathbf{p}} \times
    \mathcal{S}_{\mathbf{v}}
    \right]$ \\[1ex]
Neither 
    & $\hat{\mathbf{u}}_k\!\left[
    \mathcal{E}_{\boldsymbol{\xi}} \times
    \mathcal{S}_Z \times
    \mathcal{S}_{\mathbf{v}}
    \right]$ \\
\end{tabular}
\end{center}

\paragraph{Temporal matching protocol.}
The search region is a 2D subset of the pixel plane. 
Matching proceeds as follows:
\begin{enumerate}
    \item \textit{Discretise input domain.} Sample 
          the uncertainty sets above. Each sample 
          specifies a candidate 
          $(\Delta T,\,\mathbf{p}_i,\,
          \mathbf{v}_i^{\mathbf{p}})$.
    \item \textit{Project.} Pass each sample through 
          $\hat{\mathbf{u}}_k$ 
          (Equation~\ref{eq:u_pred}) to obtain 
          candidate pixels. Discard those outside
          $[0,w]\times[0,h]$ and deduplicate on
          integer pixel coordinates. The surviving
          candidates are the search region. Their
          convex hull bounds it if a contiguous
          region is wanted. If no candidate
          survives, do not attempt a match.
    \item \textit{Coarse match.} Evaluate SSD between 
          the reference patch at 
          $\mathbf{u}_i^{k-1}$ and the candidate 
          pixels. Select the candidate with lowest 
          SSD.
    \item \textit{Fine match.} Run pyramidal LK from 
          the coarse winner to obtain sub-pixel 
          accuracy.
    \item \textit{Validate.} Forward-backward check: 
          run LK backward from the result; if 
          $\|\mathbf{u}'' - \mathbf{u}_i^{k-1}\| 
          > \epsilon_{\text{fb}}$, reject.
    \item \textit{Solver initialisation.} The input 
          sample 
          $(\Delta T,\,\mathbf{p}_i,\,
          \mathbf{v}_i^{\mathbf{p}})$ that produced 
          the winning candidate provides the 
          initial estimate 
          $(\hat{\mathbf{p}}_i^0,\,
          \hat{\mathbf{v}}_i^{\mathbf{p},0})$ for 
          the joint solver 
          (Section~\ref{sec:joint_solver}). Point data is overwritten by stereo triangulation when available. 
\end{enumerate}
SSD rather than NCC is sufficient for temporal 
matching: the reference and target are from the same 
camera separated by one frame interval, so brightness 
and contrast are effectively unchanged.
\subsection{Stationarity Testing}
\label{sec:statistics}

\subsubsection{Feature Movement Detection}
\label{sec:feature_movement}

\paragraph{Individual gate.}
The EKF predicts measurement 
$\hat{\mathbf{z}}_{k,i} = h_i(\hat{\mathbf{x}}_k^-)$. 
The innovation:
\begin{equation}
    \mathbf{y}_{k,i} = \mathbf{z}_{k,i} 
    - h_i(\hat{\mathbf{x}}_k^-)
\end{equation}
Under the stationarity hypothesis, 
$\mathbf{y}_{k,i} \sim 
\mathcal{N}(\mathbf{0}, \mathbf{S}_{k,i})$, where 
the innovation covariance combines state uncertainty 
projected into measurement space with sensor noise 
\citep[p.~236]{barshalom2001estimation}:
\begin{equation}
    \mathbf{S}_{k,i} 
    = \mathbf{H}_{k,i}\,\mathbf{P}_k^-\,
    \mathbf{H}_{k,i}^\top + \mathbf{R}_{k,i}
\end{equation}
The normalised innovation squared (NIS):
\begin{equation}
    \gamma_{k,i} = \mathbf{y}_{k,i}^\top\,
    \mathbf{S}_{k,i}^{-1}\,\mathbf{y}_{k,i} 
    \sim \chi^2_{\nu_i}
\end{equation}
where $\nu_i = 4$ (stereo) or $\nu_i = 2$ (mono). If 
$\gamma_{k,i} > \chi^2_{\alpha,\,\nu_i}$, reject 
stationarity.

\paragraph{Joint consistency check.}
After individual gating, test the remaining inlier set 
$\mathcal{S}$ for collective consistency 
\citep[p.~236]{barshalom2001estimation}:
\begin{equation}
    \gamma_\text{joint} 
    = \sum_{i \in \mathcal{S}} \gamma_{k,i} 
    \sim \chi^2_{\nu_\text{total}}, \qquad 
    \nu_\text{total} 
    = \sum_{i \in \mathcal{S}} \nu_i
\end{equation}
If $\gamma_\text{joint}
> \chi^2_{\alpha,\,\nu_\text{total}}$: the inlier
set is inconsistent. Exact $\chi^2$ behaviour
assumes the innovations are independent. The sum is
otherwise an approximation.

\subsubsection{Feature Admission}
\label{sec:admission_stats}

Before a feature point enters the EKF, it must pass a stationary test which depends on the correspondence type. For MS/SS/SM
correspondences:
\begin{equation}
    \gamma_v = (\hat{\mathbf{v}}_i^{\mathbf{p}})^\top\,
    \boldsymbol{\Sigma}_{vv}^{-1}\,
    \hat{\mathbf{v}}_i^{\mathbf{p}} 
    \sim \chi^2_3
\end{equation}
For MM correspondences:
\begin{equation}
    \gamma_v = \frac{\hat{v}_{\perp,i}^2}
    {\sigma_{v_\perp}^2} \sim \chi^2_1
\end{equation}
If $\gamma_v$ exceeds the threshold: do not admit. The NIS gate ejects features later if they were in fact moving. Points that pass are augmented into the EKF (Section~\ref{sec:augment}) with covariance from the point-velocity solve
$(A^\top A)^{-1}$ (Section~\ref{sec:joint_solver}).

\paragraph{Limitation.}
For MM, motion along the epipolar plane is 
undetectable by this test (absorbed into the position 
estimate). The NIS gate 
(Section~\ref{sec:feature_movement}) catches 
such points one frame after EKF admission, since the 
epipolar geometry changes between frames.

\subsubsection{Stationarity Assumption}

The dynamics $\dot{\mathbf{p}}_i^B 
= -\mathbf{v} 
- \boldsymbol{\omega} \times \mathbf{p}_i^B$ 
(Equation~\ref{eq:pdot}) assume the point is 
fixed in the world. If the point is moving, the 
propagated position is wrong, the predicted pixel 
position is wrong, and the innovation is large. 
The chi-squared test 
(Section~\ref{sec:statistics}) detects this and 
excludes the point from the update.

\newpage
\section{Algorithm Overview}
\label{sec:algorithm_overview}

This section gives a conceptual map of the 
estimator. The code is the source of truth: 
\href{https://github.com/danielftg/blackbird-vio/blob/a8713f352ee5481de798098e39798ac93115d479/src/modules/algo.py}{algo.py} 
contains the orchestration; tuning parameters are 
maintained in 
\href{https://github.com/danielftg/blackbird-vio/blob/a8713f352ee5481de798098e39798ac93115d479/src/constants/algorithm.yaml}{algorithm.yaml}; 
Point-Set and \texttt{Point} datatypes are in 
\href{https://github.com/danielftg/blackbird-vio/blob/a8713f352ee5481de798098e39798ac93115d479/src/modules/points.py}{points.py}; 
the joint solver lives in 
\href{https://github.com/danielftg/blackbird-vio/blob/a8713f352ee5481de798098e39798ac93115d479/src/modules/solver.py}{solver.py}; 
the EKF in 
\href{https://github.com/danielftg/blackbird-vio/blob/a8713f352ee5481de798098e39798ac93115d479/src/modules/ekf.py}{ekf.py}.

\subsection{Point Sets and Roles}
Tracked points are partitioned by role:
\begin{itemize}
    \item $\mathcal{F}$: feature points 
    augmented into the EKF state. Stationary, 
    used as inertial references. Their 
    measurement information enters via the pose 
    update; they do not participate in the joint 
    solver.
    \item $\mathcal{F}_{\mathrm{pre}}$: candidate 
    feature points awaiting their first 
    two-frame solve and admission test. Enter 
    the joint solver.
    \item $\mathcal{I}$: interest points steered 
    by the focus. Not in the EKF. Enter the 
    joint solver.
\end{itemize}

\noindent Each \texttt{Point} carries its pixel 
observations at $k{-}1$ and $k$ (stereo or mono 
per camera), its 3D state and covariance in 
$B_{k-1}$ from the previous solve and in $B_k$ 
from the current solve, plus an id and age. The 
populated fields determine its stage (one-frame 
vs two-frame), correspondence type (SS/SM/MS/MM, 
see \S\ref{sec:joint_solver}), and search 
strategy (\S\ref{sec:search_region}).

\subsection{Accumulator}
The accumulator persists across iterations: the 
three point sets, the EKF posterior 
$(\hat{\mathcal{X}}_{k-1}^+,\,\mathbf{P}_{k-1}^+)$, 
the previous image pair, the focus, and the last 
timestamp. At the end of each iteration $k$ is 
pushed into $k{-}1$.

\subsection{Input and Output}
\label{sec:alg_io}

\paragraph{Input.} Per frame: stereo pair 
$(L_k, R_k)$, timestamp $t_k$, control input 
$\mathbf{u}_{k-1}$, focus point 
$\mathbf{F}_{k-1}$ and spread $\sigma_{F,k-1}$.

\paragraph{Output.} 
\begin{itemize}
    \item Core EKF state 
    $\hat{\mathcal{X}}_k^{\mathrm{core}} 
    = \langle T,\mathbf{v},\boldsymbol{\omega},
    \mathbf{g}^B,\mathbf{d}^B\rangle$ with 
    $\mathbf{P}_k^{\mathrm{core}} 
    \in \mathbb{R}^{18\times 18}$. Feature rows 
    and columns are stripped from the EKF state 
    and covariance before output; internally the 
    EKF retains them.
    \item Solver pose change 
    $\Delta\hat{T}_{\mathrm{solver}}$ and 
    $\boldsymbol{\Sigma}_{\Delta\xi}$ from the 
    joint solver. This is the recommended 
    transform for propagating the downstream 
    world model between frames.
    \item Point cloud $\mathcal{C}_k$ tagged by 
    role and stage. Stage-2 points contribute 
    $(\hat{\mathbf{p}}_i^{B_k},
    \hat{\mathbf{v}}_i^{B_k},
    \boldsymbol{\Sigma}_i^{B_k})$ from the joint 
    solver; stage-1 (stereo-only) points 
    contribute $(\hat{\mathbf{p}}_i,
    \boldsymbol{\Sigma}_{p,i})$ from stereo 
    triangulation; feature points contribute 
    position and covariance from the EKF state.
\end{itemize}

\subsection{Pseudocode}

\paragraph{Initialise ($k = 0$).}~
\begin{algorithmic}[1]
    \State Receive $(L_0,\,R_0,\,t_0,\,
           \mathbf{F}_{-1},\,\sigma_{F,-1})$
    \State FAST + Shi-Tomasi on $L_0,\,R_0$ 
           \hfill (\S\ref{sec:cv})
    \State Focus-weighted grid selection into 
           $\mathcal{I}$; grid selection into 
           $\mathcal{F}_{\mathrm{pre}}$ 
           \hfill (\S\ref{sec:cv})
    \State NCC stereo match each point; 
           stage-1 stereo triangulation for 
           stereo pairs gives 
           $(\hat{\mathbf{p}}_i,\,
           \boldsymbol{\Sigma}_{p,i})$ 
           \hfill (\S\ref{sec:depth_estimation})
    \State Initialise EKF: 
           $(\hat{\mathcal{X}}_0,\,
           \mathbf{P}_0)$ with no feature 
           points 
           \hfill (\S\ref{sec:system_model})
    \State Output 
           $(\hat{\mathcal{X}}_0,\,
           \mathbf{P}_0,\,\mathcal{C}_0)$; 
           push accumulator $k \to k{-}1$
\end{algorithmic}

\paragraph{Iterate ($k = 1, 2, \ldots$).}~
\begin{algorithmic}[1]
    \Statex \textbf{--- Receive and propagate ---}
    \State Receive $(L_k,\,R_k,\,t_k,\,
           \mathbf{u}_{k-1},\,\mathbf{F}_{k-1},\,
           \sigma_{F,k-1})$; 
           $\Delta t = t_k - t_{k-1}$
    \State Propagate 
           $(\hat{\mathcal{X}}_{k-1}^+,\,
           \mathbf{P}_{k-1}^+) \to 
           (\hat{\mathcal{X}}_k^-,\,
           \mathbf{P}_k^-)$ 
           \hfill (\S\ref{sec:ekf_eqs}, 
           \S\ref{sec:system_model})
    \State FAST + Shi-Tomasi on $L_k,\,R_k$; 
           store keypoint pool 
           \hfill (\S\ref{sec:cv})

    \Statex
    \Statex \textbf{--- Pre-update relative 
           pose ---}
    \State Compute $(\Delta\hat{T}^-,\,
           \boldsymbol{\Sigma}_{\Delta\xi}^-)$ 
           from $\hat{T}_{k-1}^+,\,\hat{T}_k^-$ 
           \hfill (\S\ref{sec:relative_pose})

    \Statex
    \Statex \textbf{--- Search and track ---}
    \State For each point with 
           $\mathbf{u}_i^{k-1}$: build 
           candidate set; SSD coarse match; 
           pyramidal LK refinement; 
           forward-backward validation 
           \hfill (\S\ref{sec:search_region})
    \State Track failures: drop from 
           $\mathcal{I}$; marginalise from 
           $\mathcal{F}$ 
           \hfill (\S\ref{sec:augment})
    \State Stereo promotion: NCC search in 
           other camera for mono points at $k$ 
           \hfill (\S\ref{sec:cv})

    \Statex
    \Statex \textbf{--- EKF update ---}
    \State Per-feature NIS gate at 
           $\chi^2_{\alpha_{\mathrm{NIS}},\,
           \nu_i}$; moving points demoted from 
           $\mathcal{F}$ to $\mathcal{I}$ 
           \hfill (\S\ref{sec:feature_movement})
    \State Joint consistency check on 
           surviving inliers; if fail: demote 
           all $\mathcal{F}$ to $\mathcal{I}$  
           \hfill (\S\ref{sec:feature_movement})
    \State Update on inliers plus gravity 
           pseudo-measurement 
           \hfill (\S\ref{sec:ekf_eqs}, 
           \S\ref{sec:measurement})

    \Statex
    \Statex \textbf{--- Post-update relative 
           pose ---}
    \State Recompute 
           $(\Delta\hat{T}^+,\,
           \boldsymbol{\Sigma}_{\Delta\xi}^+)$ 
           from $\hat{T}_{k-1}^+,\,\hat{T}_k^+$ 
           — this is the pose prior for the 
           joint solver 
           \hfill (\S\ref{sec:relative_pose})

    \Statex
    \Statex \textbf{--- Joint solver ---}
    \State Stage-2 points in 
           $\mathcal{F}_{\mathrm{pre}} \cup 
           \mathcal{I}$:
    \State \quad Classify correspondence type
           (SS, SM, MS, MM); MM gated out
           entirely for now
    \State \quad Build per-point position 
           priors from stereo triangulation 
           at $k{-}1$ and $k$ where available 
           \hfill (\S\ref{sec:point_priors})
    \State \quad Solve jointly with pose prior 
           $(\Delta\hat{T}^+,\,
           \boldsymbol{\Sigma}_{\Delta\xi}^+)$ 
           \hfill (\S\ref{sec:joint_solver})
    \State \quad Transform per-point 
           $(\hat{\mathbf{p}}_i^{B_k},\,
           \hat{\mathbf{v}}_i^{B_k},\,
           \boldsymbol{\Sigma}_i^{B_k})$ via 
           $\mathbf{T}_{\Delta,i}$

    \Statex
    \Statex \textbf{--- Feature admission ---}
    \State $\mathcal{F}_{\mathrm{pre}}$ points 
           with first stage-2 solve: velocity 
           consistency test 
           \hfill 
           (\S\ref{sec:admission_stats})
    \State \quad Pass: augment 
           $(\hat{\mathbf{p}}_i^{B_k},\,
           \boldsymbol{\Sigma}_{p,i}^{B_k})$ 
           into EKF 
           \hfill (\S\ref{sec:augment})
    \State \quad Fail: demote to $\mathcal{I}$

    \Statex
    \Statex \textbf{--- Replenish ---}
    \State Mark points past their
           (per-instance jittered) $n_{\max,i}$
           for retirement
    \State Pre-emptive replenish: points
           reaching $n_{\max,i}$ next frame get 
           replacements now
    \State Replenish 
           $\mathcal{F}_{\mathrm{pre}}$ and 
           $\mathcal{I}$ from keypoint pool; 
           NCC stereo match; stage-1 stereo 
           triangulation for new stereo points 
           \hfill (\S\ref{sec:cv}, 
           \S\ref{sec:depth_estimation})

    \Statex
    \Statex \textbf{--- Output ---}
    \State Extract core EKF state 
           $\hat{\mathcal{X}}_k^{\mathrm{core}},\,
           \mathbf{P}_k^{\mathrm{core}}$ 
           (strip feature rows/columns)
    \State Assemble $\mathcal{C}_k$ from EKF 
           (for $\mathcal{F}$), joint solver 
           (for stage-2 
           $\mathcal{F}_{\mathrm{pre}},\,
           \mathcal{I}$), and stereo 
           triangulation (for stage-1 
           $\mathcal{F}_{\mathrm{pre}},\,
           \mathcal{I}$)
    \State \Return 
           $\hat{\mathcal{X}}_k^{\mathrm{core}},\,
           \mathbf{P}_k^{\mathrm{core}},\,
           \Delta\hat{T}_{\mathrm{solver}},\,
           \boldsymbol{\Sigma}_{\Delta\xi}
           ^{\mathrm{solver}},\,
           \mathcal{C}_k$
    \State Push accumulator $k \to k{-}1$
    \State Discard the points marked for
           retirement; marginalise those in
           $\mathcal{F}$ from the EKF
           \hfill (\S\ref{sec:augment})
\end{algorithmic}

\noindent \textbf{Review note.} Retirement on age is
the intent. The code compares $n_{\max,i}$ against
the global $n_{\max}$ instead of the point's age, so
it never fires. The results below therefore have no
maximum point lifetime.
\newpage
\section{Evaluation}
\subsection{Setup}
We have implemented the algorithm in Python and evaluated it on the VID dataset \citep{zhang2022visualinertialdynamicalmultirotordataset}, using the sequence indoor\_loadless\_hovor\_3096.1g\_79.04s. The sequence is characterised by simple dynamics: no payload, no wind, and dominated by a hover state. These characteristics make the sequence appropriate for first validation. \\

\noindent 
The dataset provides calibration, stereo video sequences and ground truth making it ideal for evaluating dynamics coupled VO systems.  

\begin{figure}[H]
    \centering
    \includegraphics[width=0.6\textwidth]
        {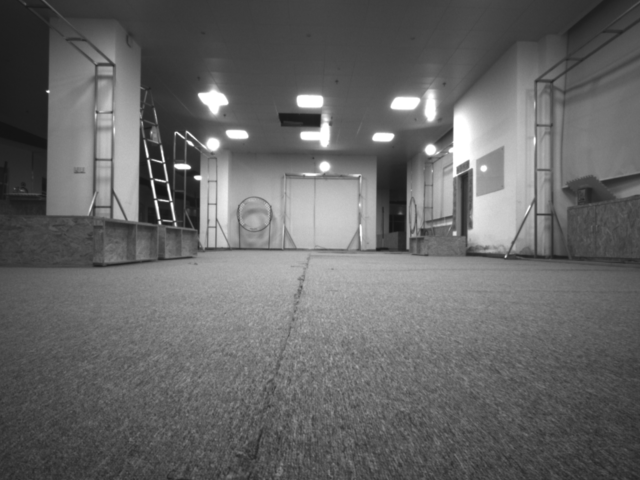}
    \caption{Representative left-camera frame from 
    the VID indoor sequence. Hangar interior at the 
    start of the bag, quadcopter stationary on the floor.}
    \label{fig:sample_frame}
\end{figure}

\noindent The sequence starts with the quadcopter stationary on the ground, then it takes off and climbs altitude, before it maintains hover, then finally it lands.

\subsection{State Estimate Accuracy}

\begin{figure}[H]
    \centering
    \includegraphics[width=\textwidth]
        {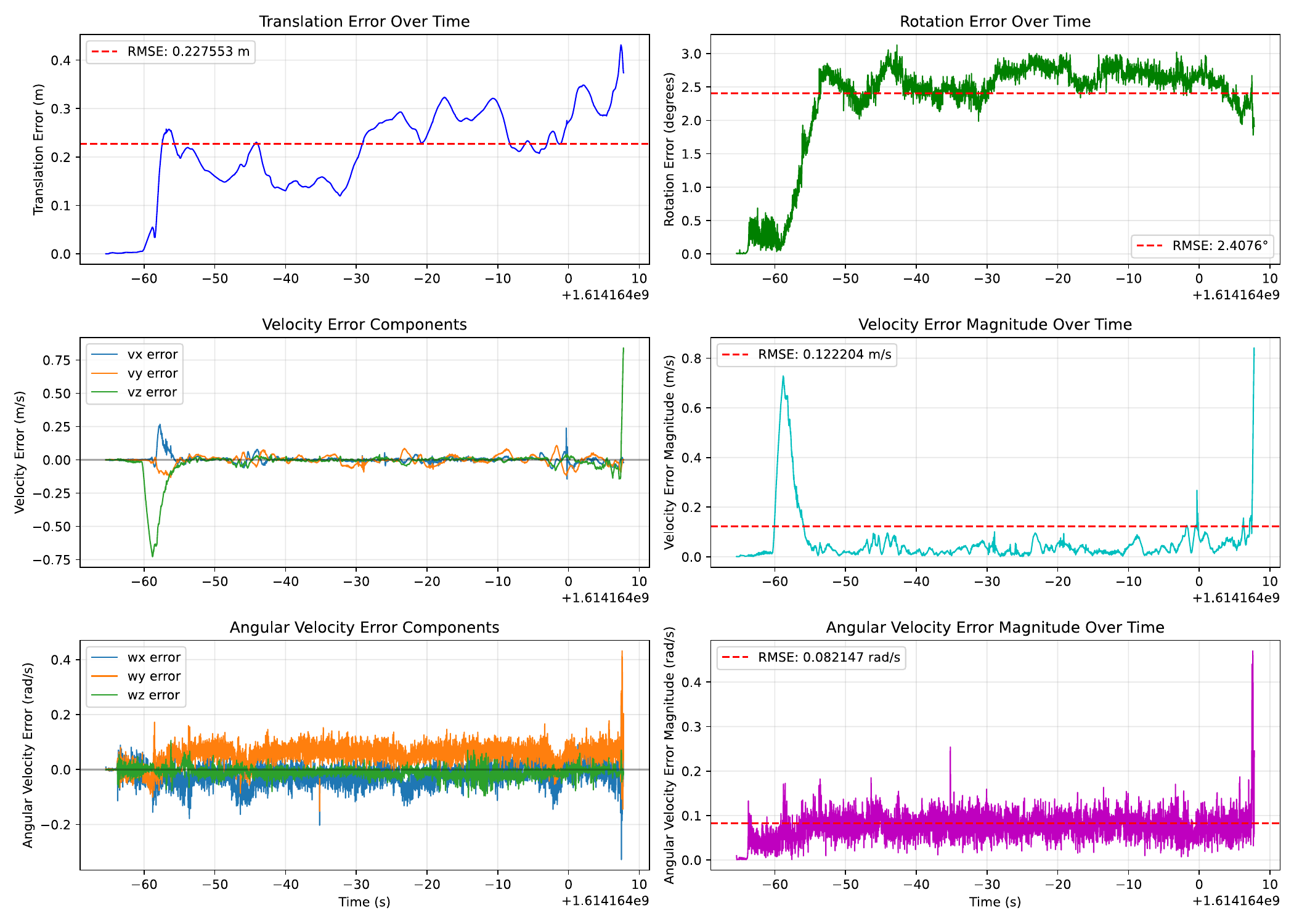}
    \caption{All but last 350 frames. Per-frame error magnitudes (translation, 
    rotation, body-frame linear and angular velocity) 
    against Vicon ground truth. Dashed red: RMSE over 
    the full evaluated window.}
    \label{fig:res_errors1}
\end{figure}
Overall our state estimates appear accurate, but degrade during the transients of take-off and landing, as illustrated by the large changes in error at the beginning and end of the time interval. This is expected due to the discontinuous nature of the normal force, but we could likely alleviate the transient by increasing the process noise associated with the disturbance vector. Measurements observing the disturbance such as an accelerometer would also alleviate the transient. \\

\noindent 
The process noise spectral densities
$\mathcal{Q}_u$, $\mathcal{Q}_d$ and
$\mathcal{Q}_\omega$ are priors, not values
validated against data. The filter is likely
overconfident as a result: the state collapses when
measurements pause for $\sim 10$ frames. A NEES
sweep on a held-out flight would settle it. \\

\noindent We are pleased with the accuracy but recognise the need for tuning and optimisation. For vision-only without IMU, this is reasonable first-implementation performance. Notably we observe a non-zero mean error in the pitch rate ($\omega_y$ error) and to a lesser extent in the roll rate ($\omega_x$ error), suggesting a bias in our model which could stem from a non-zero displacement between the centre of mass and the geometric centre.

\begin{figure}[H]
    \centering
    \includegraphics[width=\textwidth]
        {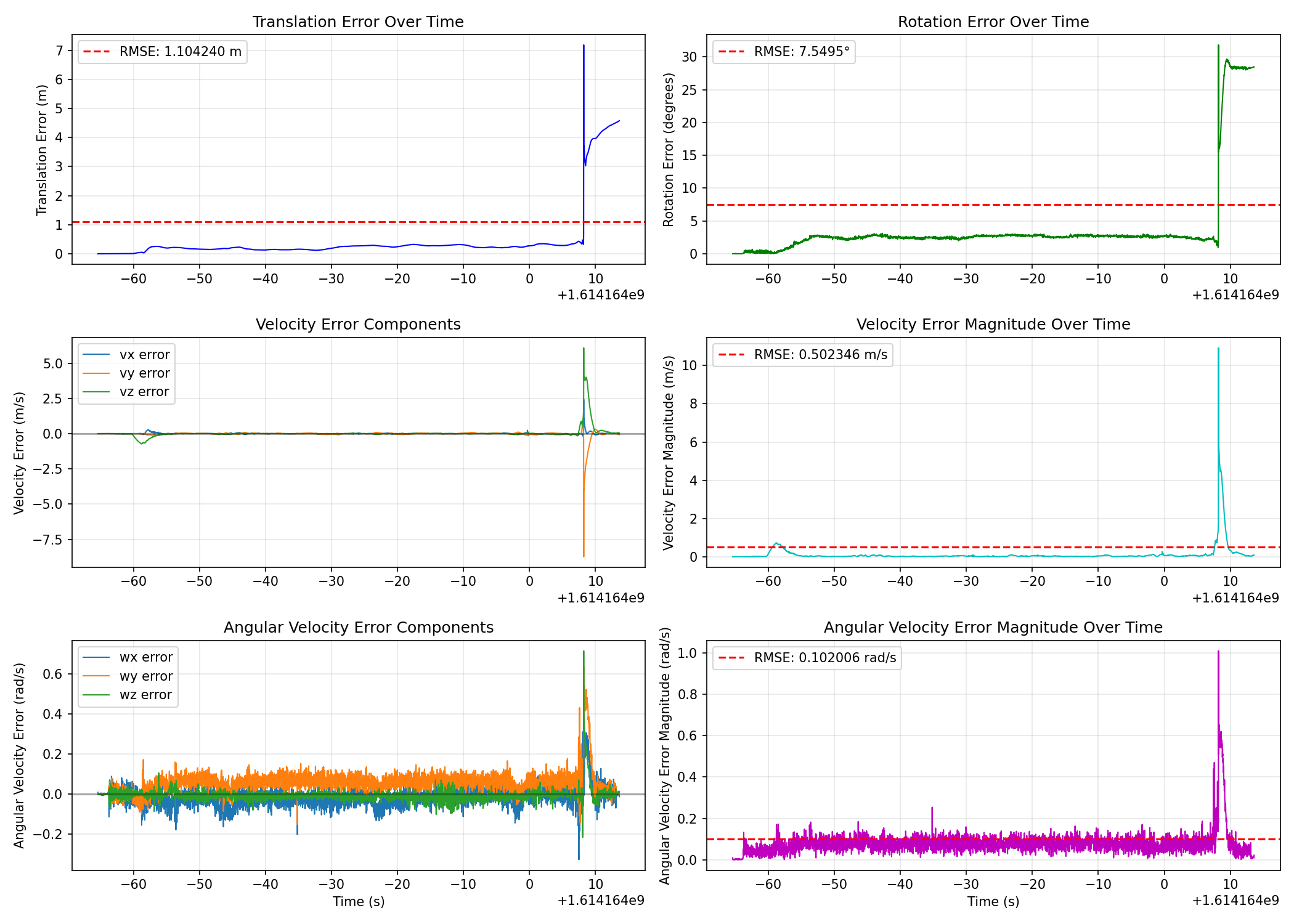}
    \caption{All frames. Per-frame error magnitudes (translation, 
    rotation, body-frame linear and angular velocity) 
    against Vicon ground truth. Dashed red: RMSE over 
    the full evaluated window.}
    \label{fig:res_errors2}
\end{figure}

\noindent Above we see the error plots for the entire sequence. This illustrates the sharp increase in error which takes place during landing. Notably the system is able to recover with time.

\begin{figure}[H]
    \centering
    \includegraphics[width=\textwidth]
        {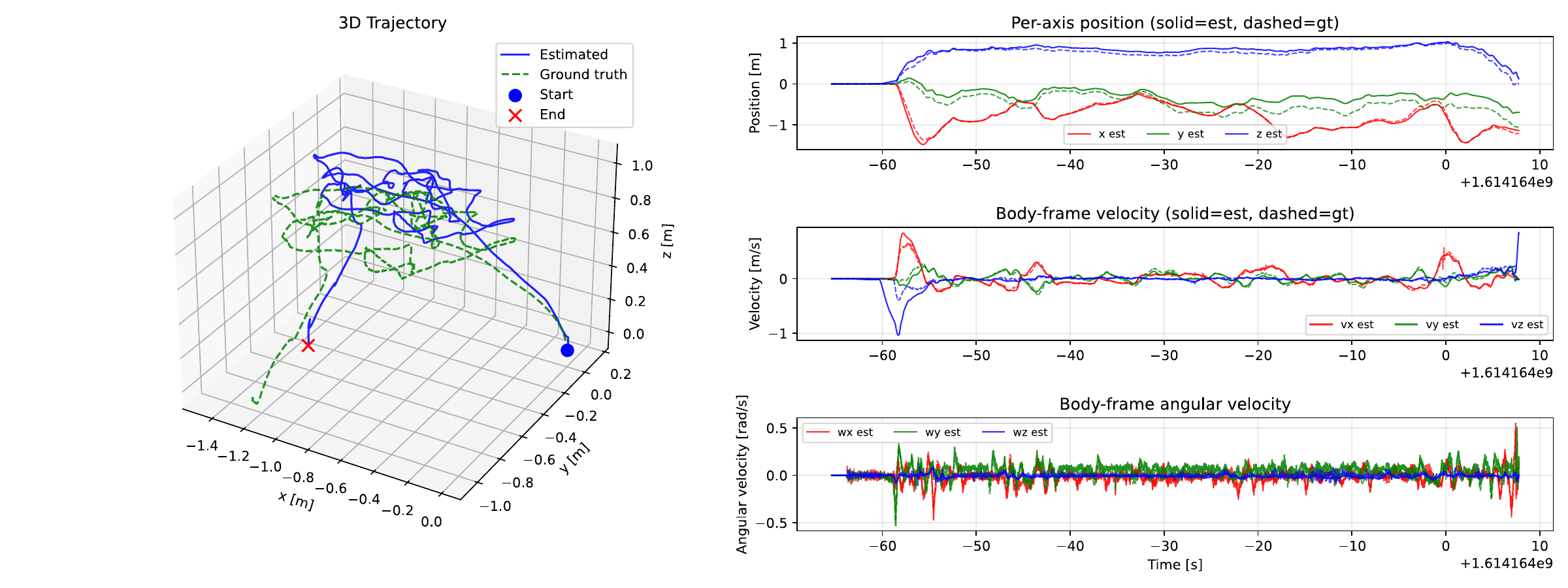}
    \caption{All but last 350 frames. Estimated vs ground-truth trajectory. 
    Left: 3D path. Right: per-axis position, body-frame 
    velocity, and angular velocity over time.}
    \label{fig:trajectory}
\end{figure}

\begin{table}[H]
    \centering
    \begin{tabular}{lrrr}
        \textbf{Quantity} & \textbf{RMSE} 
        & \textbf{Median} & \textbf{p95} \\
        \hline
        Translation [m] 
            & 1.104 & 0.229 & 3.964 \\
        Rotation [deg] 
            & 7.55  & 2.58  & 28.25 \\[1ex]
        Velocity magnitude [m/s] 
            & 0.502 & 0.034 & 0.457 \\
        \quad $|v_x|$ error [m/s] 
            & 0.065 & 0.010 & 0.065 \\
        \quad $|v_y|$ error [m/s] 
            & 0.273 & 0.019 & 0.101 \\
        \quad $|v_z|$ error [m/s] 
            & 0.417 & 0.010 & 0.420 \\[1ex]
        Angular velocity magnitude [rad/s] 
            & 0.102 & 0.075 & 0.130 \\
        \quad $|\omega_x|$ error [rad/s] 
            & 0.050 & 0.027 & 0.092 \\
        \quad $|\omega_y|$ error [rad/s] 
            & 0.082 & 0.057 & 0.115 \\
        \quad $|\omega_z|$ error [rad/s] 
            & 0.034 & 0.013 & 0.046 \\
    \end{tabular}
    \caption{All frames summary error metrics 
    (4733 frames, 79.0\,s), including ground 
    contact, takeoff, flight, and landing. The RMSE 
    figures are dominated by post-landing divergence.}
    \label{tab:summary_metrics}
\end{table}

\noindent The velocity numbers tell a clear story: median $\sim$1 cm/s, p95 (the 95th percentile) is 0.5 m/s. RMSE $\gg$ median means the error is highly concentrated in a small fraction of frames. That fraction is the takeoff and landing transient (see below). \\

\noindent Angular velocity is the opposite story: RMSE and median sit far closer together than for velocity, which points to a uniform noise floor rather than a transient. The $\omega_y$ channel is the worst, which confirms the visible bias in the plots. $\omega_z$ is the best and $\omega_x$ the middle. The asymmetry suggests, among others: residual $\mathbf{R}_{BC}$ misalignment (camera-body extrinsic), Vicon-marker-to-body alignment error, or an intrinsic limitation of vision-only rotation rate. \\

\noindent Translation median $<$ RMSE $\ll$ p95, suggests a small persistent offset (alignment or accumulated drift) and a transient spike.

\begin{table}[H]
    \centering
    \begin{tabular}{lrrrr}
        \textbf{Window [s]} 
            & \textbf{Rot [deg] med} 
            & \textbf{$|v_z|$ med} 
            & \textbf{$|v_z|$ p95} 
            & \textbf{Trans med} \\
        \hline
        0--5     & 0.19  & 0.005 & 0.018 & 0.002 \\
        5--10    & 0.52  & 0.277 & 0.689 & 0.152 \\
        10--20   & 2.45  & 0.009 & 0.048 & 0.177 \\
        20--40   & 2.51  & 0.006 & 0.023 & 0.154 \\
        40--60   & 2.74  & 0.007 & 0.032 & 0.278 \\
        60--70   & 2.61  & 0.018 & 0.063 & 0.261 \\
        70--73   & 2.23  & 0.057 & 0.146 & 0.318 \\
        73--75   & 20.15 & 1.697 & 4.001 & 3.415 \\
        75--79   & 28.35 & 0.111 & 0.276 & 4.317 \\
    \end{tabular}
    \caption{Error breakdown by time window over all frames. Takeoff occurs at $t\approx 8$\,s 
    (window 5--10). Landing occurs at 
    $t\approx 72$\,s (window 70--73). The 73--75\,s 
    window contains catastrophic divergence: 
    translation error grows from 0.32\,m to 3.4\,m 
    and rotation from 2.2° to 20° within 2\,s of 
    touchdown.}
    \label{tab:time_windows}
\end{table}

\subsection{State Diagnostics}

\begin{figure}[H]
    \centering
    \includegraphics[width=\textwidth]
        {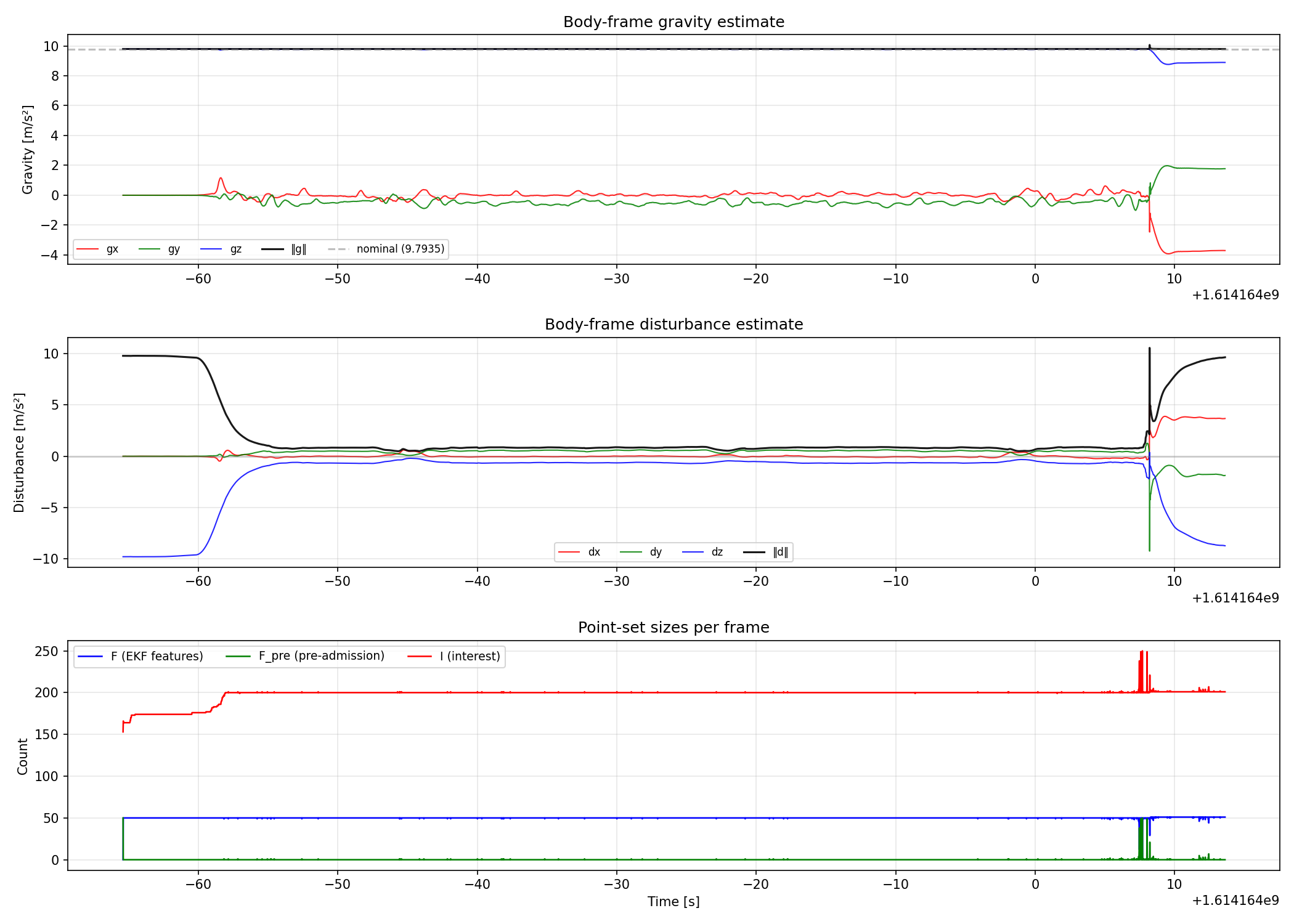}
    \caption{All frames. Internal state diagnostics. Top: body-frame 
    gravity estimate (per-axis and magnitude; grey 
    dashed: nominal $\|\mathbf{g}\| = 9.7935$\,m/s\textsuperscript{2}). 
    Middle: body-frame disturbance estimate. Bottom: 
    point-set sizes over time --- features in the EKF 
    (blue), pre-admission candidates (green), interest 
    points (red).}
    \label{fig:diagnostics}
\end{figure}

\noindent Gravity: $\|g\|$ stays locked at 9.79 throughout. The gravity pseudo-measurement is doing its job. Per-axis wandering ($\sim$0.5 m/s²) during the manoeuvre phase is gravity correctly being rotated through the body frame as the quadcopter tilts.\\

\noindent Disturbance: the most diagnostic plot. The quadcopter sits on the floor for $t<-55s$ with $d \approx (0,0,-9.8)$. The normal force exactly cancels gravity. Then the quadcopter lifts off, the filter correctly identifies that the normal force vanishes, and $\mathbf{d}^B$ transitions to near zero. The transient takes $\sim$3--5 seconds, during which the filter is splitting the residual acceleration between gravity and disturbance. \\

\noindent This is exactly the take-off discontinuity discussed above: the EKF's Gaussian disturbance model cannot represent the instantaneous transition. The filter takes seconds to re-converge. With accelerometer integration this would resolve in one frame. \\
\noindent Point counts: the retirement bug noted in Section~\ref{sec:algorithm_overview} is visible here. Both counts climb step-like to their ceilings and then stay there, $\mathcal{F}$ at 50 and $\mathcal{I}$ at $\sim$200. The steps come from points leaving instantly through marginalisation or dropping rather than ageing out, so nothing retires once a ceiling is reached. $F_{pre}\approx 0$ steady state. The one clear departure is during landing, where the NIS gate fires on high residuals between predicted and actual pixel observations. That gate is meant to catch moving points; here it catches a degraded state, marginalising the current feature points so new ones are selected.

\section{Future Work}
\label{sec:future_work}

This is a work in progress and a second version is in
preparation. Outstanding items on the system as described are
noted in place through the preceding sections rather than
collected here. \\

\noindent The direction beyond that, in brief. The estimator
splits into two processes with a defined interface: an ego-state
filter over a trailing window of image poses, taking rotor
speed, current, gyroscope and accelerometer alongside the pixel
channel, with biases, the inertia and aerodynamic
coefficients estimated online; and an environment state owning
correspondence, per-point depth, and the gate on what it hands
over. Then a C++ port with the vision primitives on programmable
hardware under a bounded sensor-to-actuator latency budget, and
closed-loop control on top.

\newpage
\addcontentsline{toc}{section}{References}
\bibliography{references}

\end{document}